\documentclass[journal]{IEEEtran}

\usepackage{hyperref}
\usepackage{cite}
\usepackage{amssymb}
\usepackage{mathtools}
\usepackage{amsmath}
\usepackage[pdftex]{graphicx}
\usepackage{multirow}
\usepackage[caption=false,font=normalsize,labelfont=sf,textfont=sf]{subfig}

\usepackage{xcolor}

\begin{document}

\title{ColorMapGAN: Unsupervised Domain Adaptation for Semantic Segmentation Using Color Mapping Generative Adversarial Networks}

\author{Onur Tasar,~\IEEEmembership{Student member,~IEEE,}
		S L Happy,~\IEEEmembership{Member,~IEEE,}
        Yuliya Tarabalka,~\IEEEmembership{Senior member,~IEEE, \\}
        Pierre Alliez
     
\thanks{O. Tasar, and P.~Alliez are with Universit{\'e} C{\^o}te d'Azur and Inria, TITANE team, 06902 Sophia Antipolis, France. E-mail: onur.tasar@inria.fr}\thanks{S L Happy is with Inria, STARS team, 06902 Sophia Antipolis, France.}
\thanks{Y. Tarabalka is with LuxCarta Technology, Parc d'Activit{\'e} l'Argile, Lot 119b, Mouans Sartoux 06370, France}}

\markboth{}%
{Shell \MakeLowercase{\textit{et al.}}: Bare Demo of IEEEtran.cls for IEEE Journals}

\maketitle

\begin{abstract}
Due to the various reasons such as atmospheric effects and differences in acquisition, it is often the case that there exists a large difference between spectral bands of satellite images collected from different geographic locations. The large shift between spectral distributions of training and test data causes the current state of the art supervised learning approaches to output unsatisfactory maps. We present a novel semantic segmentation framework that is robust to such shift. The key component of the proposed framework is Color Mapping Generative Adversarial Networks (ColorMapGAN), which can generate fake training images that are semantically exactly the same as training images, but whose spectral distribution is similar to the distribution of the test images. We then use the fake images and the ground-truth for the training images to fine-tune the already trained classifier. Contrary to the existing Generative Adversarial Networks (GANs), the generator in ColorMapGAN does not have any convolutional or pooling layers. It learns to transform the colors of the training data to the colors of the test data by performing only one element-wise matrix multiplication and one matrix addition operations. Thanks to the architecturally simple but powerful design of ColorMapGAN, the proposed framework outperforms the existing approaches with a large margin in terms of both accuracy and computational complexity.
\end{abstract}


\IEEEpeerreviewmaketitle

\section{Introduction}\label{sec:introduction}
\IEEEPARstart{W}{ith} the continuous proliferation and improvement of satellite sensors, numerous new generation satellite missions have been created, which has made it possible to collect huge amounts of data. The massive satellite data have introduced new challenges to the remote sensing community. \emph{Semantic segmentation} or \emph{dense labeling} is the task of assigning a thematic label to each pixel in the image. Without a doubt, among the challenges that the remote sensing community is facing today, dense labeling of the satellite images is one of the most important one, as a good solution for this problem is of paramount importance to generate and automatically update the maps.

\begin{figure}
\centering
\subfloat[Training image\label{fig:intro_training_image}]{\includegraphics[width=0.49\linewidth]{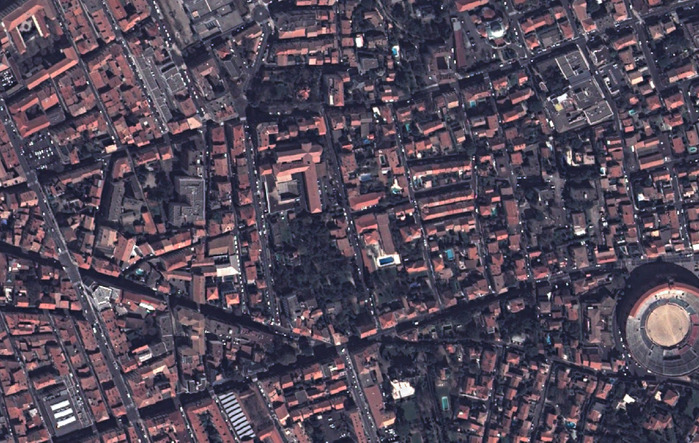}}
\hfill
\subfloat[Test image\label{fig:intro_test_image}]{\includegraphics[width=0.49\linewidth]{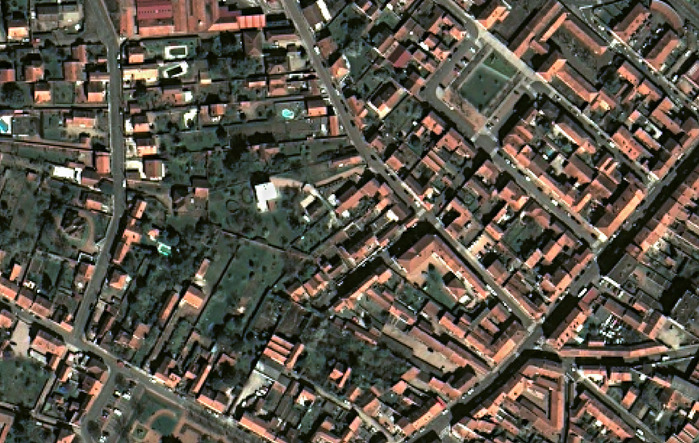}}
\hfill
\subfloat[Ground-truth for (b)\label{fig:intro_test_gt}]{\includegraphics[width=0.49\linewidth]{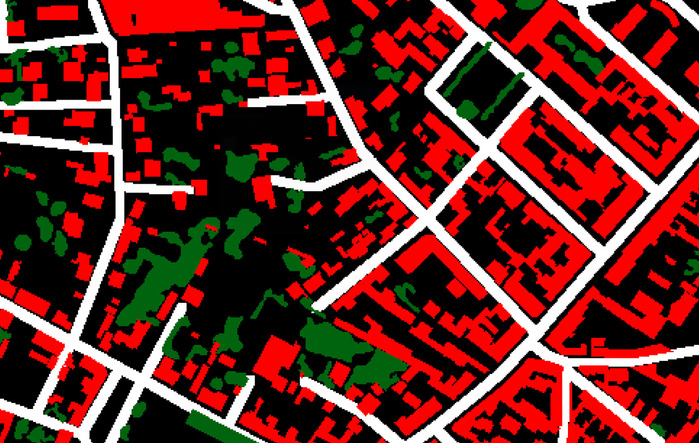}}
\hfill
\subfloat[Predicted map for (b)\label{fig:intro_test_pred}]{\includegraphics[width=0.49\linewidth]{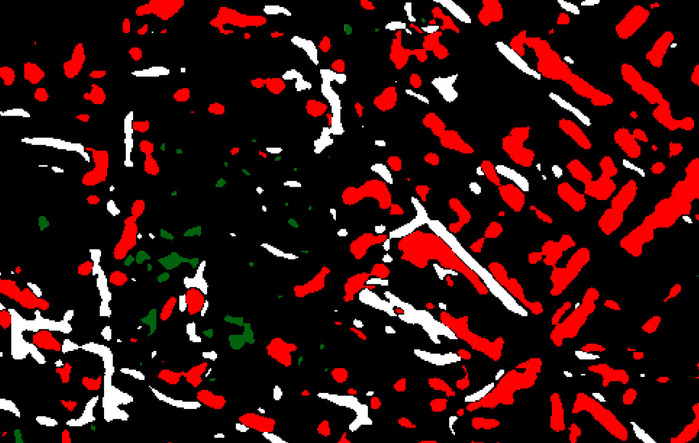}}
\caption{An illustration for the domain adaptation problem, where we depict training and test images, the ground-truth for the test image, and the predicted map by U-net. In the ground-truth and in the predicted map, red, green, and white pixels correspond to building, tree, and road classes, respectively.}
\label{fig:intro}
\end{figure}

Over the last few years, \emph{convolutional neural networks (CNNs)} have become the most commonly used tool for the task of semantic segmentation. In particular U-net\cite{ronneberger2015u} and its variants \cite{iglovikov2018ternausnet, iglovikov2018ternausnetv2, khalel:hal-02276549} are receiving a growing attention due to their great success in different domains such as medical imaging and remote sensing. The main limitation of the aforementioned CNNs is their extreme sensitivity to the training data. Although they perform very well when both training and test data come from the same distribution \cite{maggiori2017high}, their performance severely decreases when there is a large difference between spectral bands of the training and the test images. Considering that nowadays new generation satellites with a short revisit time acquire huge amounts of images from different parts of the world, one cannot assume that the distributions of the images are always similar. In addition, depending on when and where the data are collected, a large intra-class variability might be encountered in remote sensing images. For instance, color distributions of the same objects may greatly differ in the images captured in different times of the day as a result of the illumination difference. Similarly, because of the atmospheric effects, in some cases, even the images collected by the same satellite sensors might have very different radiometry, which makes the segmentation task even harder. Fig.~\ref{fig:intro} illustrates such an example, where color distributions of the training and the test data collected by the same satellite are noticeably different. As shown in the figure, even a well trained U-net, which is considered as the state of the art in semantic segmentation of satellite images\cite{huang2018large}, generates a poor map.

\emph{The unsupervised domain adaptation} assumes that annotations for any part of the test data are not available, and aims at generating high quality segmentations even when there is a large domain shift between the training and the test images. In such a setting, in order to increase the generalization capabilities of the CNNs, one of the simplest and most common methods is to diversify the training data by applying various data augmentation techniques such as gamma correction and random contrast change \cite{tasar2018incremental, tasar:hal-02276543}. However, even though these techniques help the CNNs to generalize slightly better, when spectral difference between the training and the test data is huge, the improvement is usually insufficient. Although a good solution for domain adaptation is indispensable for various applications, this problem has not been investigated intensively in the field of remote sensing. The limitations pointed out in this section have motivated us to design a methodology for this problem.

\subsection{Related Work}\label{sec:related_work}
In this section, we summarize the existing approaches on domain adaptation in the fields of computer vision and remote sensing.

\textit{Computer Vision:} The majority of the proposed approaches in the literature are based on the idea of \emph{aligning distributions} of both training and test images in a common space using \emph{generative adversatial networks (GAN)} so that the trained classifier could segment the test images well. FCNs in the Wild \cite{hoffman2016fcns} reduces the global distribution difference between the training and the test images by minimizing an adversarial loss. Aligning the distributions of the extracted features from both the training and the test images is also a common approach~\cite{sankaranarayanan2018learning, murez2018image}. The distribution alignment could also be performed in multiple layers of the network instead of only the final output~\cite{tsai2018learning, huang2018domain}. Hong~\textit{et~al.} add random noise to the training images~\cite{hong2018conditional}. They observe that perturbing the training images helps the GAN to adapt better to the test data. Saito~\textit{et~al.} introduces a maximum classifier discrepancy based approach, which alings the distributions by making use of class-specific decision boundaries~\cite{saito2018maximum}. The framework presented by Chen~\textit{et~al.} utilizes Google Street View to collect unannotated images and to use their features. Curriculum Domain Adaptation (CDA)~\cite{zhang2017curriculum} performs the alignment both globally and on the generated super-pixels. Reality Oriented Adaptation (ROAD)~\cite{chen2018road} comprises two loss functions. The first one forces the weights of the trained model from the training images with the weights of the pre-trained model from ImageNet~\cite{russakovsky2015imagenet} to be as similar as possible. The second one handles the spatial aware domain adaptation by dividing the training and the test images into grids and minimizing distance between the grids of the training and the test images.

There have been attempts to solve the domain adaptation problem by \emph{regularizing} or \emph{normalizing specific layers} of the network, or by \emph{self-learning}. The Fully Convolutional Tri-branch network (FCTN)\cite{zhang2018fctn} is one of the methods that falls into the category of self-learning. The presented network architecture consists of one encoder and three decoders. Two of the decoders pseudo-label the test image, the other one learns from the pseudo-labels and the test image. A class-balanced self learning approach is proposed in CBST\cite{zou2018unsupervised}. Introducing new normalization method, regularization technique, or new loss functions that are specific for domain adaptation problem is invesigated in~\cite{romijnders2018domain, saito2017adversarial, pan2018two, zhu2018penalizing}. Romijnders~\textit{et~al.} discuss the limitations of the traditional normalization methods such as batch normalization, and propose a new domain agnostic normalization layer that is more suitable for domain adaptation\cite{romijnders2018domain}. Saito~\textit{et~al.} introduce a new adversarial dropout regularization technique\cite{saito2017adversarial}. The IBN-Net \cite{pan2018two} combines the batch normalization with the instance normalization~\cite{ulyanov2016instance} to increase the generalization capability. Zhu~\textit{et~al.} proposes a new conservative loss~\cite{zhu2018penalizing}.

Another way to approach the unsupervised domain adaptation problem is to perform \emph{image to image translation} (I2I) between a source and a target domain. The I2I approaches aim at generating fake source images that are statistically indistinguishable from the target images. The existing I2I approaches in the literature can perform one to one \cite{zhu2017unpaired, liu2017unsupervised} or many to many \cite{huang2018multimodal, lee2018diverse} translations. If one can generate fake training images, which are style-wise consistent with the test images and semantically consistent with the original training images, the fake images could be used to train a model from scratch or to fine-tune the already trained model on the original data. The main drawback here is that usually the fake training images generated by the I2I approaches contain artificial objects and artifacts, which do not exist in the original images. Hence, annotations for the original training images and the generated fake training images do not match. As a result, the model learns wrong information. To overcome this limitation, CyCADA~\cite{hoffman2017cycada} segments the original and the fake training images using the classifier trained on the original training data, and minimizes cross entropy loss between the segmentations. However, if the domain shift between the training and the test data is large, the classifier trained from the original training images will segment the fake training image poorly. If the segmentation for the training images is very good, but the predicted map for the fake training images is extremely noisy, we cannot expect adding such a loss to prevent artificial objects appearing. Another way to enrich the training data could be to perform neural style transfer, where content of an image and style of another image are combined~\cite{gatys2016image, gatys2016preserving, johnson2016perceptual, ulyanov2016texture}. However, they also cannot guarantee that the semantic structures will be preserved. 

\textit{Remote sensing:} As thoroughly explained in the overview paper by Tuia~\textit{et al.}~\cite{tuia2016domain}, the domain adaptation methods in the field of remote sensing can be divided into four categories: selection of invariant features, adaptation of classifiers, active learning, and adaptation of the data distributions. The main goal of the methods falling into the first category is to find a subset of features from the training data, which is representative for the test data. For instance, Bruzzone~\textit{et al.} present a methodology~\cite{bruzzone2009novel} aiming at selecting the features that are both invariant to the test data and discriminative for the classes of interest by defining a new criterion function quantifying both tasks. Persello~\textit{et al.} describe a method that solves the same selection problem by a kernel based method~\cite{persello2015kernel}.

In order to adapt the classifier to the test data, the existing approaches usually either choose to adapt the classifier to the unlabeled test data directly, or favor an active learning approach, where the annotator labels a small number of representative samples from the test data to update the already trained classifier. Bruzzone~\textit{et al.} describe an approach that updates the parameters of the previously trained classifier on the unlabeled test data via the expectation-maximization (EM) algorithm~\cite{bruzzone2001unsupervised}. This approach has then been extended to another framework, where a cascade classification operation is performed~\cite{bruzzone2002partially}. In~\cite{bruzzone2002multiple}, multiple cascade classifiers are used. Bruzzone and Marconcini propose the domain adaptation support vector machine (DASVM)~\cite{bruzzone2009domain}. If adapting the classifier by an active learning approach is preferred, we can obviously expect to get better results, since some portion of the test data is provided to modify the classifier. On the other hand, the selection of the samples and labeling them manually can be too costly. In the approach proposed by Persello and Bruzzone~\cite{persello2012active}, the classifier iteratively learns from a small number of newly added samples from the test data and removes some of the training samples whose distribution does not fit with the distribution of the test data. The SVM based method presented by Matasci~\textit{et al.} also follows a similar approach~\cite{matasci2012svm}. Another kernel based active learning approach is introduced by Deng~\textit{et al.}~\cite{deng2018active}. The usage of neural networks for active learning is studied in~\cite{ghassemi2019learning}.

The adaptation of the data distributions is a common method. To do this, one can either align the distributions of the training and the test data in a common space, or align the distribution of the training data to the distribution of the test data. In both cases, we expect the model trained from the aligned training data to generate better segmentation. The aforementioned alignment can be performed by histogram matching~\cite{inamdar2008multidimensional}, \cite{Gonzalez}, graph matching~\cite{tuia2012graph}, kernel principal component analysis (KPCA)~\cite{nielsen2009kernel}, color constancy algorithms~\cite{agarwal2006overview}, or minimizing the statistical distance between the training and the test data~\cite{matasci2015semisupervised} by the maximum mean discrepancy (MMD)~\cite{matasci2015understanding}. Ma~\textit{et al.} correct the domain shift between the training and the test data for each class by centroid and covariance alignment~\cite{ma2018centroid}. Gross~\textit{et al.} propose a nonlinear feature normalization method for the alignment~\cite{gross2019nonlinear}. A new methodology performing tensor alignment is presented in~\cite{qin2018tensor}. Recently, Courty~\textit{et al.} have used the optimal transport for domain adaptation~\cite{courty2014domain, courty2016optimal}. The optimal transport for domain adaptation of remote sensing data is studied in~\cite{tardy2019assessment}. Besides, after GANs had become popular in the field of computer vision, some papers have studied their application to remote sensing. For instance, Benjdira~\textit{et al.}~\cite{benjdira2019unsupervised} use CycleGAN~\cite{zhu2017unpaired} to generate fake training data that resemble test data. The generated fake training data are then used to fine-tune the already trained model from the original training data. However, unlike the computer vision benchmarks, remote sensing images contain a lot of heterogeneous and complex structures. As a result, the quality of the fake remote sensing images generated by CycleGAN is usually not as good as desired.

\subsection{Contributions}
We study the problem of unsupervised domain adaptation for semantic segmentation, where training and test images are collected from completely different geographic locations, they have significant spectral distribution difference, and annotations for any part of the test images are not available. The way we approach the problem methodologically resembles to the image to image translation (I2I) approaches; we generate fake training images as if they came from the distribution of the test images. Instead of working on images acquired over separate geographic extents, if we had paired training and test images (e.g., images acquired from exactly the same geographic extent but in different day time), the problem would be easier, we could perform paired I2I \cite{isola2017image}. 

The main challenge here is to generate fake training images that look like test images without impairing the semantic meaning in the original training images, even when the training and the test images are unpaired. If the semantic structures of the original training images are preserved in the fake training images, we can utilize the annotations for the original training images as well as the fake training images to fine-tune the already trained model from the original training data. When generating fake training images with the style of the test images, most of the I2I approaches presented in Sec.~\ref{sec:related_work} fail to preserve the exact structures of the original training images, especially when we deal with satellite images containing a lot of complex and heterogeneous objects. Thus, the fake training images and the annotations do not mach, which leads the model to learn incorrect information. 

Contributions of this work are as follows:

\textit{1) ColorMapGAN:}
Our main contribution is novel ColorMapGAN, which can generate fake training images that are semantically exactly the same as the original training images (i.e., location and shape of the objects such as roads, building, trees, etc. are exactly the same in the fake and in the original training images) and that are visually indistinguishable from the test images (i.e., the objects such as trees and buildings in the fake training images and the test images have similar spectral distributions). To do this, ColorMapGAN transforms the colors of the training images into the colors of the test images without doing any structural changes on the objects of the training images.

\textit{2) Higher accuracy:} In our experiments, we perform city to city domain adaptation. We utilize the fake training images generated by our approach and by CycleGAN \cite{zhu2017unpaired}, UNIT~\cite{liu2017unsupervised}, MUNIT~\cite{huang2018multimodal}, DRIT~\cite{lee2018diverse}, histogram matching~\cite{Gonzalez}, and gray world algorithm~\cite{buchsbaum1980spatial} to fine-tune the model trained from the original training images. We also compare our approach with AdaptSegNet~\cite{tsai2018learning}, which aims at adapting the classifier to the test data directly. Despite its architecturally simple design, our framework enables to generate predicted maps with much higher accuracy than the others.

\textit{3) Lower complexity:} Unlike the existing GANs in the literature, our generator does not perform any convolution or pooling operations. It transforms the colors of the training images with only one element-wise matrix multiplication and one matrix addition. Because our generator is architecturally substantially simpler compared to the existing GANs, the training time of ColorMapGAN is significantly lower than the other learning based approaches. In our experiments, we compare our approach with CycleGAN~\cite{zhu2017unpaired}, UNIT~\cite{liu2017unsupervised}, MUNIT~\cite{huang2018multimodal}, DRIT~\cite{lee2018diverse}, histogram matching~\cite{Gonzalez}, and gray world algorithm~\cite{buchsbaum1980spatial} in terms of running times.

\section{Generative Adversarial Neural Networks}\label{sec:gans}
In machine learning, we can divide the models trained in a supervised setting into two groups: discriminative and generative models. In the field of image analysis, the discriminative models are usually trained to learn a mapping from a high dimensional input to class labels as in image categorization and segmentation problems. On the other hand, the generative models aim to estimate the distribution of the data samples so that new samples could be drawn from the estimation. In 2014, Goodfellow \textit{et~al.} proposed the generative adversarial networks (GANs) \cite{goodfellow2014generative}, which is a novel approach to train a generative model. 
 
GANs usually comprise a generative model $G$ and a discriminative model $D$. The goal of $G$ is to estimate the distribution of the real data and to output fake data from the estimation. $G$ takes a random noise $z$ as input, and represents a mapping to data space $G(z)$. We denote the distribution of the real data $x$ by $p(x)$ and a prior on input noise variables by $p(z)$. Let us assume that the real data $x$ and the fake data $G(z)$ are indicated by 1 and 0, respectively. $D$ outputs a scalar between 0 and 1, and aims to maximize the probability of labeling $x$ and $G(z)$ correctly. In other words, the goal of $D$ is to discriminate between the real and the fake data. The objective function for $D$ that is maximized during training is described as:
\begin{equation}\label{eq:D_objective_fun}
\max_{\substack{D}}V(D) = \mathbb{E}_{x \sim p(x)}\text{log}[D(x)] ~+~ \mathbb{E}_{z \sim p(z)}[\text{log}(1 - D(G(z)))],
\end{equation} 
where $\mathbb{E}$ is the expected value. $G$ is simultaneously trained to minimize the objective function defined as:
\begin{equation}\label{eq:G_objective_fun}
\min_{\substack{G}}V(G) = \mathbb{E}_{z \sim p(z)}[\text{log}(1 - D(G(z)))].
\end{equation}
As a result, the minimax game played between G and D could be formulated as:
\begin{equation}\label{eq:DG_minimax_game}
\begin{split}
\min_{G}\max_{D}V(D, G) = \mathbb{E}_{x \sim p(x)} \text{log} D(x) ~+~ \\ 
\mathbb{E}_{z \sim p_{(z)}} \text{log} (1 - D(G(z))).
\end{split}
\end{equation}
Once $D$ and $G$ are simultaneously trained for a sufficiently long time, $D$ becomes good at discriminating the fake and the real data, and $G$ becomes better at generating fake data that are indistinguishable from the real data. 

\begin{figure*}
	\centering
	\includegraphics[width=0.9\linewidth]{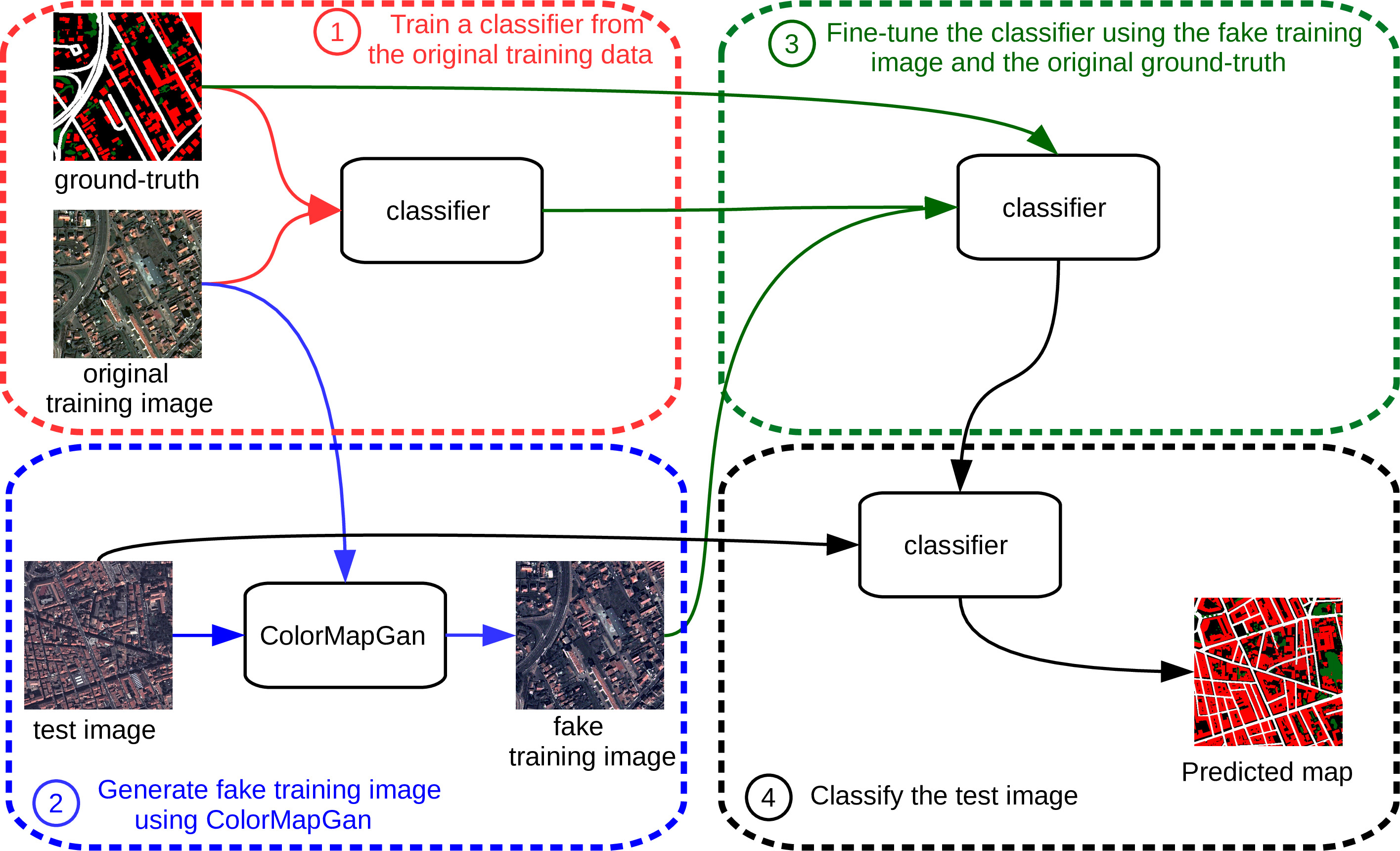}
	\caption{The overall framework.}
	\label{fig:overall_framework}
\end{figure*}

Although GAN works well with shallow multi-layer perceptrons, they suffer from instability problems during training when a more complex network is used. Several approaches have been presented to address the instability issues. In DCGAN \cite{radford2015unsupervised}, instead of multilayer perceptrons, deep convolutional networks are used in both $G$ and $D$, and certain architectural constraints are introduced for a more stable training. In WGAN \cite{arjovsky2017wasserstein}, rather than the logarithms in Eq.~\eqref{eq:DG_minimax_game}, Wasserstein distance is used to compute distance between the distributions, and gradient clipping is applied in the training stage. WGAN-GP\cite{gulrajani2017improved} is an extension of WGAN, where a gradient penalty is performed to solve the limitations of the gradient clipping. LSGAN \cite{mao2017least} proves that adopting the least squares loss function in Eq.~\eqref{eq:DG_minimax_game} allows more stabilized training.

Finally, the original GANs were extended to conditional GANs in \cite{mirza2014conditional}, where instead of generating the fake data from noise $z$, both $G$ and $D$ are conditioned on some extra information $y$. $y$ can be class labels or data from other modalities. In this architecture, $G$ learns a mapping from combination of $z$ and $y$ to the data space. If we consider $y$ as the source data and $x$ as the target data, $G$ aims to learn a mapping from source domain to target domain. Inspired from this idea, conditional GANs have been used for several I2I works \cite{isola2017image, zhu2017unpaired}. 

\section{Methodology}
\subsection{The Overall Framework}
Fig. \ref{fig:overall_framework} depicts the overall framework that consists of 4 steps as follows:
\begin{enumerate}
\item \emph{Training the initial classifier:} We train a U-net on the original training data.
\item \emph{ColorMapGAN:} We generate fake training images that are semantically exactly the same as the original training images, but visually as similar as possible to the test images using the proposed ColorMapGAN.
\item \emph{Fine-tuning:} We fine-tune the model obtained in step 1 using the fake training images and the ground-truth for the original training images.  
\item \emph{Classification:} Finally, we classify the test images.
\end{enumerate}

We use a slighly modified version of U-net \cite{ronneberger2015u} as the classifier. We replace rectified linear activation units (ReLU) by leaky rectified linear activation units (Leaky-ReLU) for a better performance \cite{xu2015empirical}. We also remove the batch normalization operation in each layer, since it uses the memory inefficiently. In the framework, the steps 1, 3, and 4 are self-explanatory, whereas step 2 needs further explanation.

\begin{figure*}
	\centering
	\includegraphics[width=0.95\linewidth]{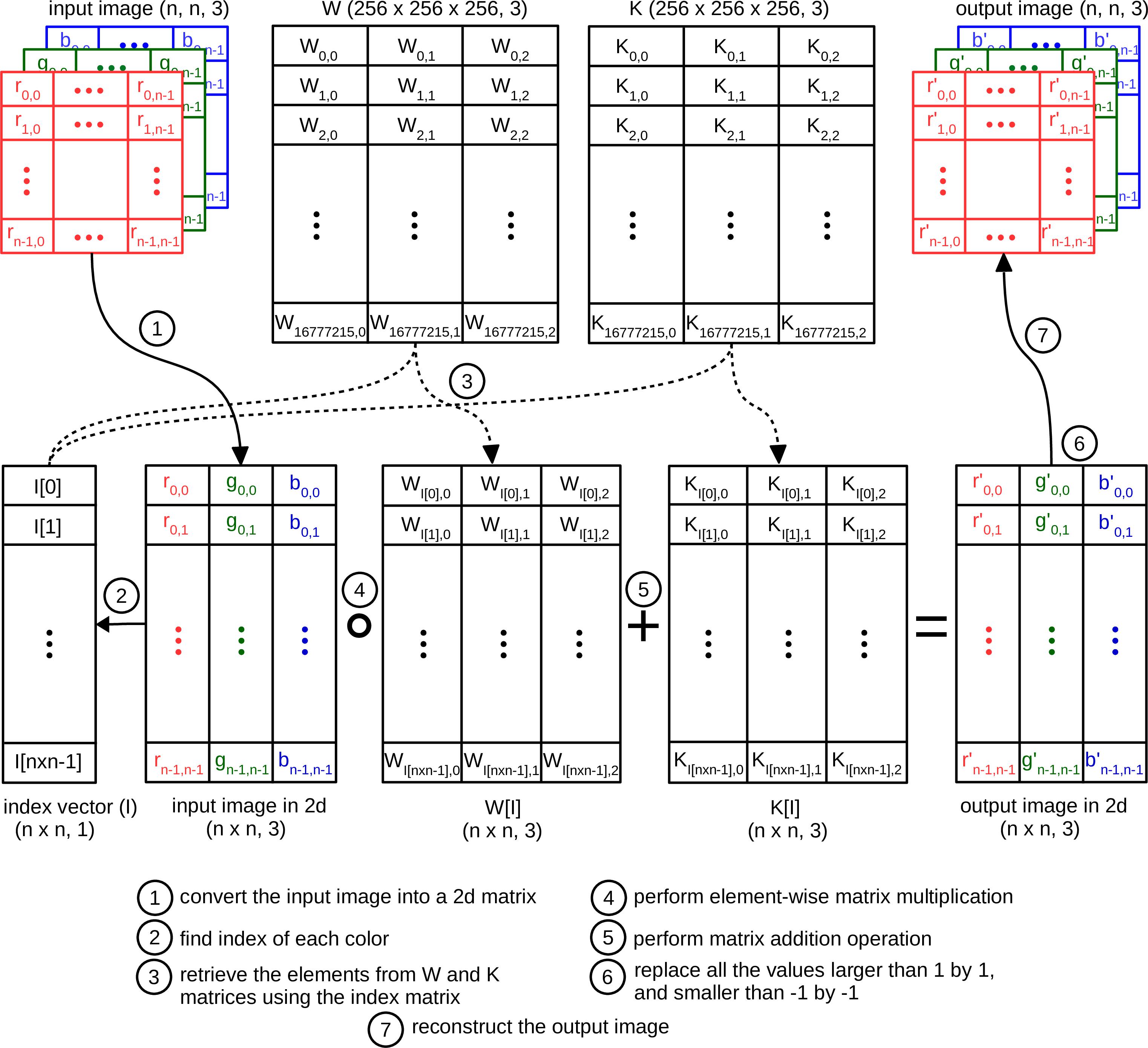}
	\caption{The overall flowchart for the feedforward pass of the generator.}
	\label{fig:generator}
\end{figure*}

\subsection{ColorMapGAN}
%In this section, we explain ColorMapGAN which is the main contribution of this study. 

\textit{1) Generator:} The novelty of the proposed ColorMapGAN is in the architecturally simple but powerful design of its generator.

We denote a set of training image patches by $X=\{x_1, x_2, \ldots, x_N \}$ and a set of test image patches by $Y=\{y_1, y_2, \ldots, y_M\}$. $G(X)$ corresponds to the set of fake training patches generated by $G$. The goal of $G$ is to generate $G(X)$, whose spectral distribution is as similar as possible to the distribution of $Y$, while keeping $G(X)$ and $X$ semantically exactly the same. Contrary to the existing GANs in the literature, we do not use convolutional or pooling layers in $G$ to preserve the exact semantics of $X$ in $G(X)$. 

Let us assume that $X$ and $Y$ are composed of 8 bit images comprising red, green, and blue channels. We denote by $R=\{0, 1, \ldots, 255\}$, $G=\{0, 1, \ldots, 255\}$, and $B=\{0, 1, \ldots 255\}$, the values each color band of the pixels can take. We denote all the possible $16,777,216$ ($256 \times 256 \times 256$) colors by $RGB$, which is defined as:
\begin{equation}
RGB = R \bigtimes G \bigtimes B,
\end{equation}
where $\bigtimes$ stands for the Cartesian product. In order to transform $RGB$ to another color matrix $R'G'B'$, we use a scale $W$ and a shift $K$ matrices with the same size as $RGB$. $R'G'B'$ could be computed as:
\begin{equation}\label{eq:rgb_to_rpgbgb}
R'G'B' = RGB \circ W + K,
\end{equation}
where $\circ$ is the element-wise product. The only learnable parameters of our $G$ are $W$ and $K$. Before the training starts, we initialize $W$ with ones and $K$ with zeros. Hence, at the end of the first iteration, the input and the output of $G$ are exactly the same. 

The main bottleneck for computing Eq.~\eqref{eq:rgb_to_rpgbgb} is the computational complexity. Since each of $RGB$, $W$, and $K$ matrices has more than 50 millions of elements (256 $\times$ 256 $\times$ 256 $\times$ 3), it is not feasible to perform the operation defined in Eq.~\eqref{eq:rgb_to_rpgbgb} on a GPU. However, the number of colors in a training image patch is much lower than the number of all the possible colors. Therefore, it is sufficient to update only the elements of $W$ and $K$ which transform the colors that are available in the training patch. To do this, we use an index vector $I$ that is defined as:
\begin{equation}
I = r \times 256 \times 256 + g \times 256 + b,
\end{equation}
where $r$, $g$, $b$ are red, green, blue values of all the pixels in the training patch. After the elements of $I$ are found, we normalize and center each $x_i \in X$ and $y_j \in Y$ by first dividing by $127.5$ and then subtracting $1$ so that each color channel ranges between $-1$ and $1$. We then partially update $R'G'B'$ as:
\begin{equation}
R'G'B'[I] = RGB[I] \circ W[I] + K[I],
\end{equation}
where $[\cdot]$ operation corresponds to retrieving the rows of an arbitrary 2D matrix indexed by the  given vector. Then, in $R'G'B'[I]$, we replace the elements that are bigger than 1 and smaller than -1 by 1 and -1, respectively. To range all the values in $R'G'B'[I]$ between 0 and 255, we then use the denormalization function $DN$ that is defined as:
\begin{equation}
DN(p) = \lfloor (p + 1) \times 127.5 \rfloor,
\end{equation}
where $p$ is a 2D input matrix. The final output of $G$ can be obtained by reshaping $DN(R'G'B'[I])$ back to the shape of the input patch. In each training iteration, we update only $W[I]$ and $K[I]$. 

Fig.~\ref{fig:generator} illustrates the overall flowchart for the feedforward pass of $G$.

\begin{figure}
	\centering
	\includegraphics[width=0.9\linewidth]{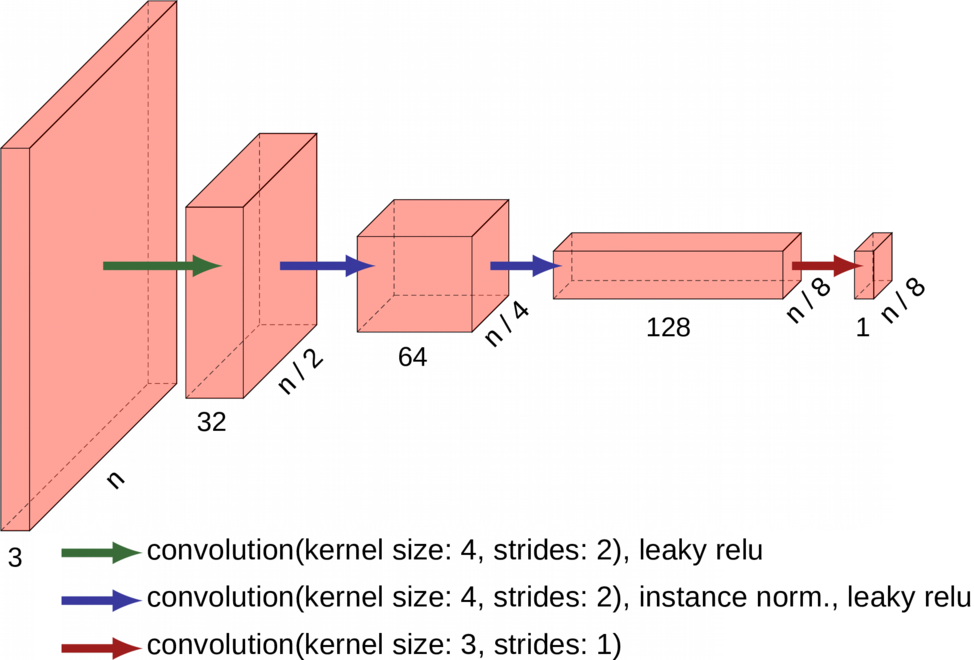}
	\caption{The architecture of the discriminator. n corresponds to the patch size, and the number below each layer is the depth.}
	\label{fig:discriminator}
\end{figure}

\textit{2) Discriminator: } The discriminator in ColorMapGAN is architecturally the same as the discriminator of CycleGAN\cite{zhu2017unpaired} (see Fig.~\ref{fig:discriminator}). Instead of outputting a single scalar for the whole image patch to determine if the patch is real or fake, this discriminator generates a two dimensional matrix. Each element of the matrix is used to locally determine whether the input patch is real or fake. We then take average of all the elements of the matrix to yield a final value. 

As we mention in Sec.~\ref{sec:gans}, GANs suffer from the instability issues; therefore, other objective functions have been proposed as an alternative to Eqs.~\ref{eq:D_objective_fun}, \ref{eq:G_objective_fun}, and \ref{eq:DG_minimax_game}. We prefer to use the functions presented in LSGAN~\cite{mao2017least}. Thus, for $D$, the objective function that is minimized during training becomes:
\begin{equation}\label{eq:D_objective_fun_LSGAN}
\min_{\substack{D}} V(D) = \mathbb{E}_{x \sim p(x)}[(D(x) - 1)^2] ~+~ \mathbb{E}_{y \sim p(y)}[(D(G(y)))^2].
\end{equation}
The objective function for $G$ is defined as:
\begin{equation}\label{eq:G_objective_fun_LSGAN}
\min_{\substack{G}} V(G) = \mathbb{E}_{y \sim p(y)}[(D(y) - 1)^2].
\end{equation}
We train $D$ and $G$ simultaneously by minimizing the Eqs.~\ref{eq:D_objective_fun_LSGAN} and \ref{eq:G_objective_fun_LSGAN}. 

\section{Experiments}
\subsection{Methods Used for Comparison}
\textit{U-net}\cite{ronneberger2015u}: We simply train a U-net from the training images and segment the test images without performing any type of domain adaptation techniques.

\textit{CycleGAN}\cite{zhu2017unpaired}: In this methodology, in addition to $G$, there is one more generator $F$. $G$ is used to learn a mapping from training images to test images, whereas $F$ learns a mapping from test images to training images. The methodology requires that the mapping from training images to test images and another consecutive mapping from test images to training images reproduce the original training images. This constraint is enforced by minimizing the L1 norm between $F(G(X))$ and $X$, and between $G(F(Y))$ and $Y$. 

\textit{UNIT}\cite{liu2017unsupervised}: The generator of UNIT has two encoders $E_{x}$ and $E_{y}$ and two decoders $G_{x}$ and $G_{y}$. The encoders are used to embed $X$ and $Y$ to a common space. The fake images are generated by $G_{x}(E_{y}(Y))$ and $G_{y}(F_{x}(X))$. In an ideal transformation, $X$ and $G_{x}(E_{y}(Y))$, and $Y$ and $G_{y}(E_{x}(X))$ have similar statistics.

\textit{MUNIT}\cite{huang2018multimodal}: It decomposes the images from both domains into content and style codes. To generate fake images, the content code of one domain and the style code of another domain are combined. AdaIN~\cite{huang2017arbitrary} is utilized for the combination.

\textit{DRIT}\cite{lee2018diverse}: Methodologically DRIT is almost the same as MUNIT. The only difference is that the content code of one domain and the style code of another domain are combined by concatenating them.

\textit{Histogram matching}\cite{Gonzalez}: For each color channel, we match the histogram of the training images with the histogram of the test images to correct the spectral shift between the images.

\textit{Gray world}\cite{buchsbaum1980spatial}: It is one of the color constancy algorithms~\cite{agarwal2006overview}. This algorithm assumes that the average color of the image should be natural gray, and any deviation from gray is caused by the illuminant. This assumption is used to remove the effect of the illuminant. We use gray world algorithm to standardize the training and the test data.

\textit{AdaptSegNet single}\cite{tsai2018learning}: It aims at training a domain invariant network that performs well in segmenting both training and test images. To do this, the classifier generates predicted maps from the training and the test images, and the discriminator forces the predicted maps for the test images to look like the predictions for the training images. In the original paper, DeepLab v2 \cite{chen2018deeplab} is used as the classifier. However, Atrous Spatial Pyramid Pooling (ASPP) in this network reduces the segmentation performance on satellite images significantly, especially when the image contains objects covering a small area. Hence, we remove ASPP from the network and upsample the final classification layer directly to get the predicted map with exactly the same size as the training images. 

\textit{AdaptSegNet multiple}\cite{tsai2018learning}: In the same paper, in addition to aligning the final predictions for the training and test images, the experimental results with 2 classification layers and 2 discriminators are also presented. We compare our method with this strategy as well. 

To make a fair comparison between ColorMapGAN and CycleGAN, UNIT, MUNIT, DRIT, histogram matching, and gray world, we replace the step 2 in our framework with these algorithms. For gray world, we also modify the step 4. Instead of segmenting the original test image, we segment the test image in which the illuminant effect is removed.

\begin{figure}
\centering
\subfloat[\textit{Bad Ischl}]{\includegraphics[width=0.49\linewidth]{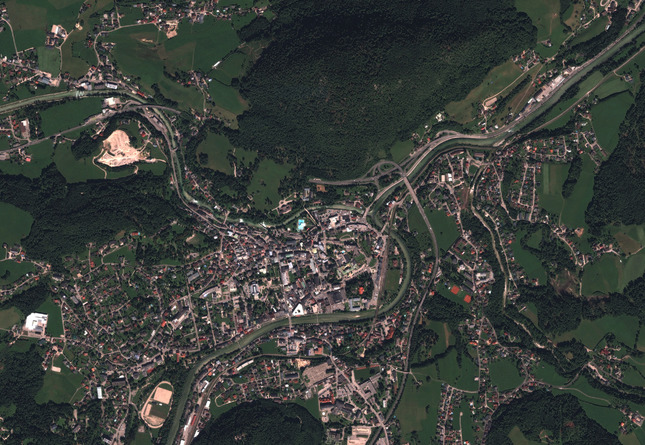}}
\hfill
\subfloat[\textit{Villach}]{\includegraphics[width=0.49\linewidth]{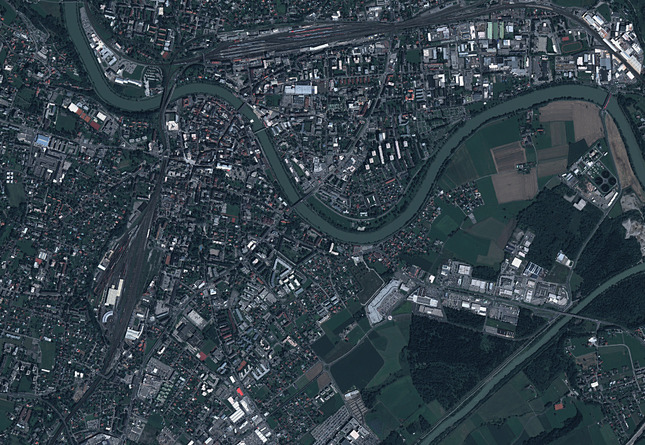}}
\hfill
\subfloat[\textit{B{\'e}ziers}]{\includegraphics[width=0.49\linewidth]{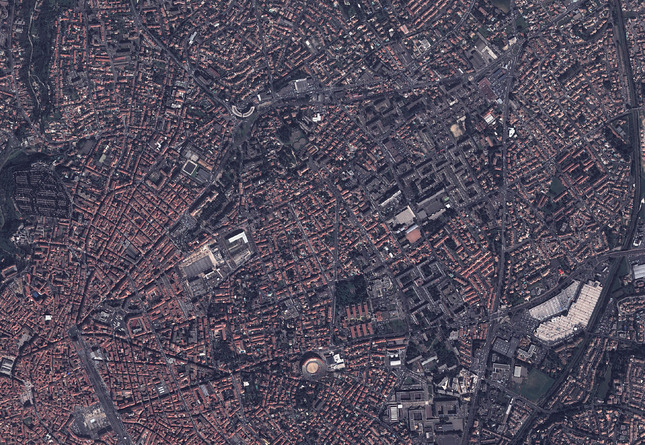}}
\hfill
\subfloat[\textit{Roanne}]{\includegraphics[width=0.49\linewidth]{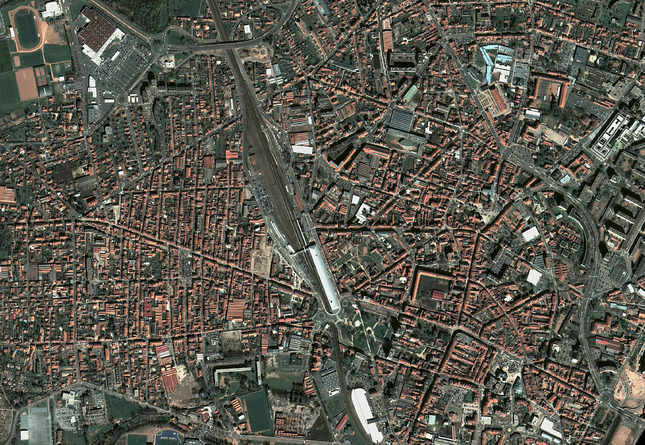}}
\caption{Example close-ups from the Luxcarta data set.}
\label{fig:lux_data_ex_images}
\end{figure}

\begin{table}
\centering
\caption{The statistics for the luxcarta data set.}
\label{table:luxcarta_cities}
\begin{tabular}{c|c|c|c|c|c}
\hline
\multirow{2}{*}{\textbf{City}} & \multirow{2}{*}{\textbf{$\#$ of patches}} & 
                                 \multirow{2}{*}{\textbf{Area (km\textsuperscript{2})}} & 
                                 \multicolumn{3}{c}{\textbf{Class frequency (\%)}} \\
\cline{4-6}
& & & \textbf{building} & \textbf{road} & \textbf{tree} \\
\hline

\textit{Bad Ischl}   & 457 & 27.71 &  5.51 &  6.03 & 35.38 \\
\textit{Villach}     & 749 & 43.59 &  9.26 & 10.63 & 19.91 \\
\textit{B{\'e}ziers} & 407 & 25.75 & 19.09 & 17.62 & 10.91 \\
\textit{Roanne}      & 384 & 25.84 & 18.44 &  8.33 & 14.78 \\
\hline
\end{tabular}
\end{table}

\subsection{Experimental Setup}
To conduct our experiments, we use the Luxcarta data set containing Pl{\'e}iades images acquired over 4 cities in Europe: \textit{Bad Ischl (Austria), Villach (Austria), B{\'e}ziers (France),} and \textit{Roanne (France)}. The images are converted to 8 bit, and their spatial resolution is reduced to 1m by the data set providers. The images in the data set contain red, green, and blue channels. The full annotations for \textit{building}, \textit{road}, and \textit{tree} classes are provided. An example close-up from each city is shown in Fig.~\ref{fig:lux_data_ex_images}. We split the cities into two pairs, where the first pair consists of \textit{Bad Ischl} and \textit{Villach}, and the second pair comprises \textit{B{\'e}ziers} and \textit{Roanne}. To make the experimental setup suitable for the unsupervised domain adaptation problem, when we split the cities into pairs, we pay attention that radiometry of both cities in each pair is as different as possible, and the objects belonging to the same class (e.g., \textit{building}) have similar structural characteristics. For instance, buildings in \textit{B{\'e}ziers} and \textit{Roanne} are densely grouped and have mostly rectangular shape, whereas buildings in \textit{Bad Ischl} and \textit{Villach} are more sparsely distributed and mostly square-like shaped. However, as can be seen in Fig.~\ref{fig:lux_data_ex_images}, there is a large domain shift between spectral bands of the cities in each pair. Our final assumption is that we have access to annotations of only the training city. For instance, when we classify \textit{Roanne} in pair 2, we suppose that its annotations are not available; only the ground-truth of \textit{B{\'e}ziers} is accessible. Similarly, to classify \textit{B{\'e}ziers}, we assume that the ground-truth of only \textit{Roanne} is accessible. 

In the preprocessing step, we split each satellite image into $256 \times 256$ training patches with an overlap of $32$ pixels between neighboring patches. Table~\ref{table:luxcarta_cities} reports for each city the number of patches, the total area covered, and the class frequencies. For the quantitative performance assessment, we use Intersection over Union (IoU)\cite{csurka2013good} as the evaluation metric.  

\subsection{Training Details}\label{sec:training_details}
In the first step of the framework, we train a U-net with Adam optimizer, where the learning rate is 0.0001, exponential decay rate for the first and the second moment estimates are 0.9 and 0.999. In each iteration, we randomly sample a batch of 32 training patches. We apply online data augmentation with random horizontal/vertical flips and 0/90/180/270 degrees of rotations. We use sigmoid cross entropy as the loss function and ignore the background class while computing the loss. We optimize the network for 2,500 iterations. 

In the stage of generating fake training images, in each training iteration of ColorMapGAN, we randomly sample only one patch from the training city and one patch from the test city. We use Adam optimizer to update both the generator and the discriminator. Since the generator is architecturally much simpler than the discriminator, we prefer to optimize it with a larger learning rate. For the generator, the learning rate is 0.0005, whereas we set it to 0.0001 for the discriminator. We train ColorMapGAN for 8,000 iterations, since we verify by visual inspection that visually appealing results are obtained for this number of iterations. We fine-tune the previously trained network on the fake training images for 750 iterations. For all the compared methods, we use default parameters that are specified in the related papers.

%BAD ISCHL
\begin{figure*}
\centering
\subfloat[Bad Ischl]{\includegraphics[width = 0.24\linewidth]{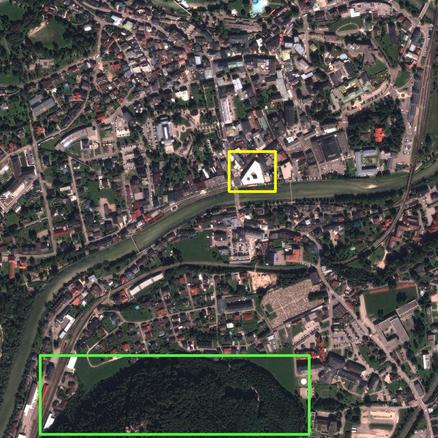}}
\hfill
\subfloat[CycleGAN\cite{zhu2017unpaired}\label{fig:bad_ischl_cycle_gan}]{\includegraphics[width = 0.24\linewidth]{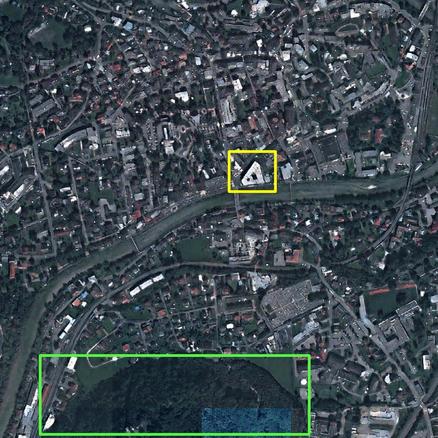}}
\hfill
\subfloat[UNIT\cite{liu2017unsupervised}]{\includegraphics[width = 0.24\linewidth]{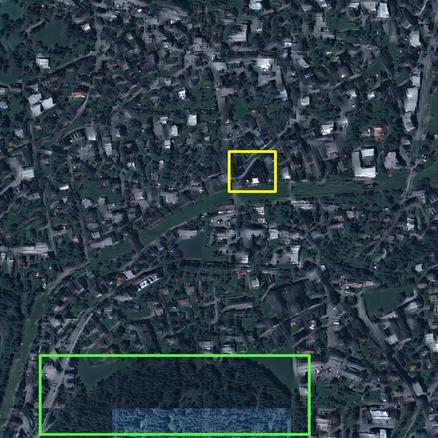}}
\hfill
\subfloat[MUNIT\cite{huang2018multimodal}]{\includegraphics[width = 0.24\linewidth]{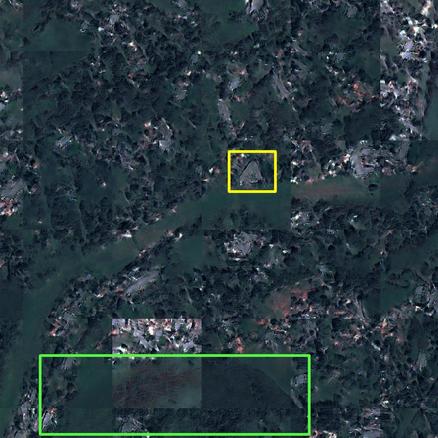}}
\hfill
\subfloat[DRIT\cite{lee2018diverse}]{\includegraphics[width = 0.24\linewidth]{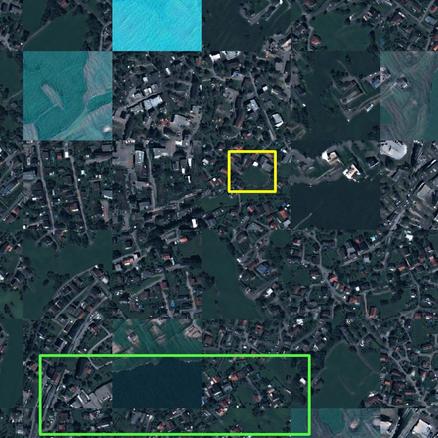}}
\hfill
\subfloat[Gray world\cite{buchsbaum1980spatial}]{\includegraphics[width = 0.24\linewidth]{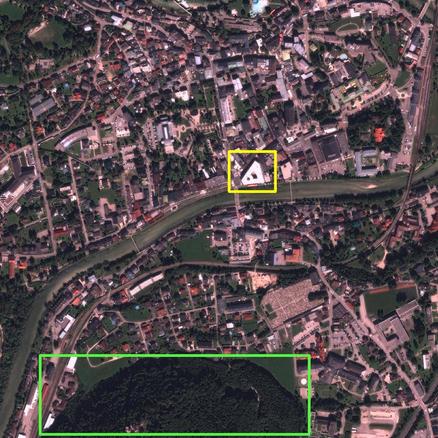}}
\hfill
\subfloat[Histogram matching\cite{Gonzalez}]{\includegraphics[width = 0.24\linewidth]{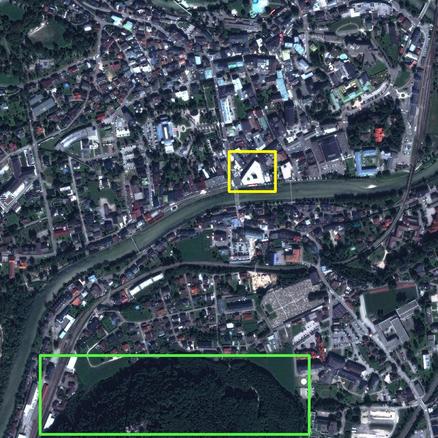}}
\hfill
\subfloat[ColorMapGAN (ours)]{\includegraphics[width = 0.24\linewidth]{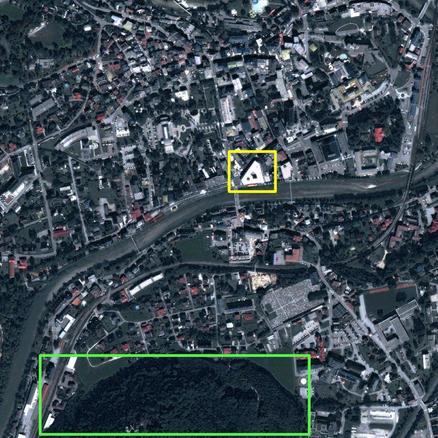}}
\caption{Original \textit{Bad Ischl} and the generated fake images that are used to generate maps for \textit{Villach}.}
\label{fig:bad_ischl_original_fake}
\end{figure*}

%VILLACH
\begin{figure*}
\centering
\subfloat[Villach\label{fig:villach_original}]{\includegraphics[width = 0.24\linewidth]{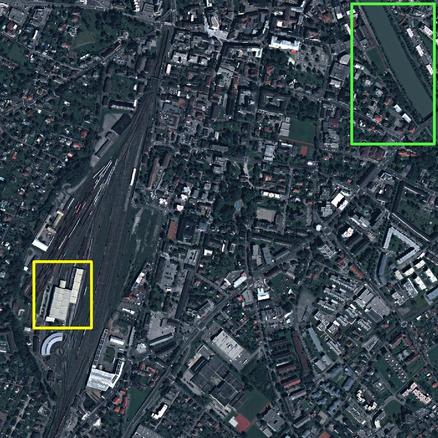}}
\hfill
\subfloat[CycleGAN\cite{zhu2017unpaired}]{\includegraphics[width = 0.24\linewidth]{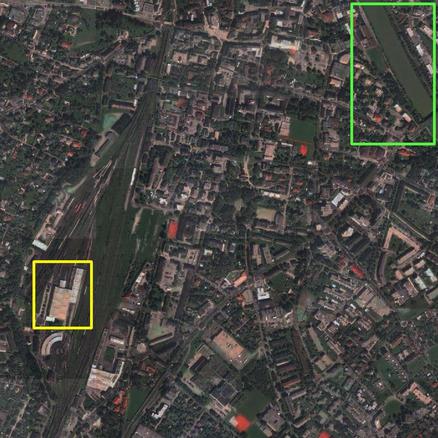}}
\hfill
\subfloat[UNIT\cite{liu2017unsupervised}]{\includegraphics[width = 0.24\linewidth]{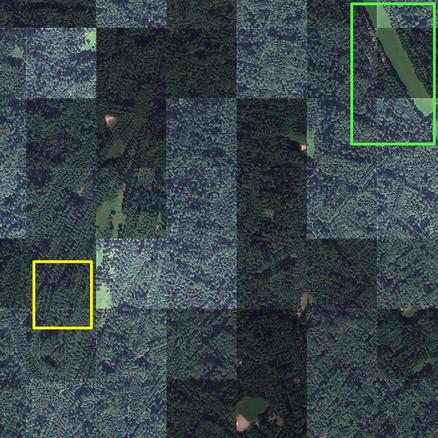}}
\hfill
\subfloat[MUNIT\cite{huang2018multimodal}]{\includegraphics[width = 0.24\linewidth]{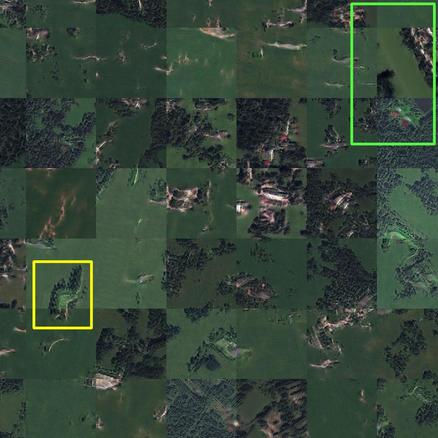}}
\hfill
\subfloat[DRIT\cite{lee2018diverse}]{\includegraphics[width = 0.24\linewidth]{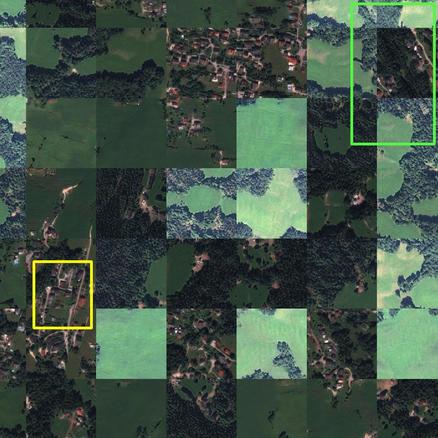}}
\hfill
\subfloat[Gray world\cite{buchsbaum1980spatial}]{\includegraphics[width = 0.24\linewidth]{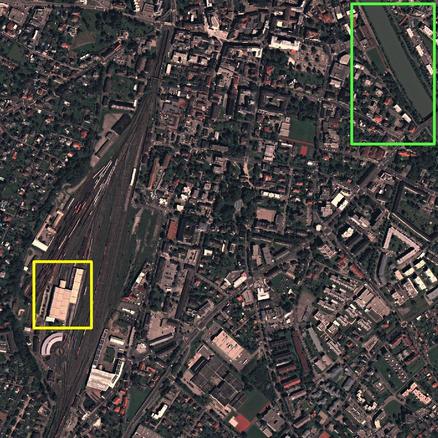}}
\hfill
\subfloat[Histogram matching\cite{Gonzalez}]{\includegraphics[width = 0.24\linewidth]{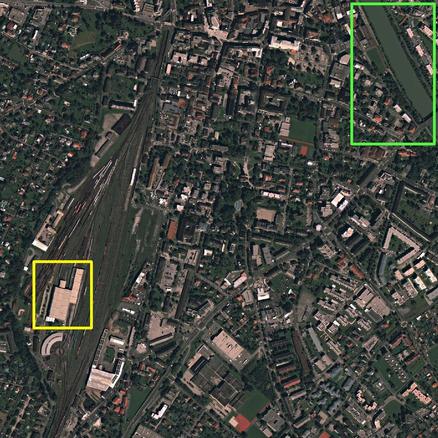}}
\hfill
\subfloat[ColorMapGAN (ours)]{\includegraphics[width = 0.24\linewidth]{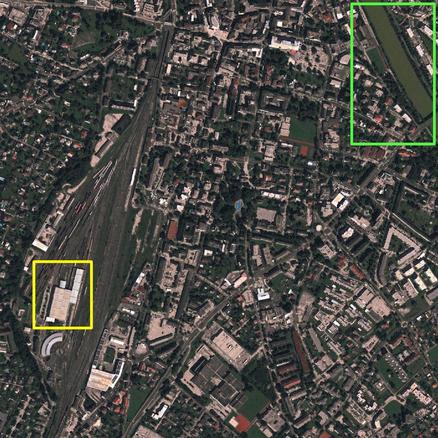}}
\caption{Original \textit{Villach} and the generated fake images that are used to generate maps for \textit{Bad Ischl}.}
\label{fig:villach_original_fake}
\end{figure*}

%BEZIERS
\begin{figure*}
\centering
\subfloat[B{\'e}ziers]{\includegraphics[width = 0.24\linewidth]{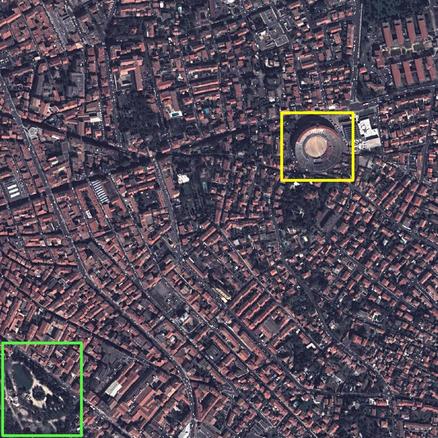}}
\hfill
\subfloat[CycleGAN\cite{zhu2017unpaired}]{\includegraphics[width = 0.24\linewidth]{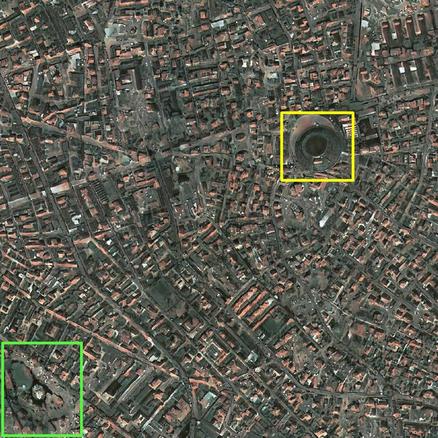}}
\hfill
\subfloat[UNIT\cite{liu2017unsupervised}]{\includegraphics[width = 0.24\linewidth]{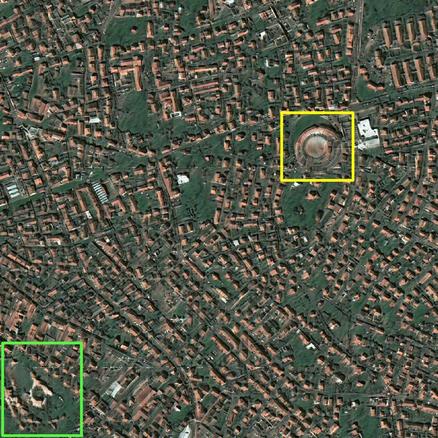}}
\hfill
\subfloat[MUNIT\cite{huang2018multimodal}]{\includegraphics[width = 0.24\linewidth]{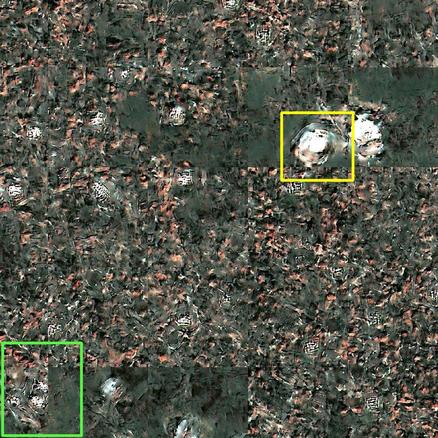}}
\hfill
\subfloat[DRIT\cite{lee2018diverse}]{\includegraphics[width = 0.24\linewidth]{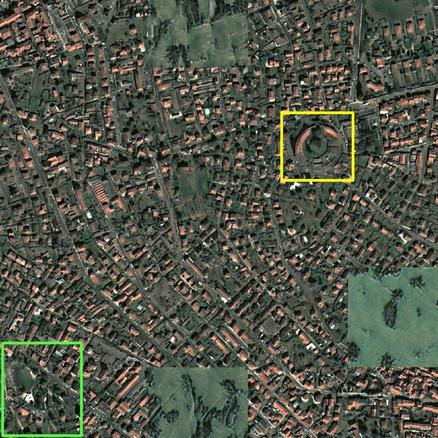}}
\hfill
\subfloat[Gray world\cite{buchsbaum1980spatial}]{\includegraphics[width = 0.24\linewidth]{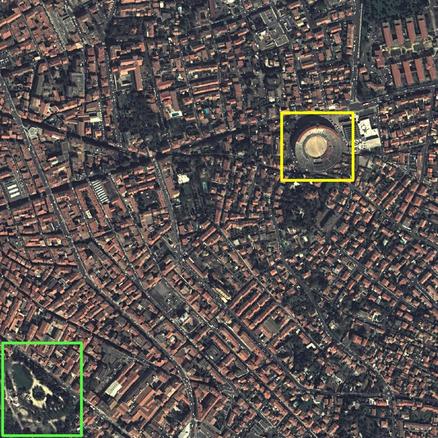}}
\hfill
\subfloat[Histogram matching\cite{Gonzalez}]{\includegraphics[width = 0.24\linewidth]{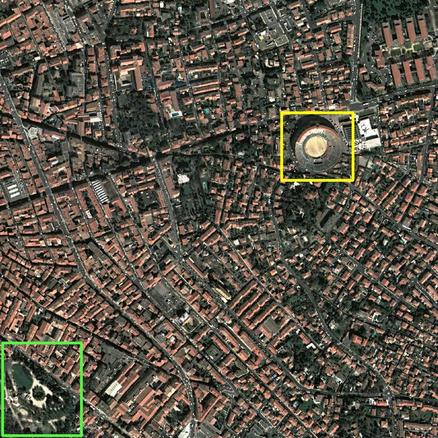}}
\hfill
\subfloat[ColorMapGAN (ours)]{\includegraphics[width = 0.24\linewidth]{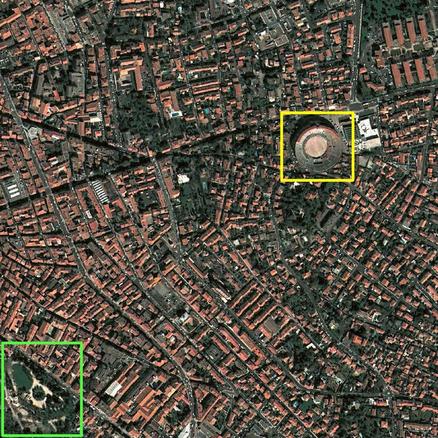}}
\caption{Original \textit{B{\'e}ziers} and the generated fake images that are used to generate maps for \textit{Roanne}.}
\label{fig:beziers_original_fake}
\end{figure*}

%ROANNE
\begin{figure*}
\centering
\subfloat[Roanne]{\includegraphics[width = 0.24\linewidth]{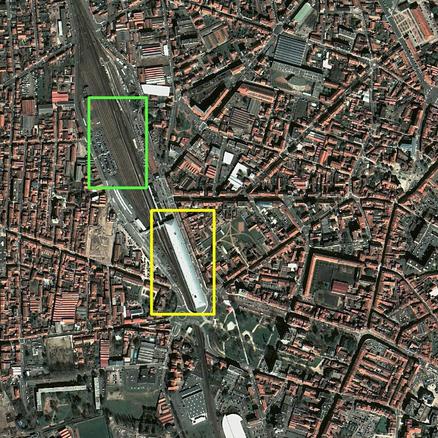}}
\hfill
\subfloat[CycleGAN\cite{zhu2017unpaired}]{\includegraphics[width = 0.24\linewidth]{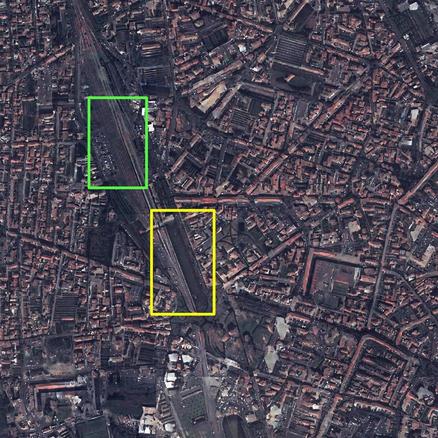}}
\hfill
\subfloat[UNIT\cite{liu2017unsupervised}]{\includegraphics[width = 0.24\linewidth]{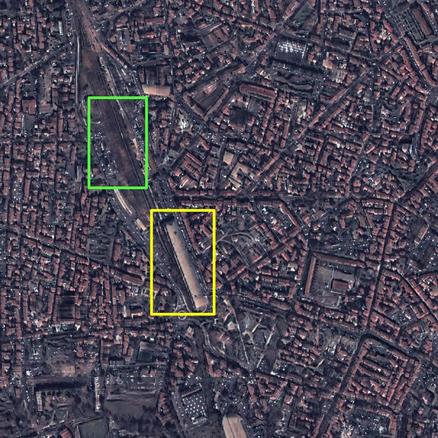}}
\hfill
\subfloat[MUNIT\cite{huang2018multimodal}]{\includegraphics[width = 0.24\linewidth]{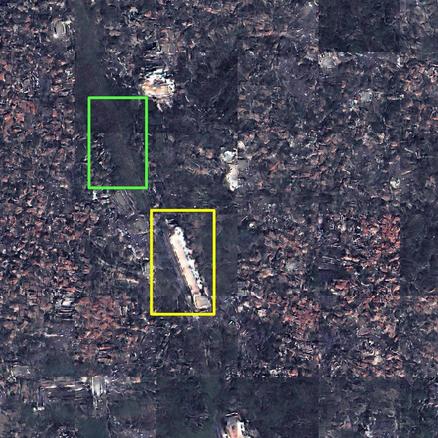}}
\hfill
\subfloat[DRIT\cite{lee2018diverse}]{\includegraphics[width = 0.24\linewidth]{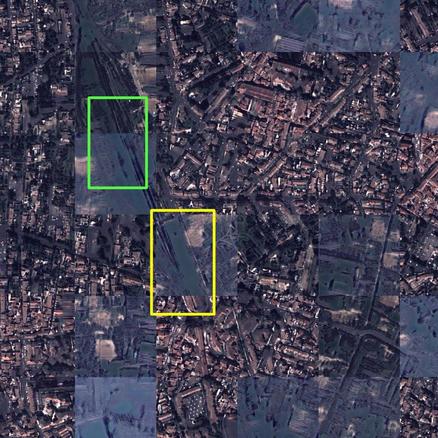}}
\hfill
\subfloat[Gray world\cite{buchsbaum1980spatial}]{\includegraphics[width = 0.24\linewidth]{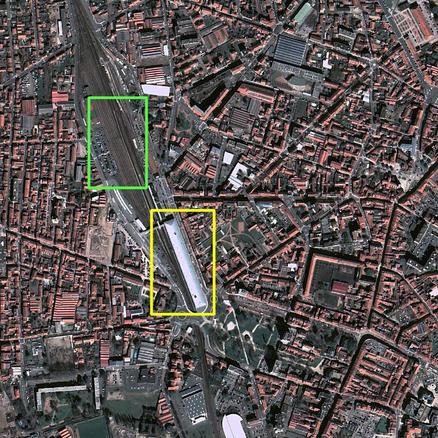}}
\hfill
\subfloat[Histogram matching\cite{Gonzalez}]{\includegraphics[width = 0.24\linewidth]{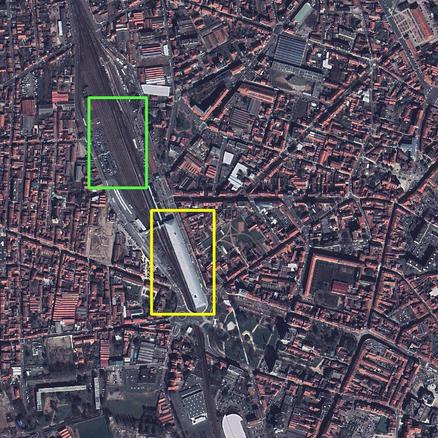}}
\hfill
\subfloat[ColorMapGAN (ours)]{\includegraphics[width = 0.24\linewidth]{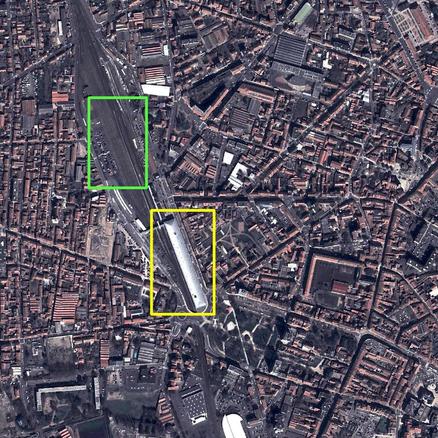}}
\caption{Original \textit{Roanne} and the generated fake images that are used to generate maps for \textit{B{\'e}ziers}.}
\label{fig:roanne_original_fake}
\end{figure*}
\begin{table*}
\centering
\caption{IoU scores for the test cities in pair 1.}
\label{table:ious_pair1}
\scalebox{0.8}{
\begin{tabular}{||p{0.01cm}p{0.1cm}|c||c|c|c|c||c|c|c|c||}
\hline			
\multicolumn{3}{||c||}{\multirow{2}{*}{\textbf{Method}}} & \multicolumn{4}{c||}{\textbf{Training: Bad Ischl, Test: Villach}} & \multicolumn{4}{c||}{\textbf{Training: Villach, Test: Bad Ischl}} \\
\cline{4-11}                         
& \multicolumn{2}{c||}{} & \textbf{building} & \textbf{road} & \textbf{tree} & \textbf{Overall} 
& \textbf{building} & \textbf{road} & \textbf{tree} & \textbf{Overall} \\ 
\hline
\multicolumn{3}{||c||}{U-net}                       & 23.61 &  0.91 & 40.53 & 21.68 & 5.84 &  0.24 &  0.50 &  2.19  \\
\hline
\multicolumn{3}{||c||}{AdaptSegNet Single\cite{tsai2018learning}} &  6.01 &  4.37 & 10.43 &  6.94 &  3.06 &  2.71 & 10.23 &  5.33 \\
\multicolumn{3}{||c||}{AdaptSegNet Multi\cite{tsai2018learning}}  & 24.59 &  9.02 & 56.08 & 29.86 & 14.26 &  4.46 & 24.66 & 14.46 \\
\hline
\multirow{7}{*}{\rotatebox{90}{Our proposed}}& \multirow{7}{*}{\rotatebox{90}{framework with}}& CycleGAN\cite{zhu2017unpaired}
                                                    & 43.03 & 28.96 & \textbf{68.86} & 46.95 & 43.62 & 38.69 & \textbf{71.68} & \textbf{51.33} \\
& & UNIT\cite{liu2017unsupervised}                  & 30.86 & 15.84 & 63.00 & 36.57 & 19.29 & 36.83 & 35.57 & 30.56 \\
& & MUNIT\cite{huang2018multimodal}                 &  0.02 &  1.38 & 47.23 & 16.21 &  6.20 &  0.13 &  0.05 &  2.13 \\
& & DRIT\cite{lee2018diverse}                       &  0.01 &  3.96 &  8.72 &  4.23 &  0.00 & 10.19 &  0.01 &  3.40 \\
& & Gray world\cite{buchsbaum1980spatial}           & 25.19 & 26.43 & 56.15 & 35.92 & 29.55 & 24.80 & 46.41 & 33.58 \\
& & Histogram matching\cite{Gonzalez}               & 24.95 & 29.34 & 61.59 & 38.63 &  6.45 &  0.92 &  1.28 &  2.88 \\
& & ColorMapGAN (ours)                              & \textbf{48.47} & \textbf{37.82} & 58.92 & \textbf{48.40} & \textbf{49.16} & \textbf{41.75} & 59.84 & 50.25 \\
\hline  
\end{tabular}}
\end{table*}

\begin{table*}
\centering
\caption{IoU scores for the test cities in pair 2.}
\label{table:ious_pair2}
\scalebox{0.8}{
\begin{tabular}{||p{0.01cm}p{0.1cm}|c||c|c|c|c||c|c|c|c||}
\hline			
\multicolumn{3}{||c||}{\multirow{2}{*}{\textbf{Method}}} & \multicolumn{4}{c||}{\textbf{Training: B{\'e}ziers, Test: Roanne}} & \multicolumn{4}{c||}{\textbf{Training: Roanne, Test: B{\'e}ziers}} \\
\cline{4-11}                         
& \multicolumn{2}{c||}{} & \textbf{building} & \textbf{road} & \textbf{tree} & \textbf{Overall} 
& \textbf{building} & \textbf{road} & \textbf{tree} & \textbf{Overall} \\ 
\hline
\multicolumn{3}{||c||}{U-net}                       & 26.13 & 11.16 &  7.79 & 15.03 & 19.85 &  0.00 &  0.00 &  6.62  \\
\hline
\multicolumn{3}{||c||}{AdaptSegNet Single\cite{tsai2018learning}} &  6.61 & 11.05 &  3.37 &  7.01 & 11.07 &  4.19 &  3.71 &  6.32 \\
\multicolumn{3}{||c||}{AdaptSegNet Multi\cite{tsai2018learning}}  & 22.42 &  5.87 & 17.84 & 15.37 & 24.27 & 10.88 & 10.45 & 15.20 \\
\hline
\multirow{7}{*}{\rotatebox{90}{Our proposed}}& \multirow{7}{*}{\rotatebox{90}{framework with}}& CycleGAN\cite{zhu2017unpaired}
                                                    & 18.19 & 24.28 &  0.19 & 14.22 &  9.92 & 16.00 &  0.54 &  8.82 \\
& & UNIT\cite{liu2017unsupervised}                  & 41.99 & 38.47 &  0.39 & 26.95 & 29.99 & \textbf{39.19} &  3.11 & 24.10 \\
& & MUNIT\cite{huang2018multimodal}                 & 10.17 &  1.66 &  0.31 &  4.05 &  7.49 &  0.84 &  0.13 &  2.82 \\
& & DRIT\cite{lee2018diverse}                       & 42.16 & 41.77 &  1.18 & 28.37 & 25.73 & 36.54 &  0.68 & 20.98 \\
& & Gray world\cite{buchsbaum1980spatial}           & 51.47 & 40.42 & 18.25 & 36.71 & 14.61 & 31.32 & \textbf{21.99} & 22.64 \\
& & Histogram matching\cite{Gonzalez}               & 18.64 &  9.87 &  4.81 & 11.11 & 20.63 &  0.02 &  0.00 &  6.88 \\
& & ColorMapGAN (ours)                              & \textbf{55.60} & \textbf{44.66} & \textbf{28.39} & \textbf{42.88} & \textbf{47.12} & 35.18 & 21.91 & \textbf{34.74} \\
\hline  
\end{tabular}}
\end{table*}

\subsection{Results}
\begin{figure*}
\subfloat[\textit{Bad Ischl}]{\includegraphics[width = 0.165\linewidth]{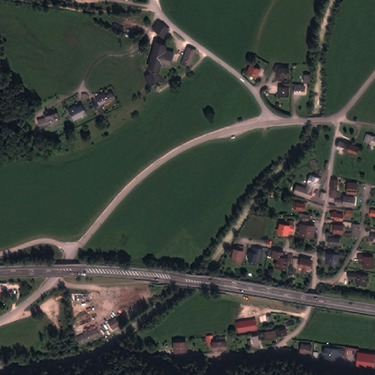}}
\hfill
\subfloat[Ground-truth]{\includegraphics[width = 0.165\linewidth]{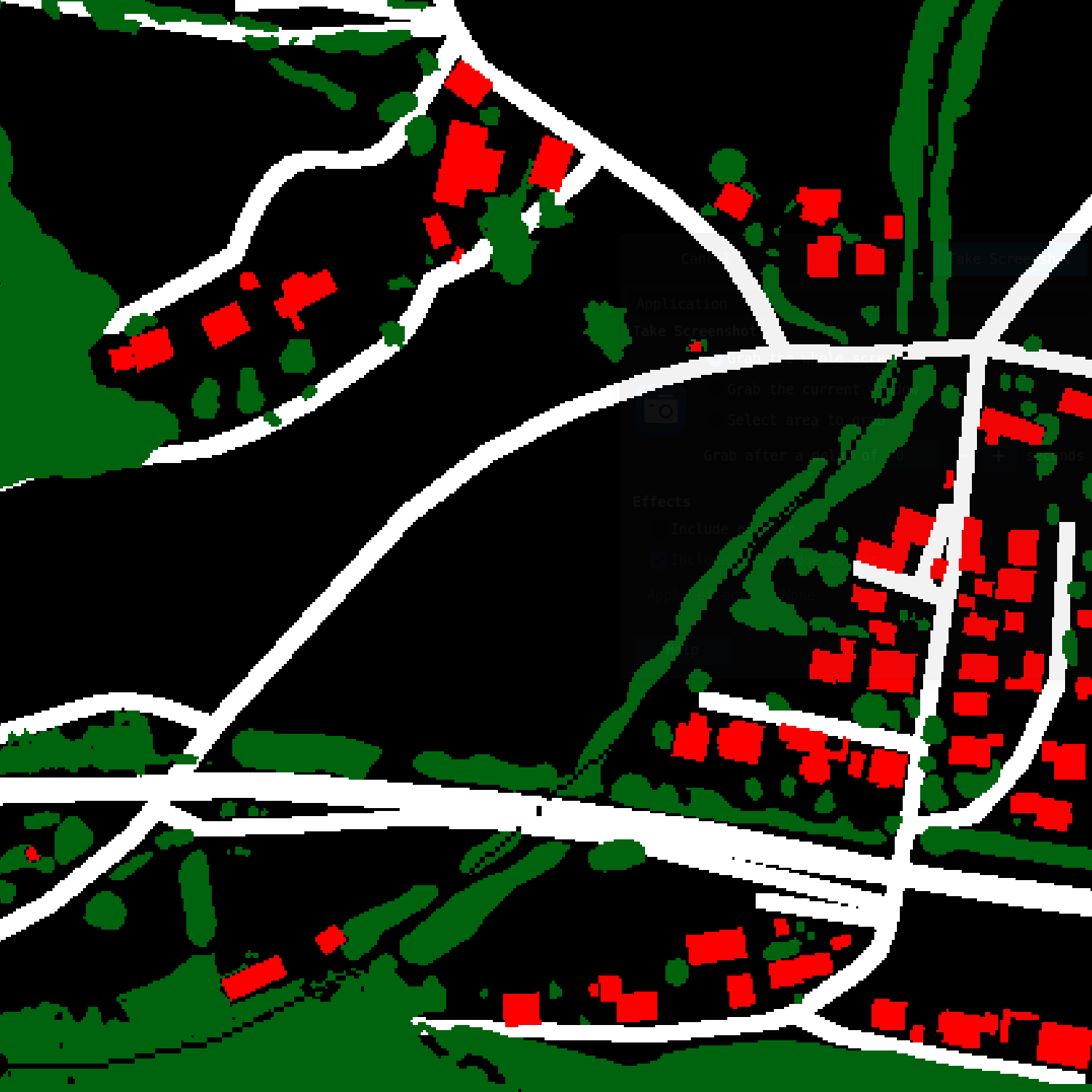}}
\hfill
\subfloat[U-net~\cite{ronneberger2015u}]{\includegraphics[width = 0.165\linewidth]{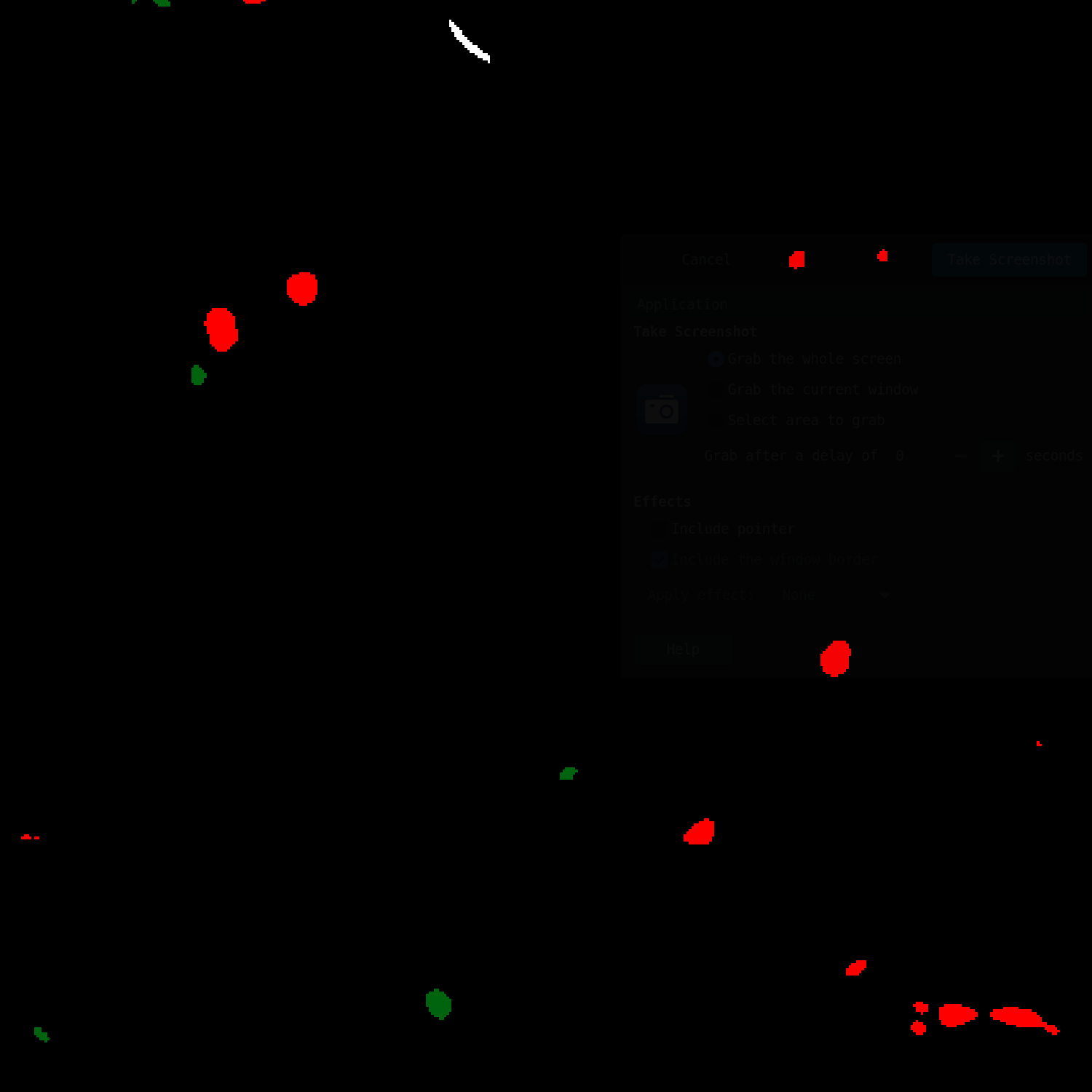}}
\hfill
\subfloat[AdaptSN S~\cite{tsai2018learning}]{\includegraphics[width = 0.165\linewidth]{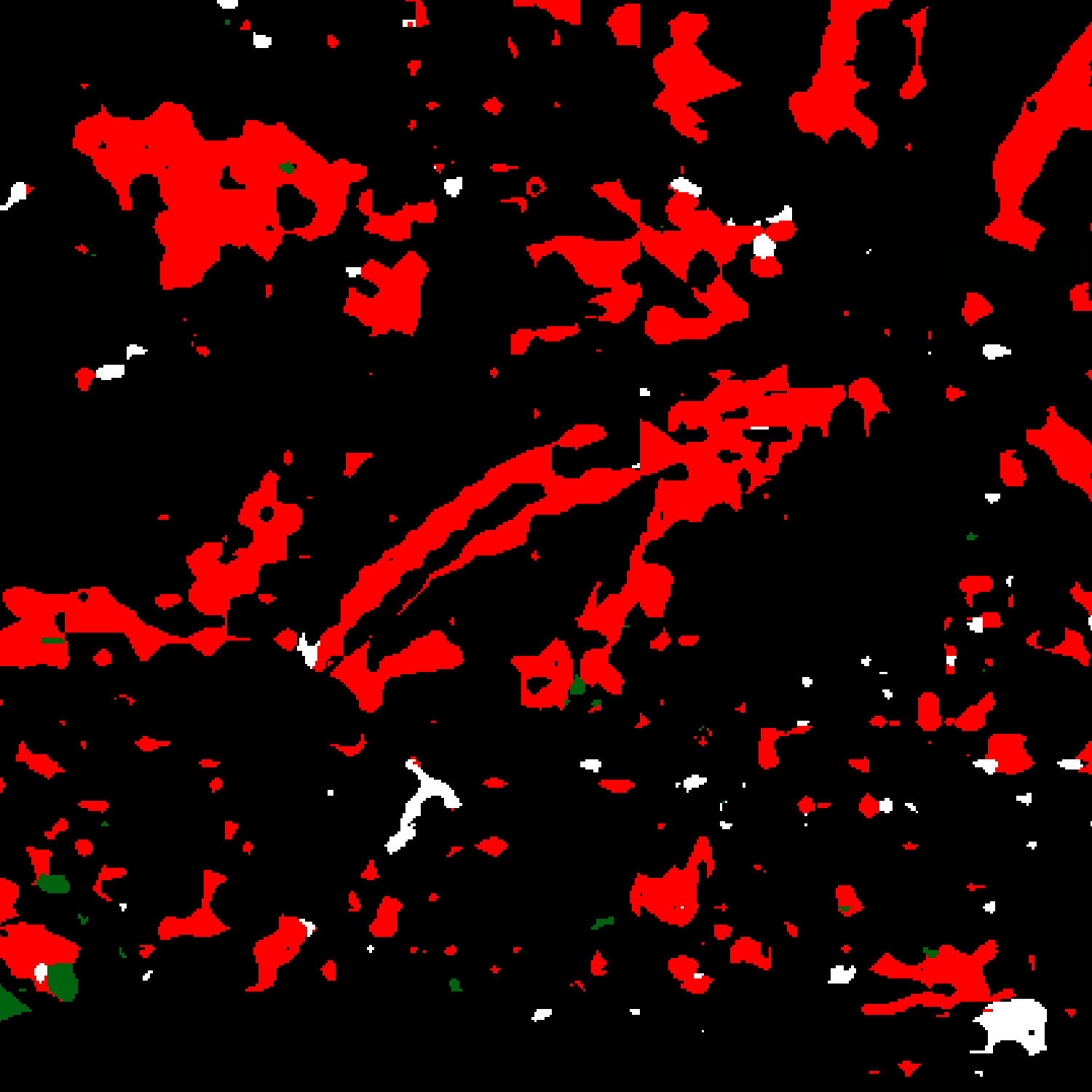}}
\hfill
\subfloat[AdaptSN M~\cite{tsai2018learning}]{\includegraphics[width = 0.165\linewidth]{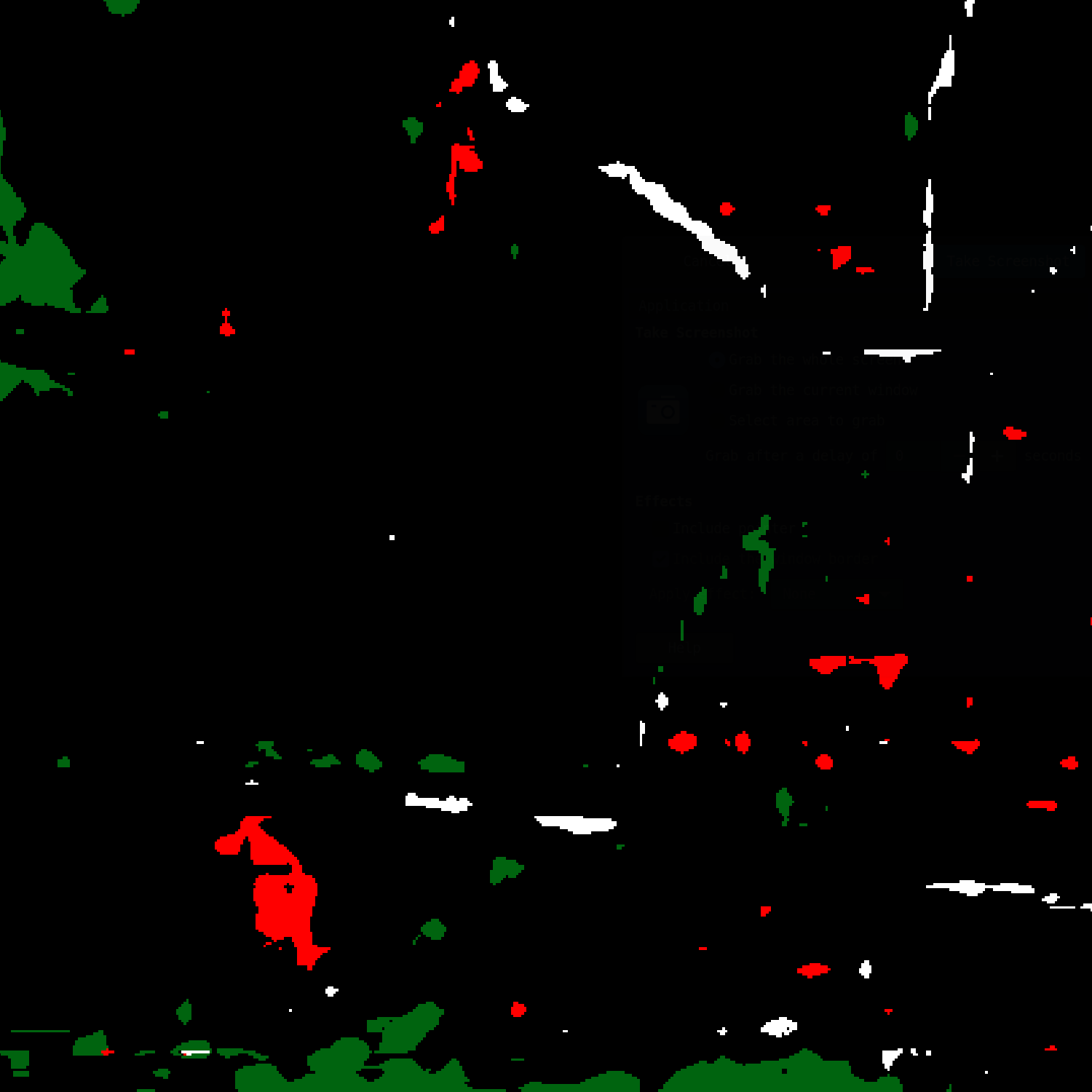}}
\hfill
\subfloat[CycleGAN~\cite{zhu2017unpaired}]{\includegraphics[width = 0.165\linewidth]{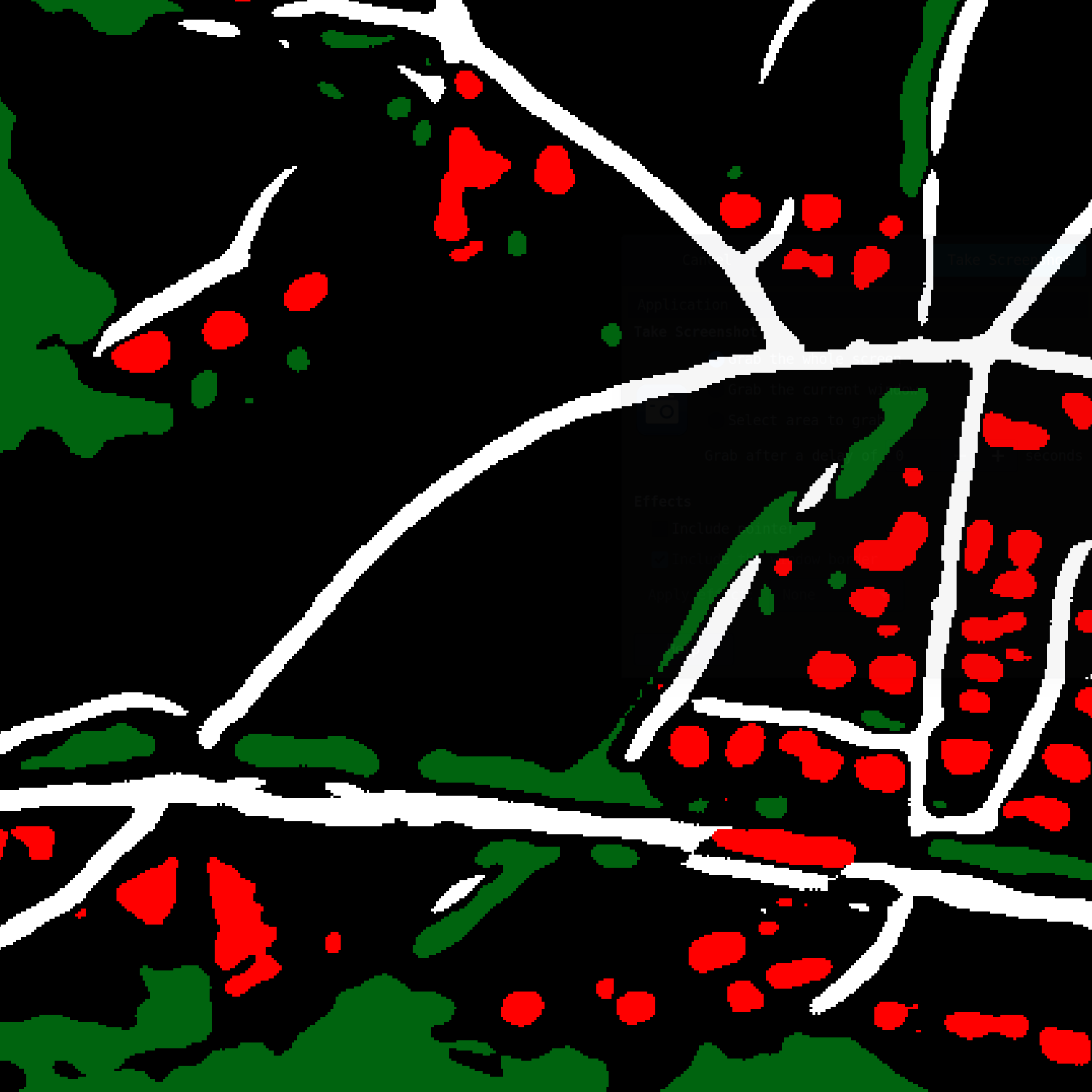}}

\subfloat[UNIT~\cite{liu2017unsupervised}]{\includegraphics[width = 0.165\linewidth]{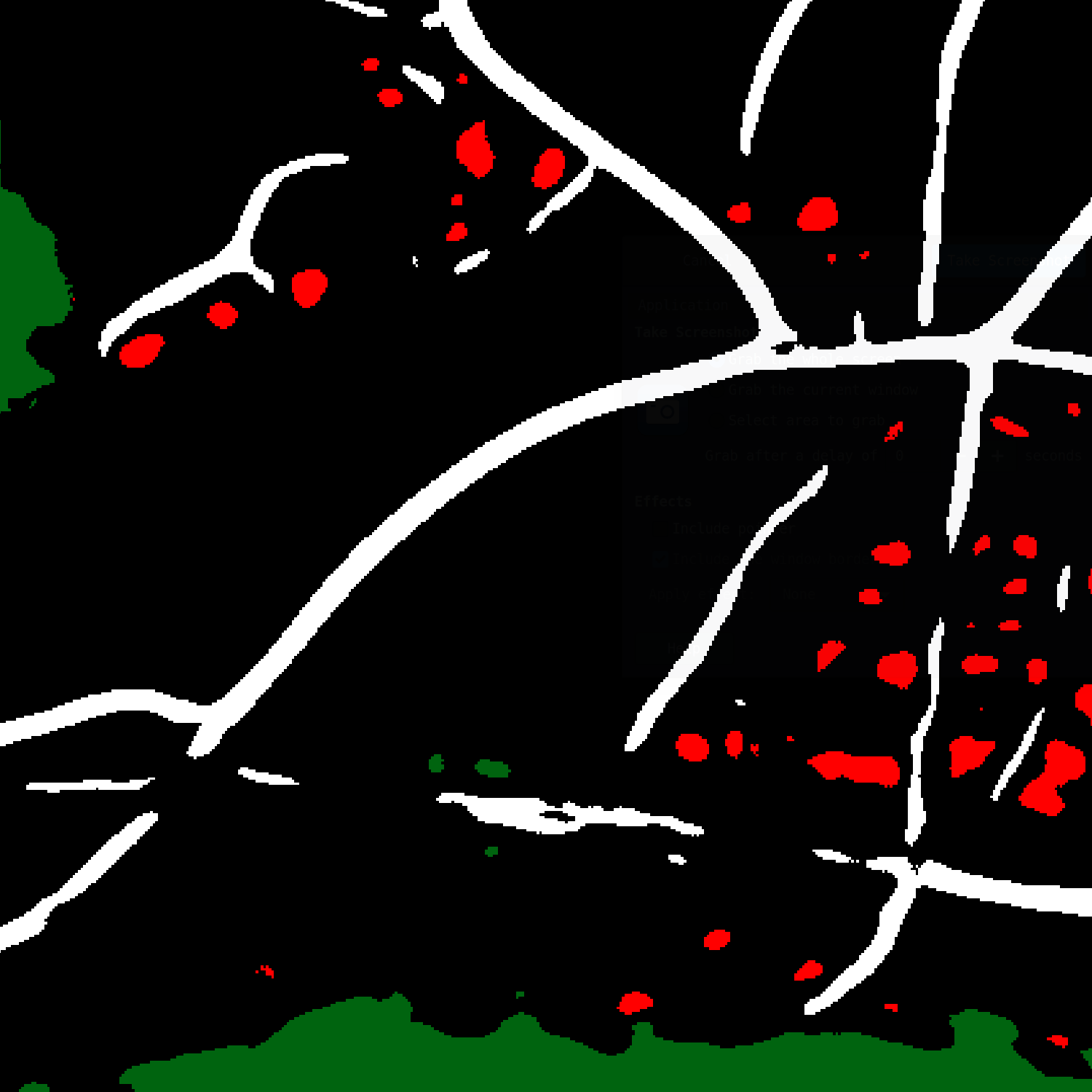}}
\hfill
\subfloat[MUNIT~\cite{huang2018multimodal}]{\includegraphics[width = 0.165\linewidth]{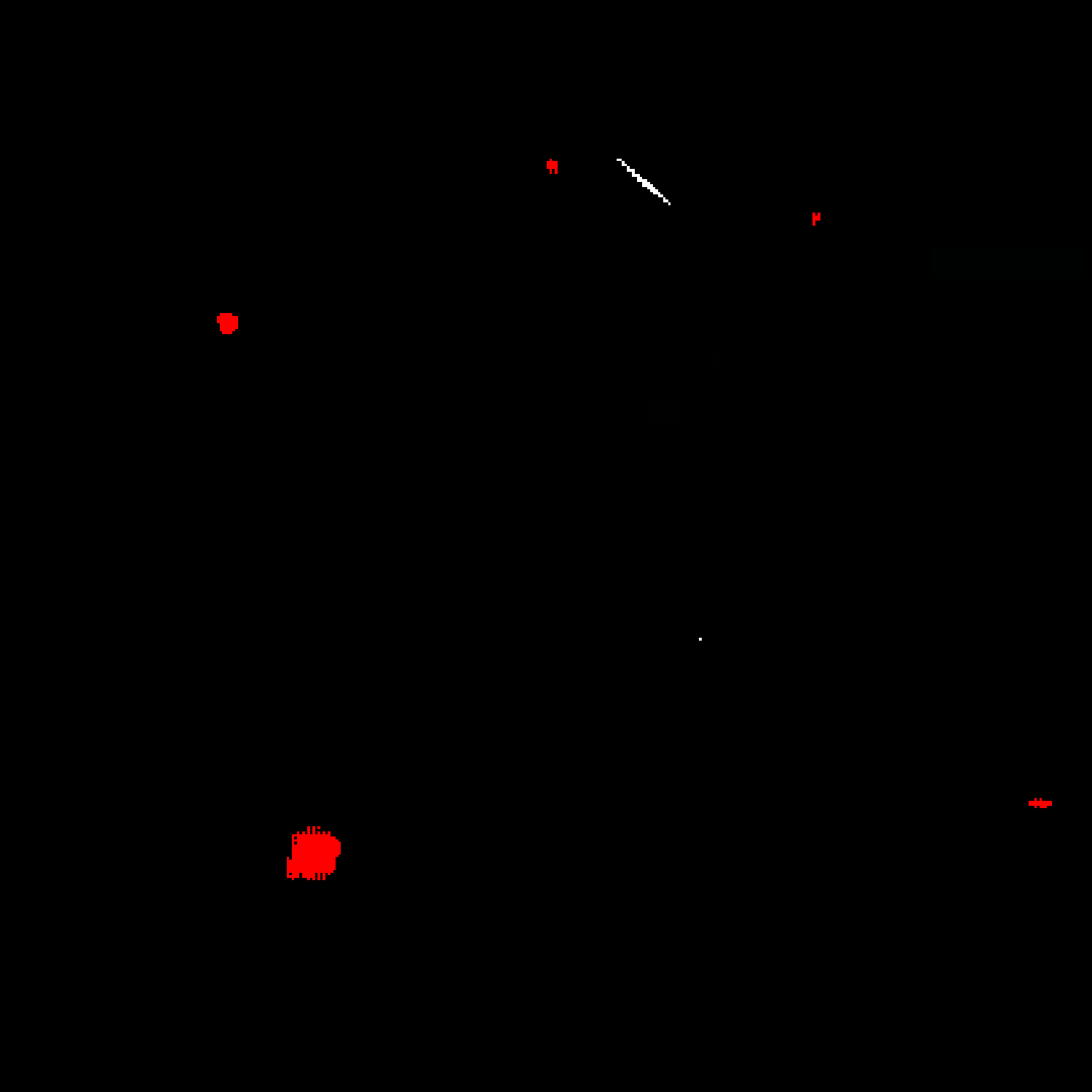}}
\hfill
\subfloat[DRIT~\cite{lee2018diverse} ]{\includegraphics[width = 0.165\linewidth]{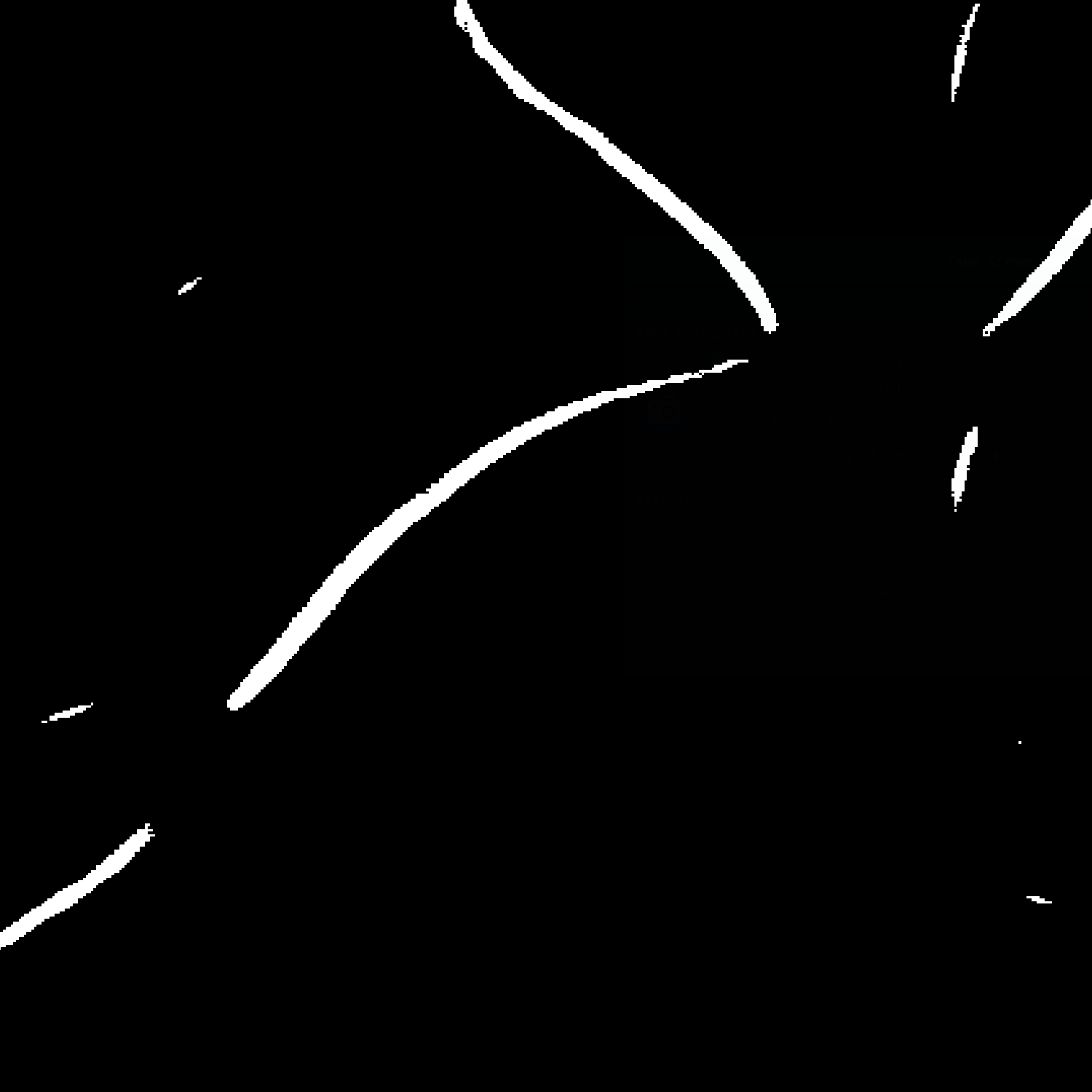}}
\hfill
\subfloat[Gray world~\cite{buchsbaum1980spatial}]{\includegraphics[width = 0.165\linewidth]{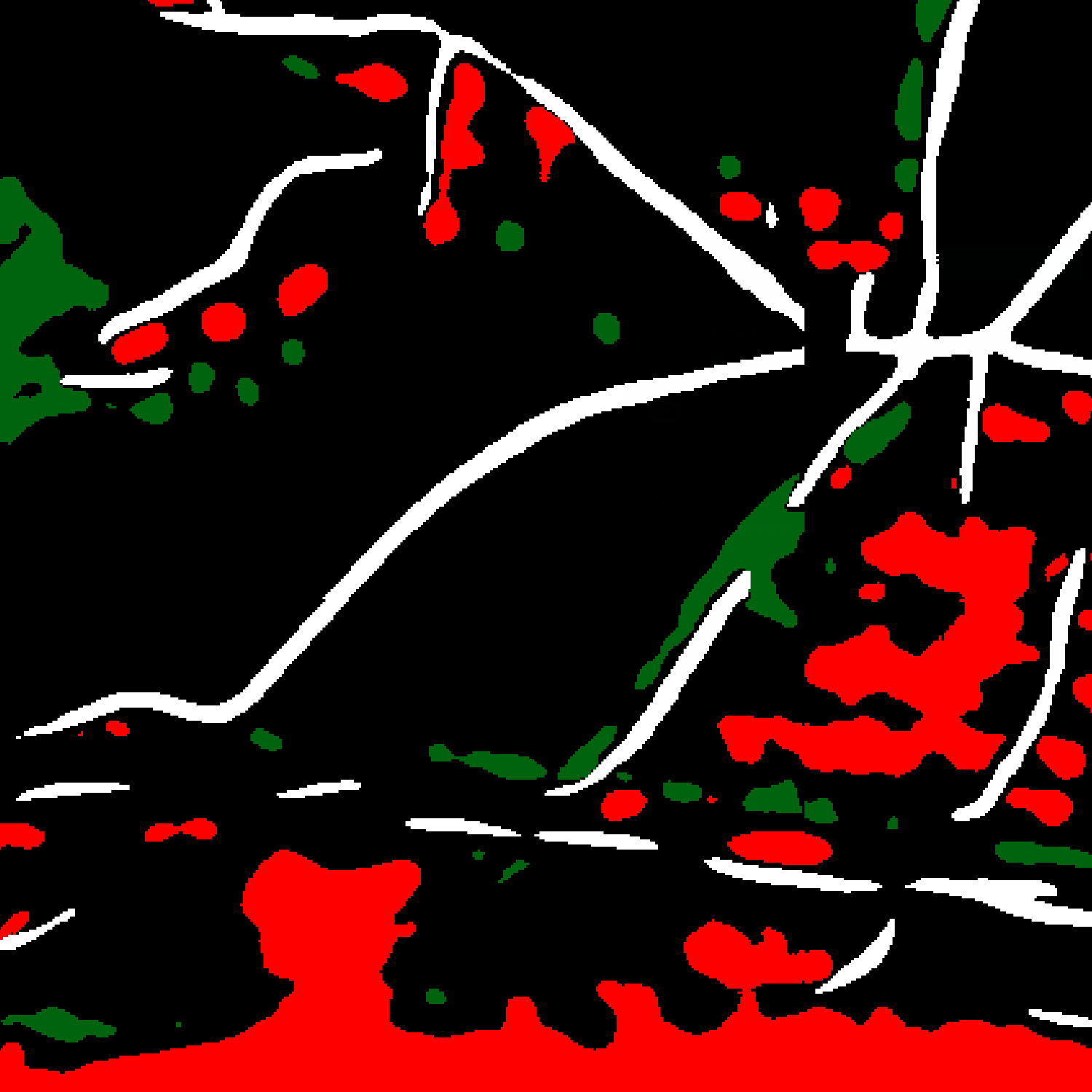}}
\hfill
\subfloat[Hist. match.~\cite{Gonzalez}]{\includegraphics[width = 0.165\linewidth]{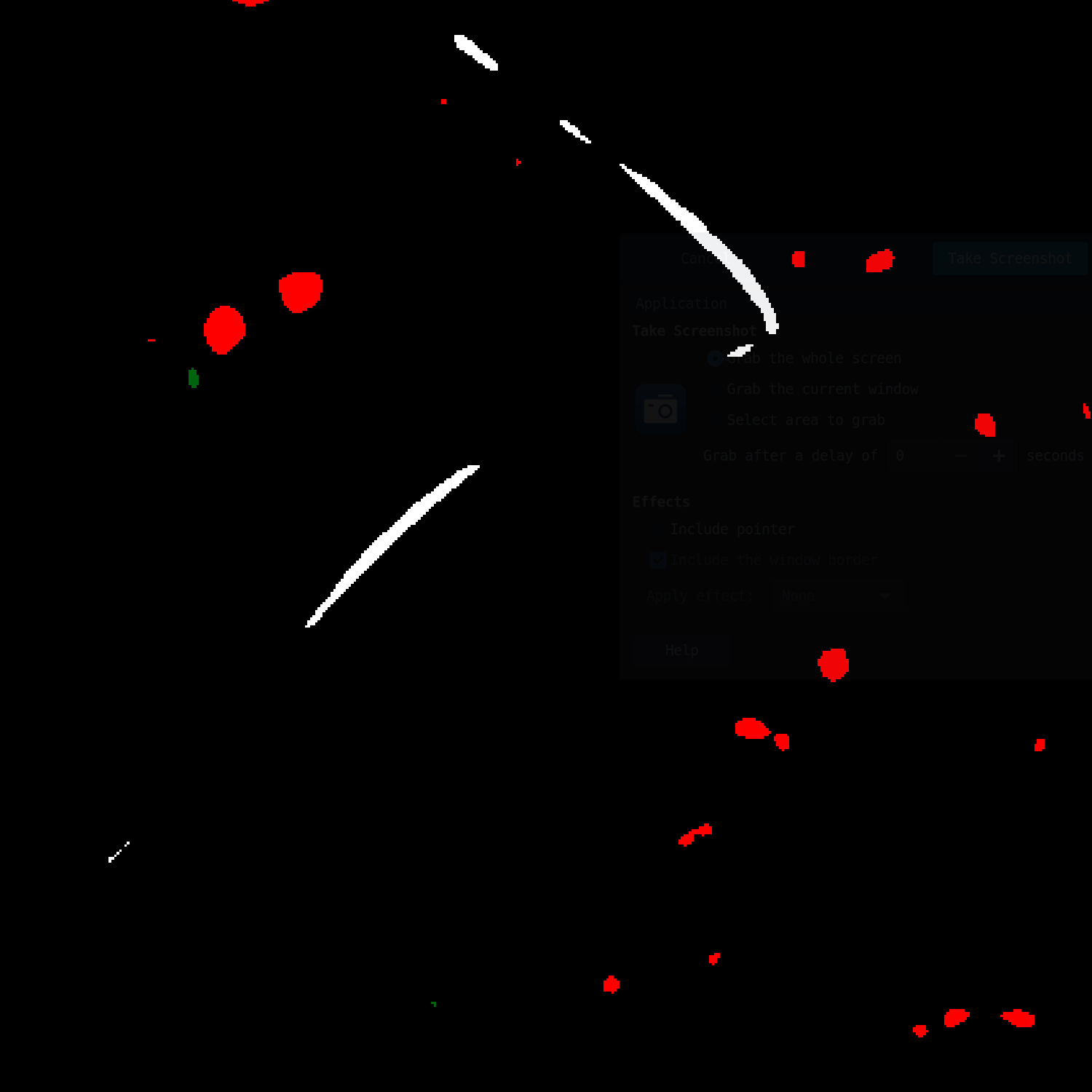}}
\hfill
\subfloat[ColorMapGAN]{\includegraphics[width = 0.165\linewidth]{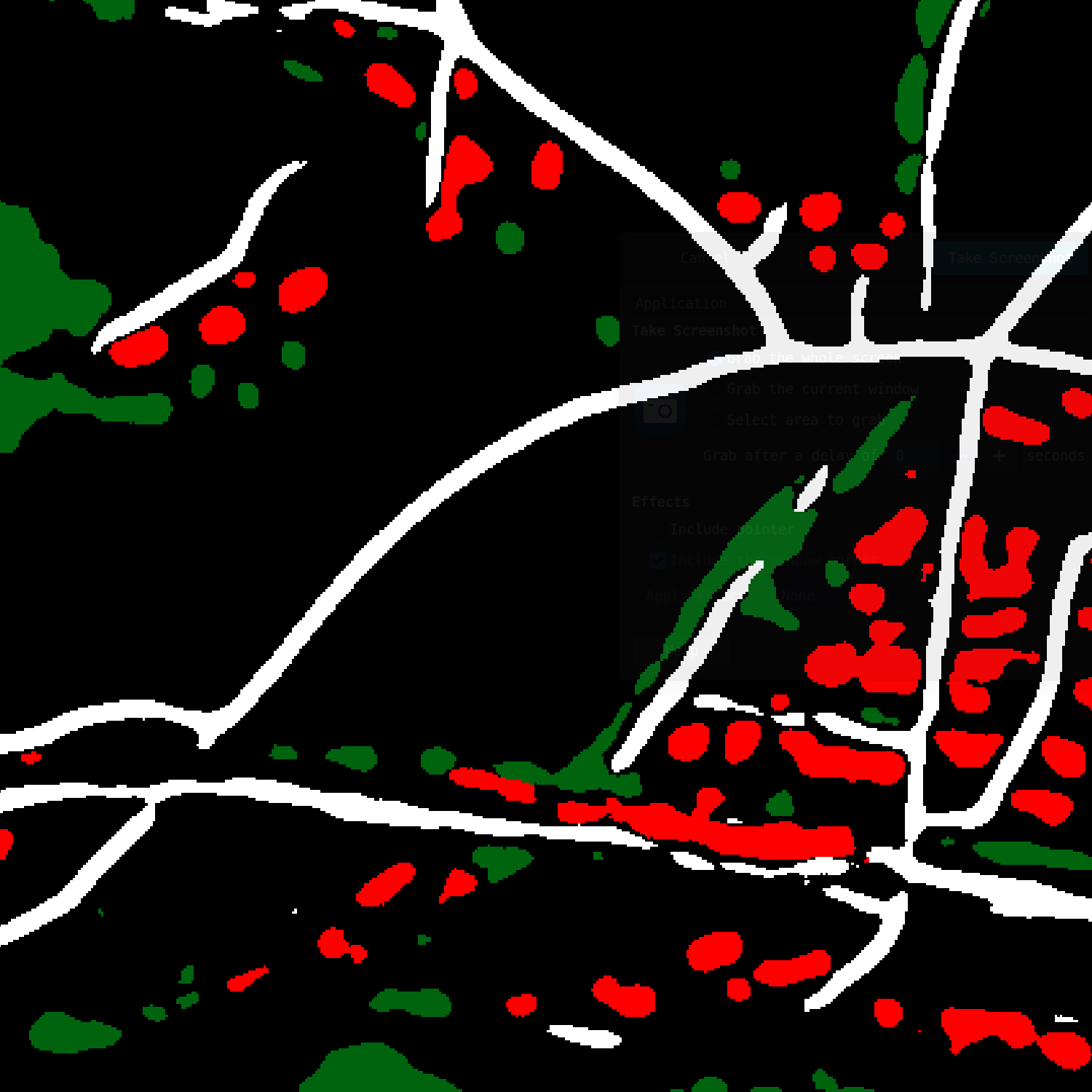}}
\caption{\textit{Bad Ischl}, ground-truth, and the predictions. \textit{Building, road, tree}, and background classes are represented by red, white, green, and black colors.}
\label{fig:bad_ischl_preds}
\end{figure*}

\begin{figure*}
\subfloat[\textit{Villach}]{\includegraphics[width = 0.165\linewidth]{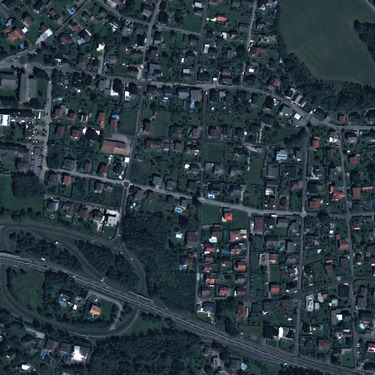}}
\hfill
\subfloat[Ground-truth]{\includegraphics[width = 0.165\linewidth]{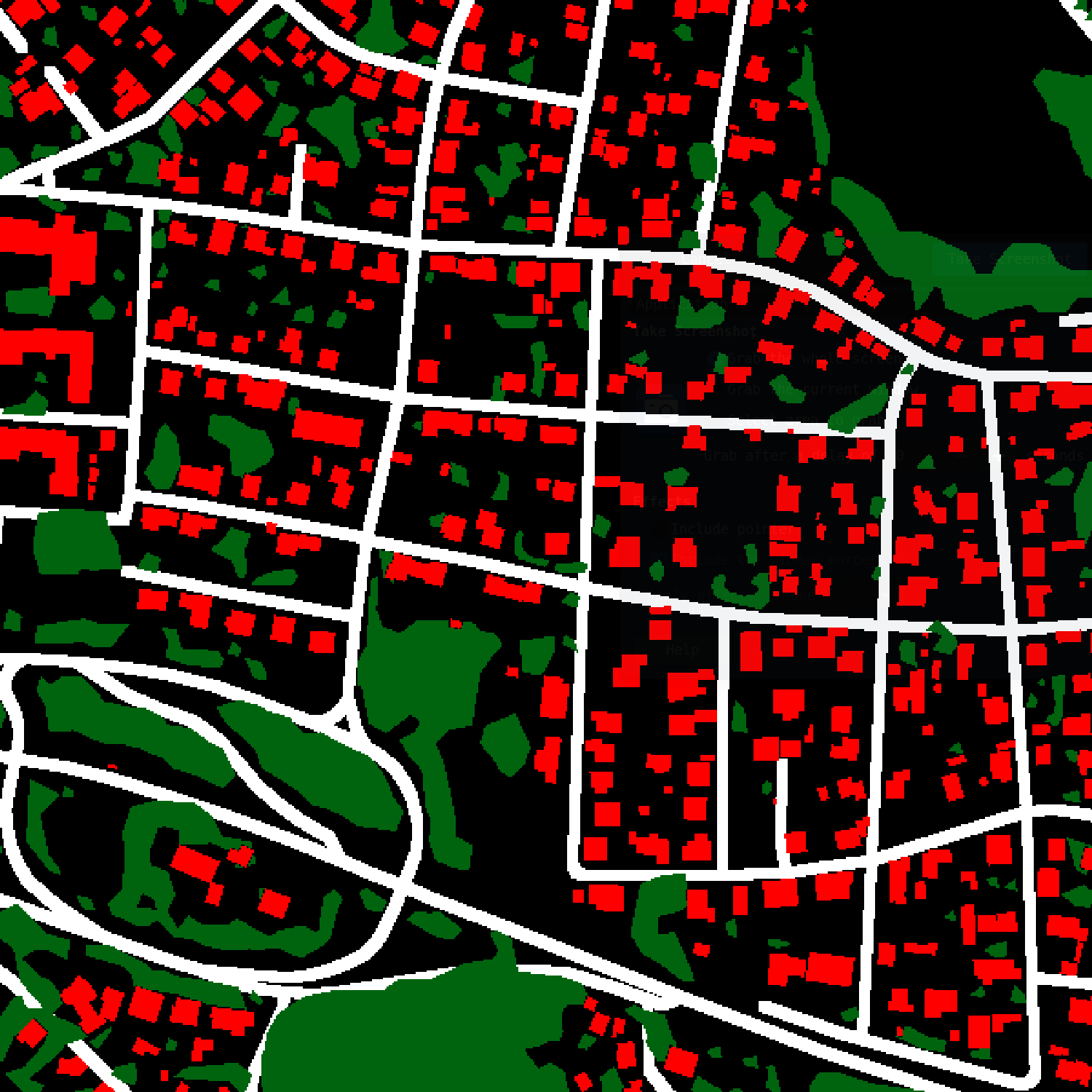}}
\hfill
\subfloat[U-net~\cite{ronneberger2015u}]{\includegraphics[width = 0.165\linewidth]{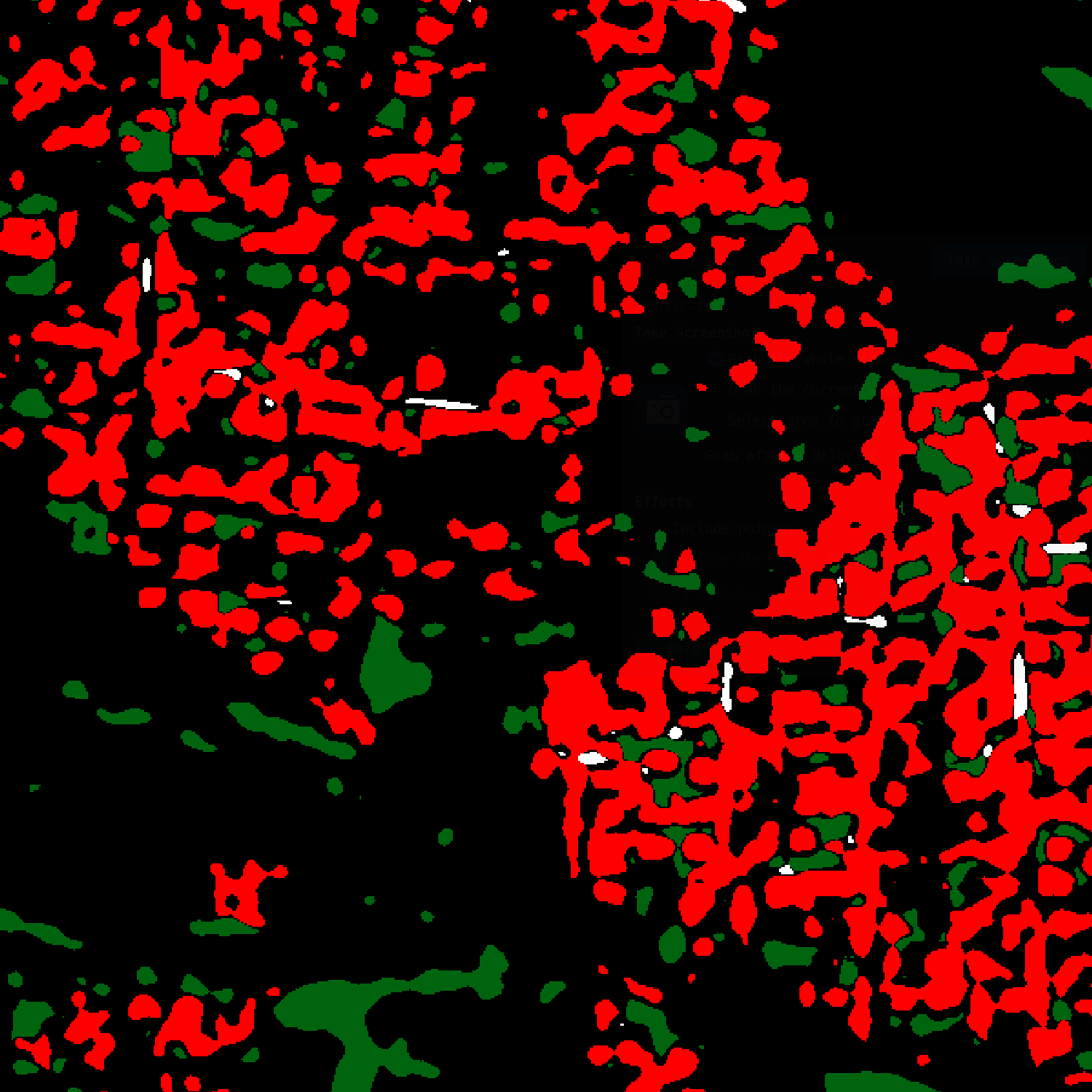}}
\hfill
\subfloat[AdaptSN S~\cite{tsai2018learning}]{\includegraphics[width = 0.165\linewidth]{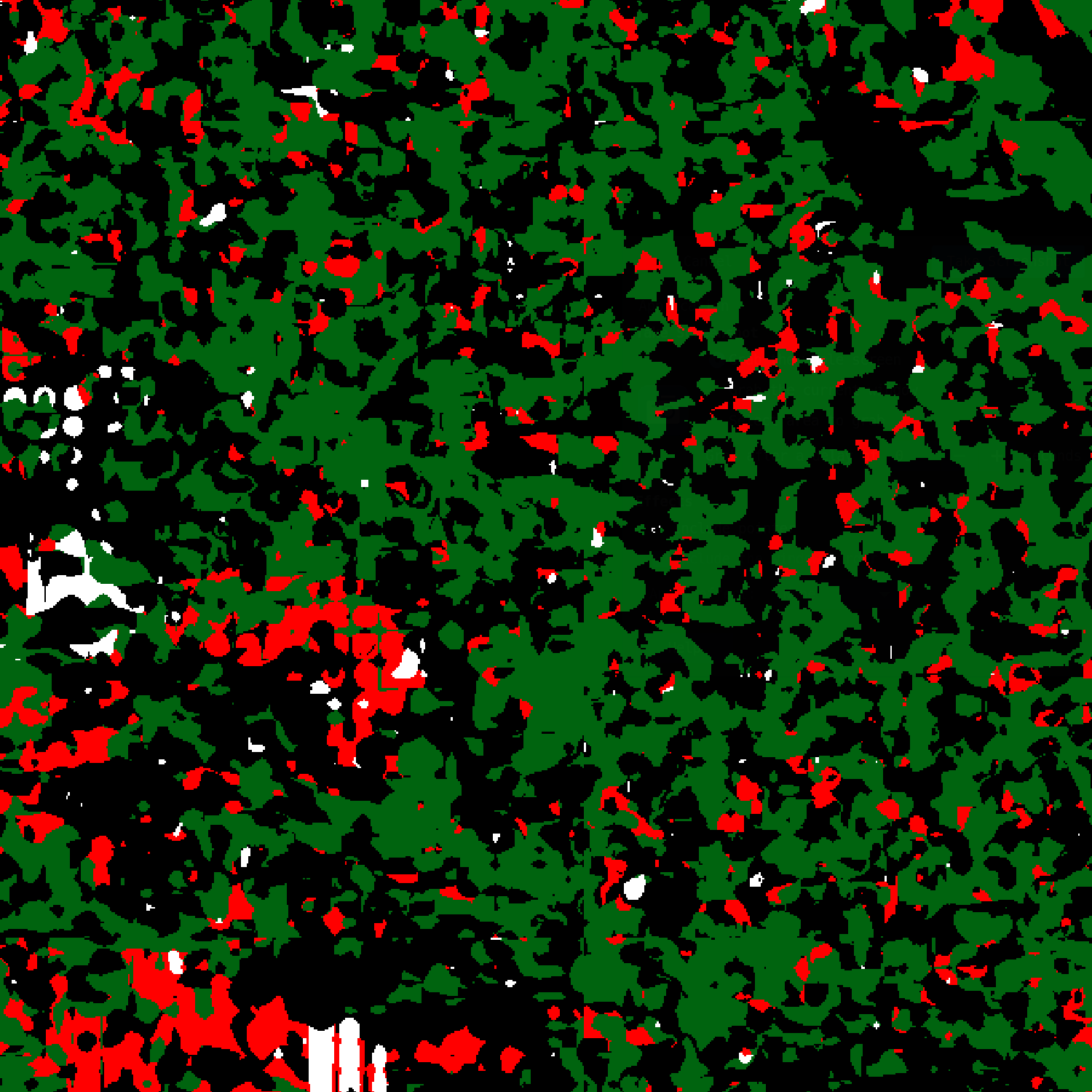}}
\hfill
\subfloat[AdaptSN M~\cite{tsai2018learning}]{\includegraphics[width = 0.165\linewidth]{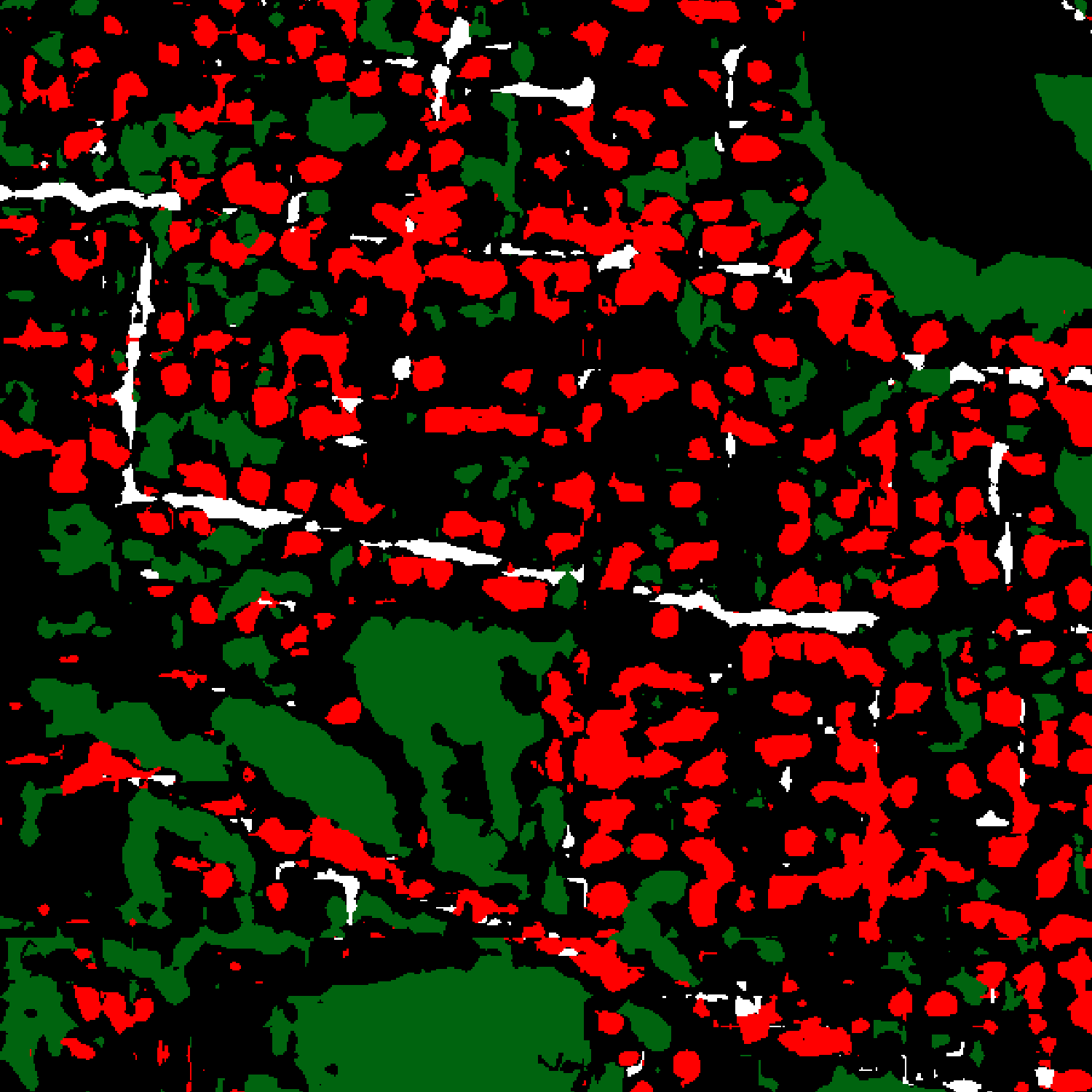}}
\hfill
\subfloat[CycleGAN~\cite{zhu2017unpaired}]{\includegraphics[width = 0.165\linewidth]{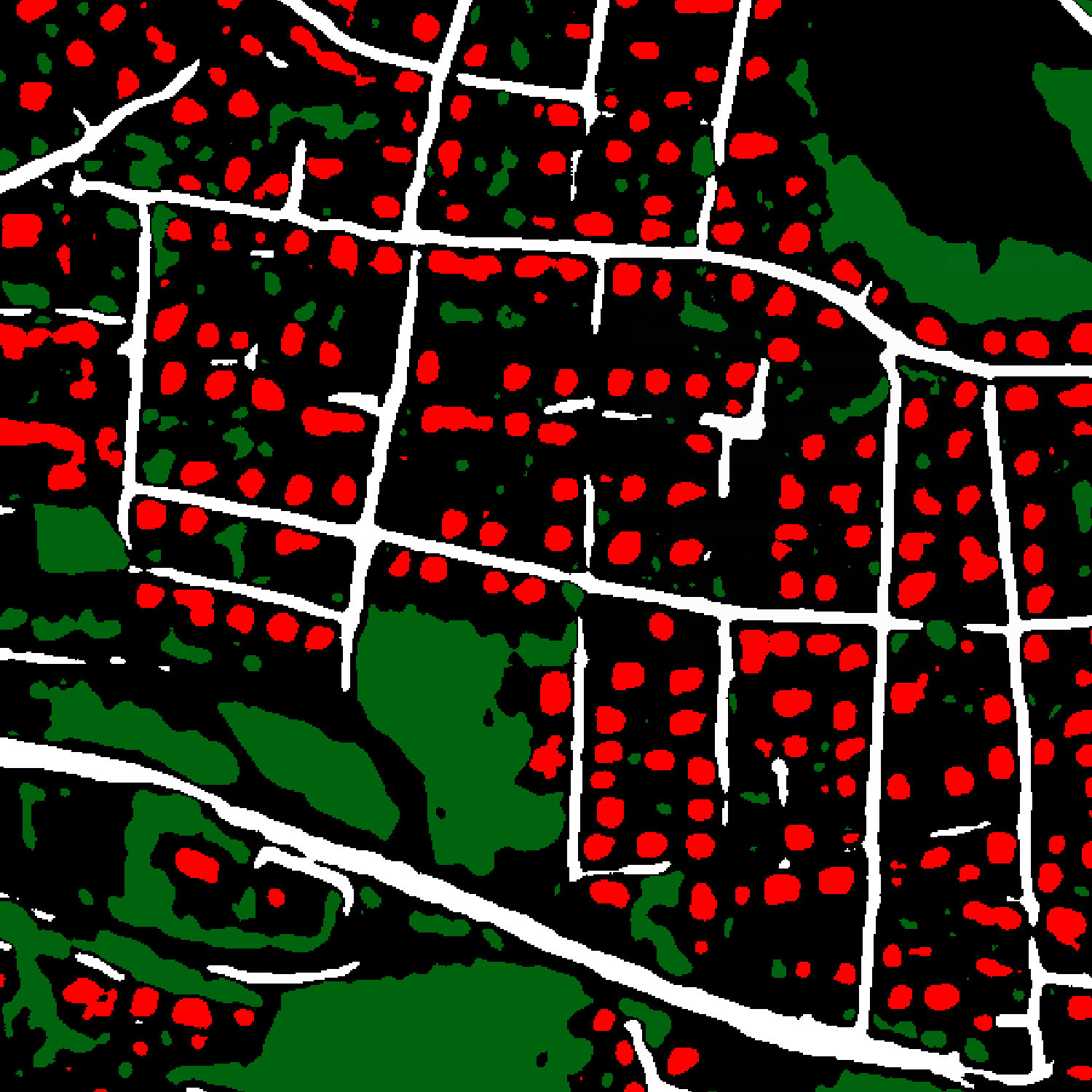}}

\subfloat[UNIT~\cite{liu2017unsupervised}]{\includegraphics[width = 0.165\linewidth]{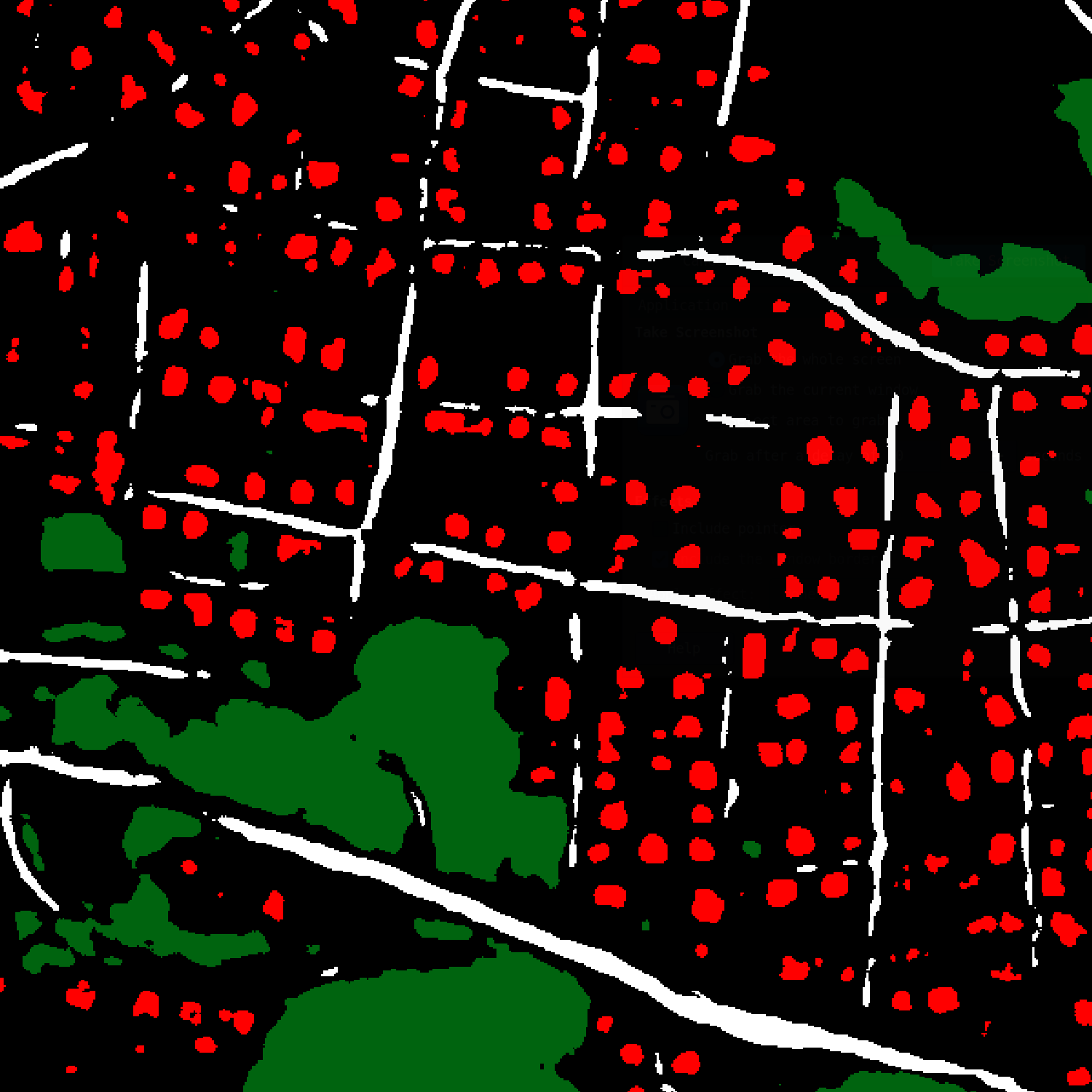}}
\hfill
\subfloat[MUNIT~\cite{huang2018multimodal}]{\includegraphics[width = 0.165\linewidth]{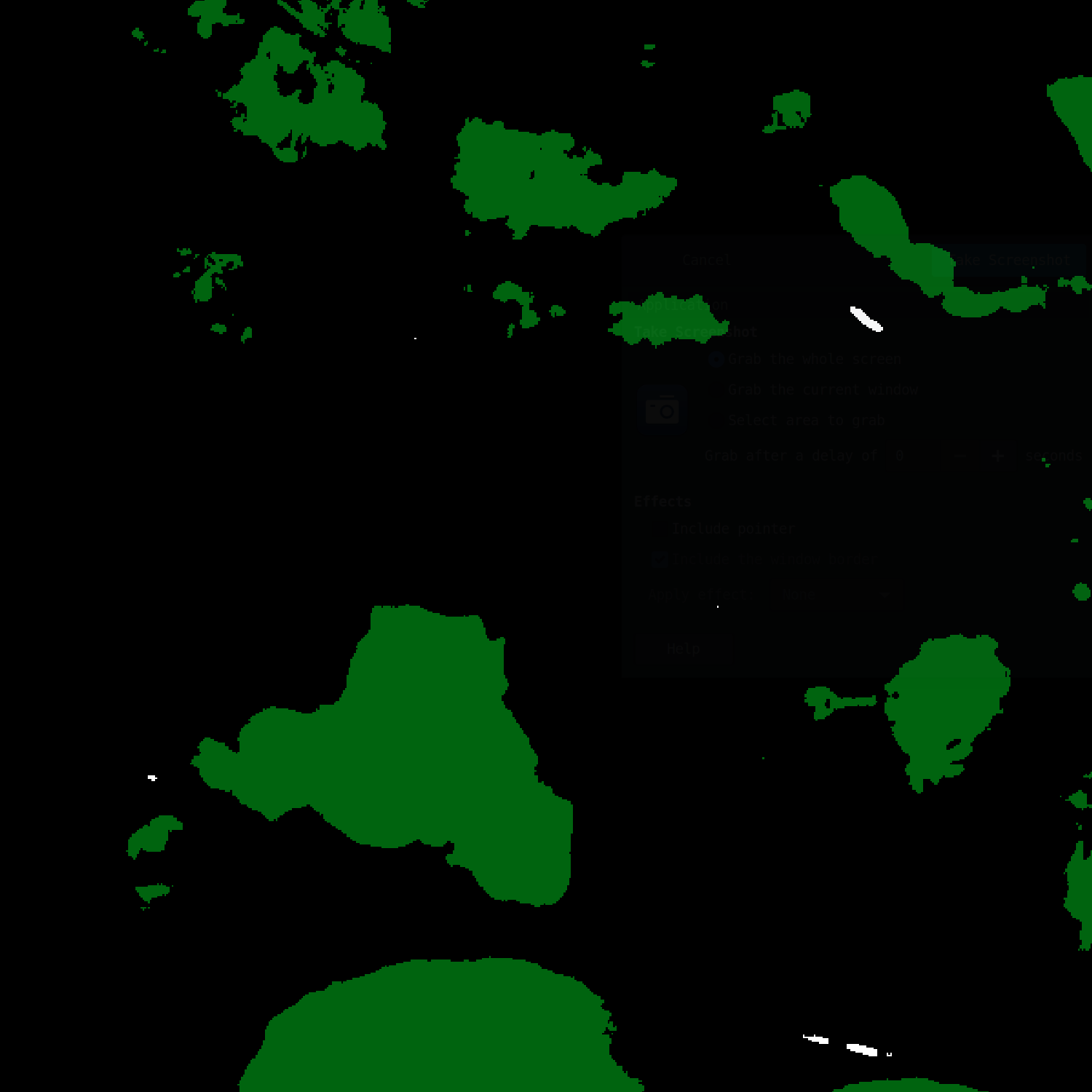}}
\hfill
\subfloat[DRIT~\cite{lee2018diverse} ]{\includegraphics[width = 0.165\linewidth]{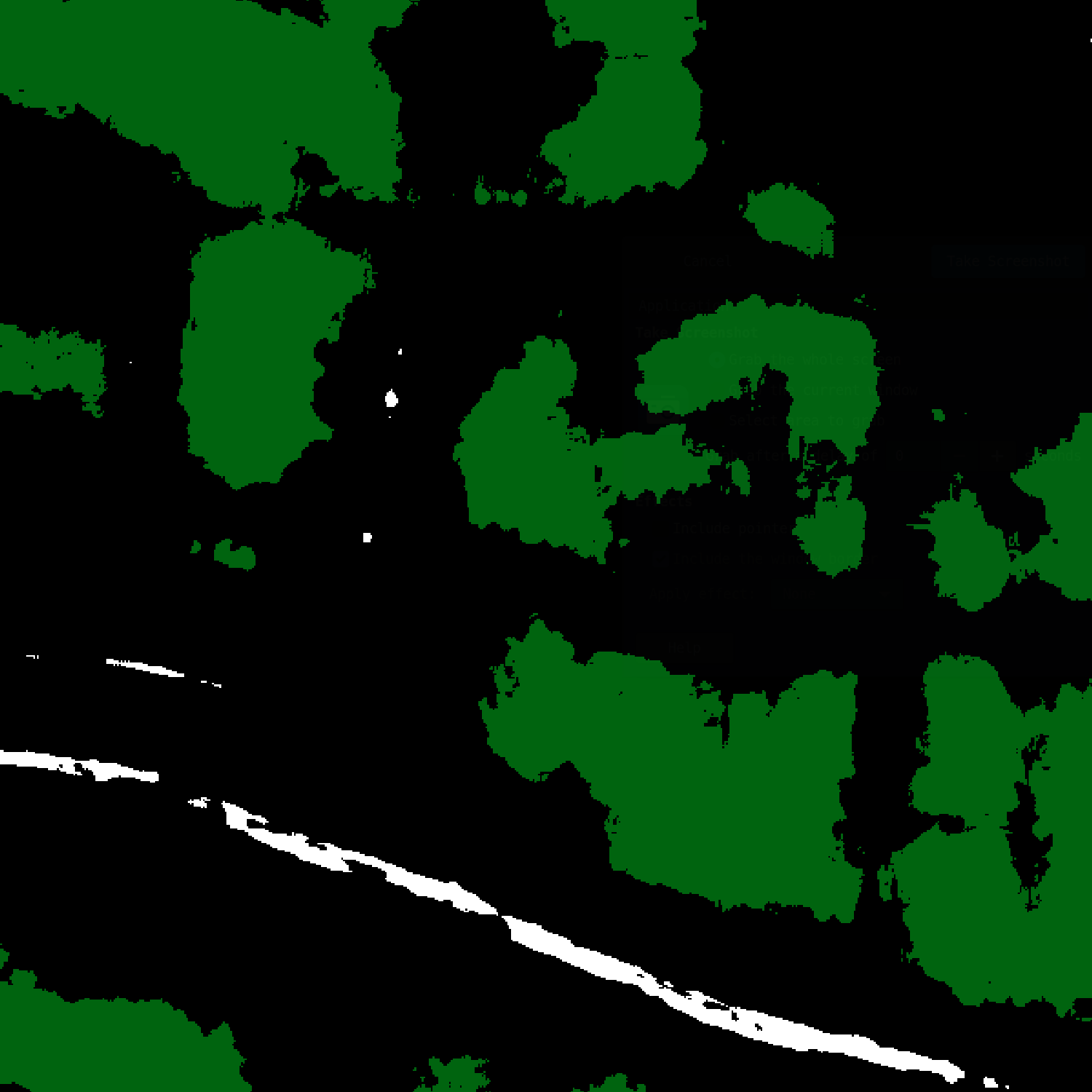}}
\hfill
\subfloat[Gray world~\cite{buchsbaum1980spatial}]{\includegraphics[width = 0.165\linewidth]{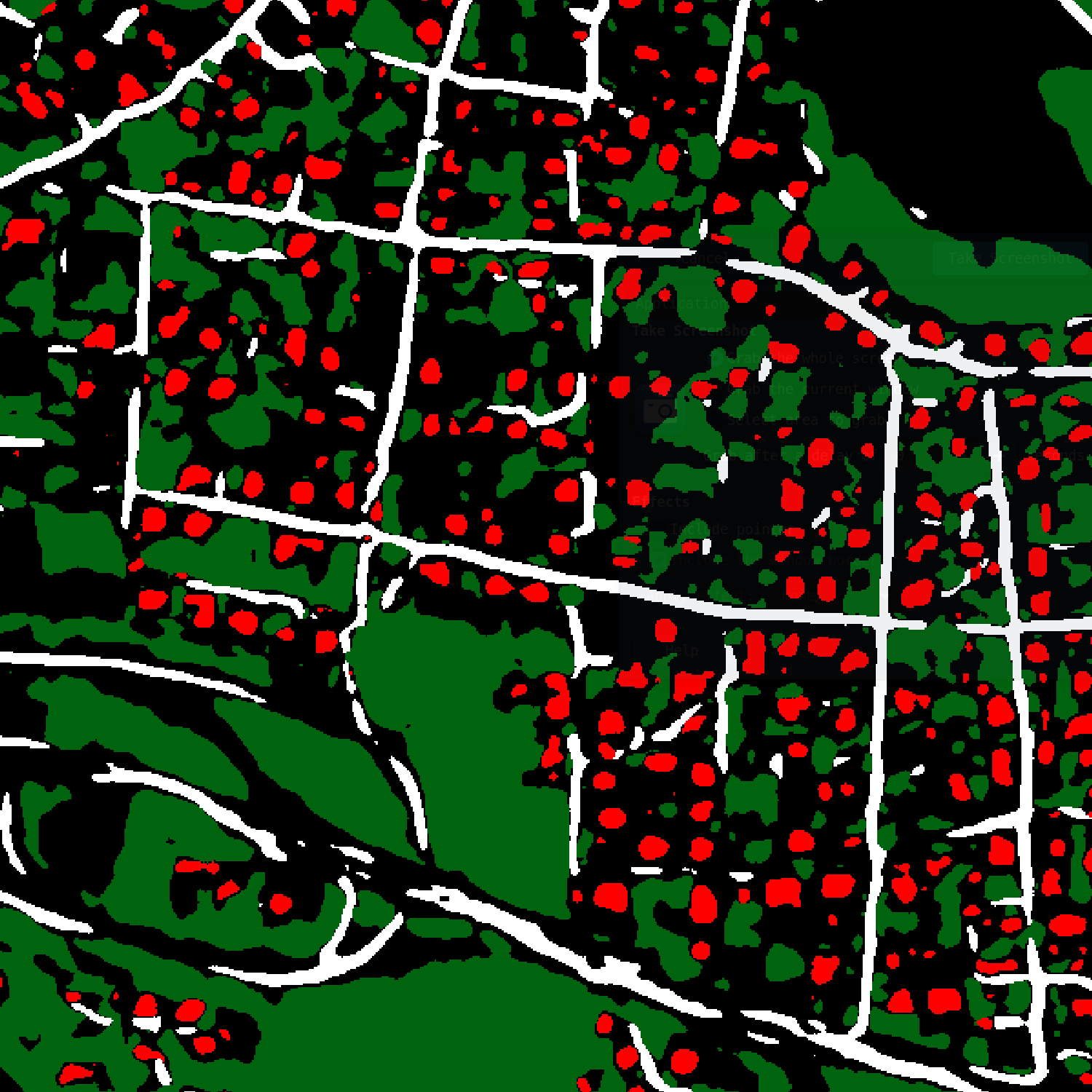}}
\hfill
\subfloat[Hist. match.~\cite{Gonzalez}]{\includegraphics[width = 0.165\linewidth]{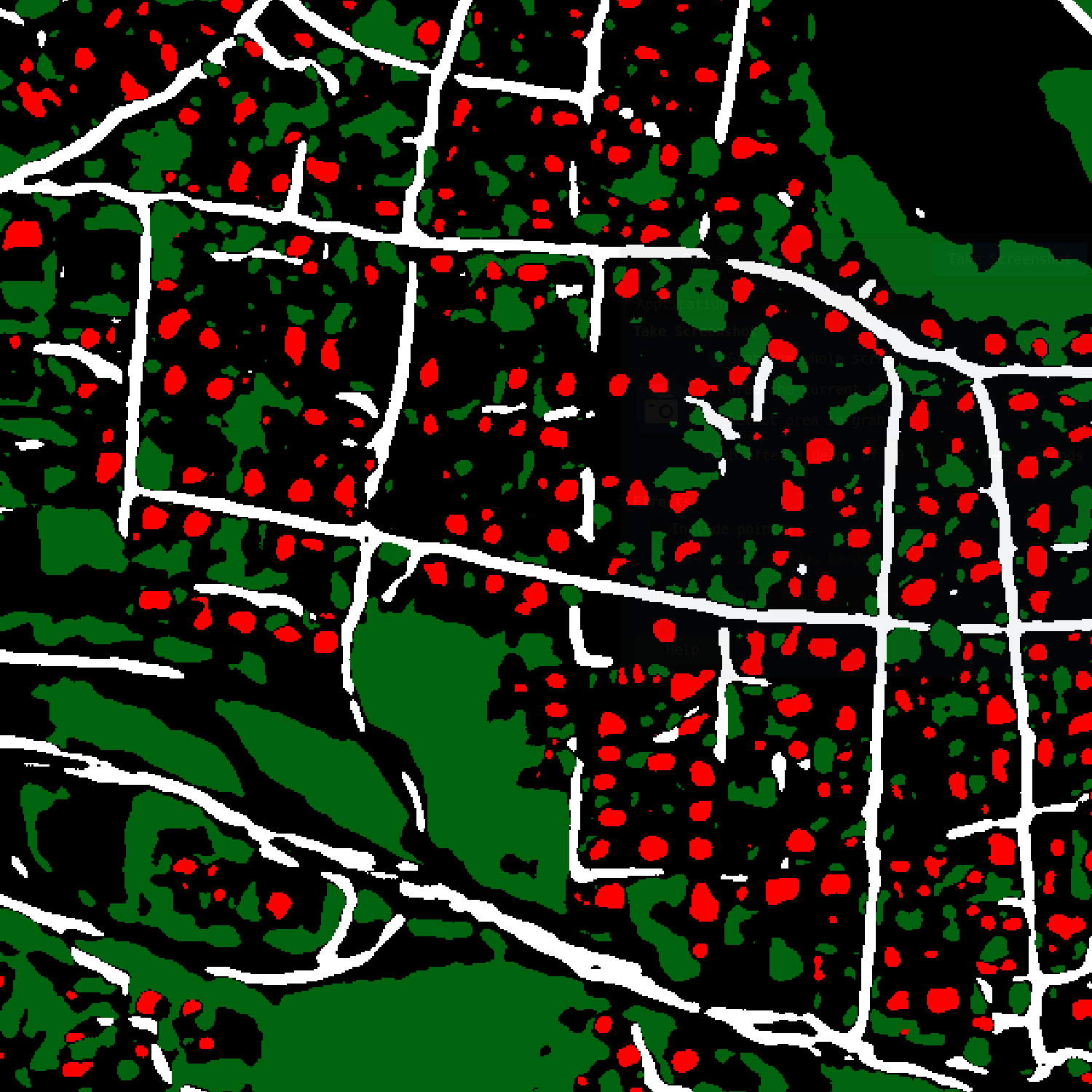}}
\hfill
\subfloat[ColorMapGAN]{\includegraphics[width = 0.165\linewidth]{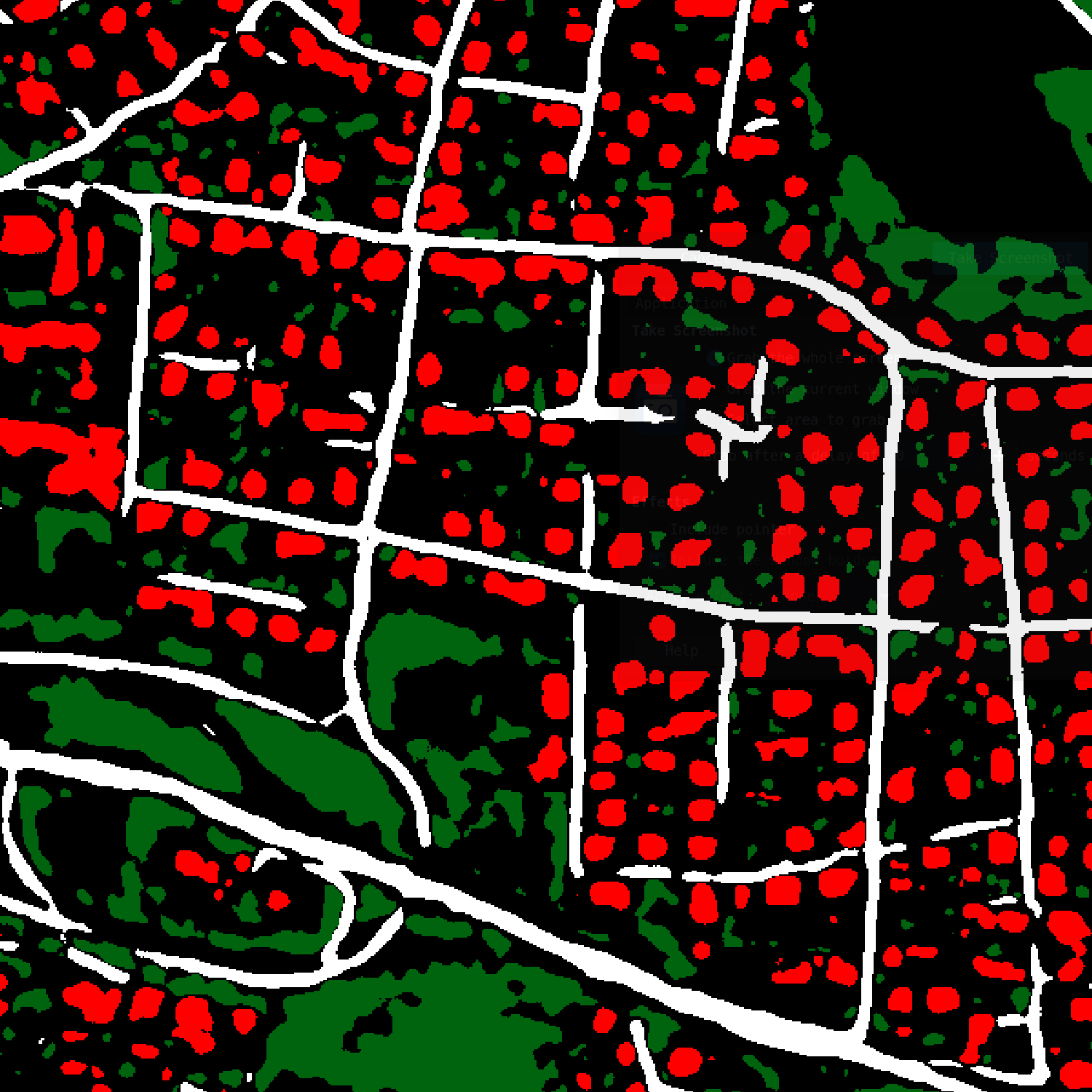}}
\caption{\textit{Villach}, ground-truth, and the predictions. \textit{Building, road, tree}, and background classes are represented by red, white, green, and black colors.}
\label{fig:villach_preds}
\end{figure*}

\begin{figure*}
\subfloat[\textit{B{\'e}ziers}]{\includegraphics[width = 0.165\linewidth]{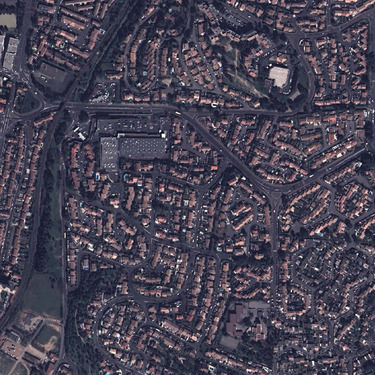}}
\hfill
\subfloat[Ground-truth]{\includegraphics[width = 0.165\linewidth]{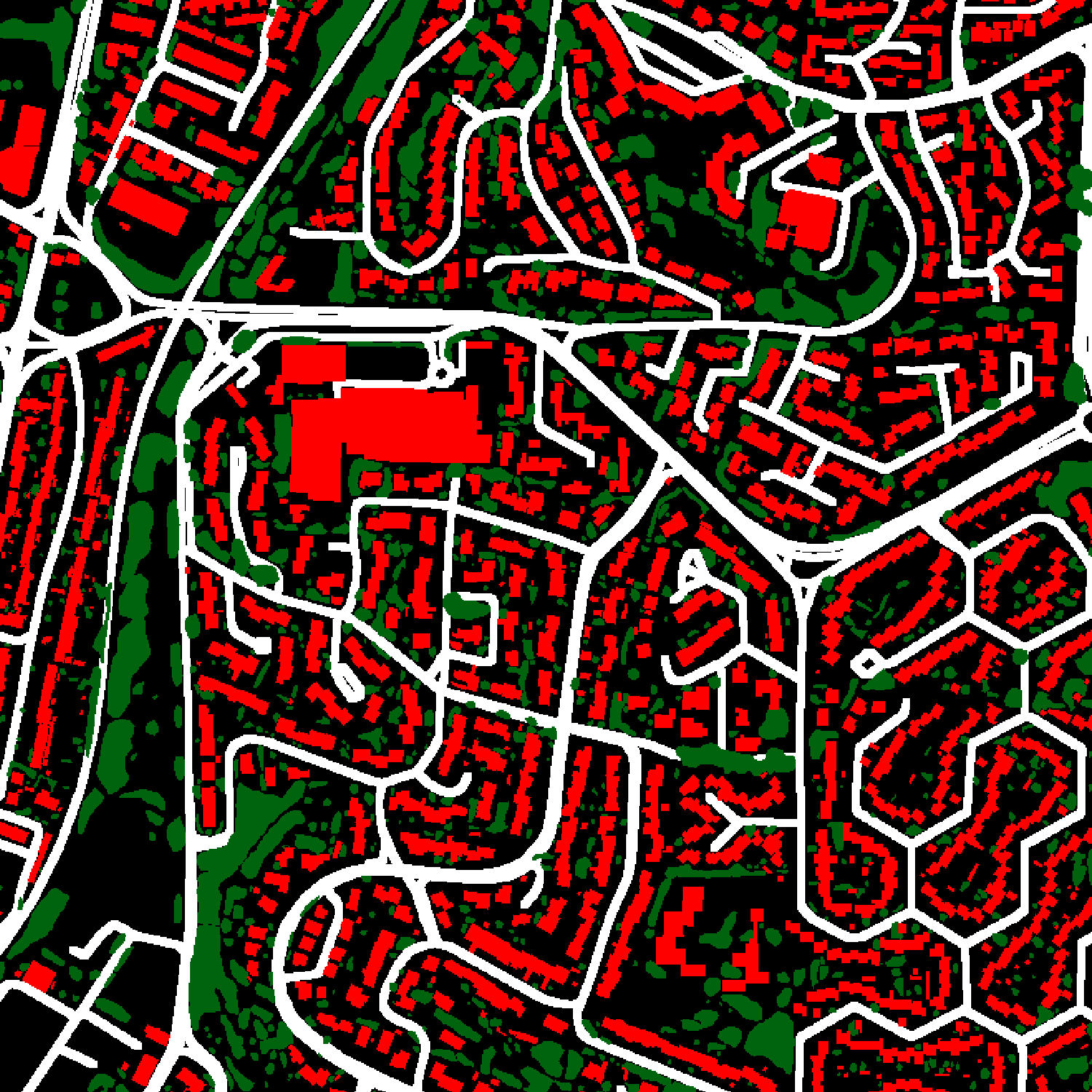}}
\hfill
\subfloat[U-net~\cite{ronneberger2015u}]{\includegraphics[width = 0.165\linewidth]{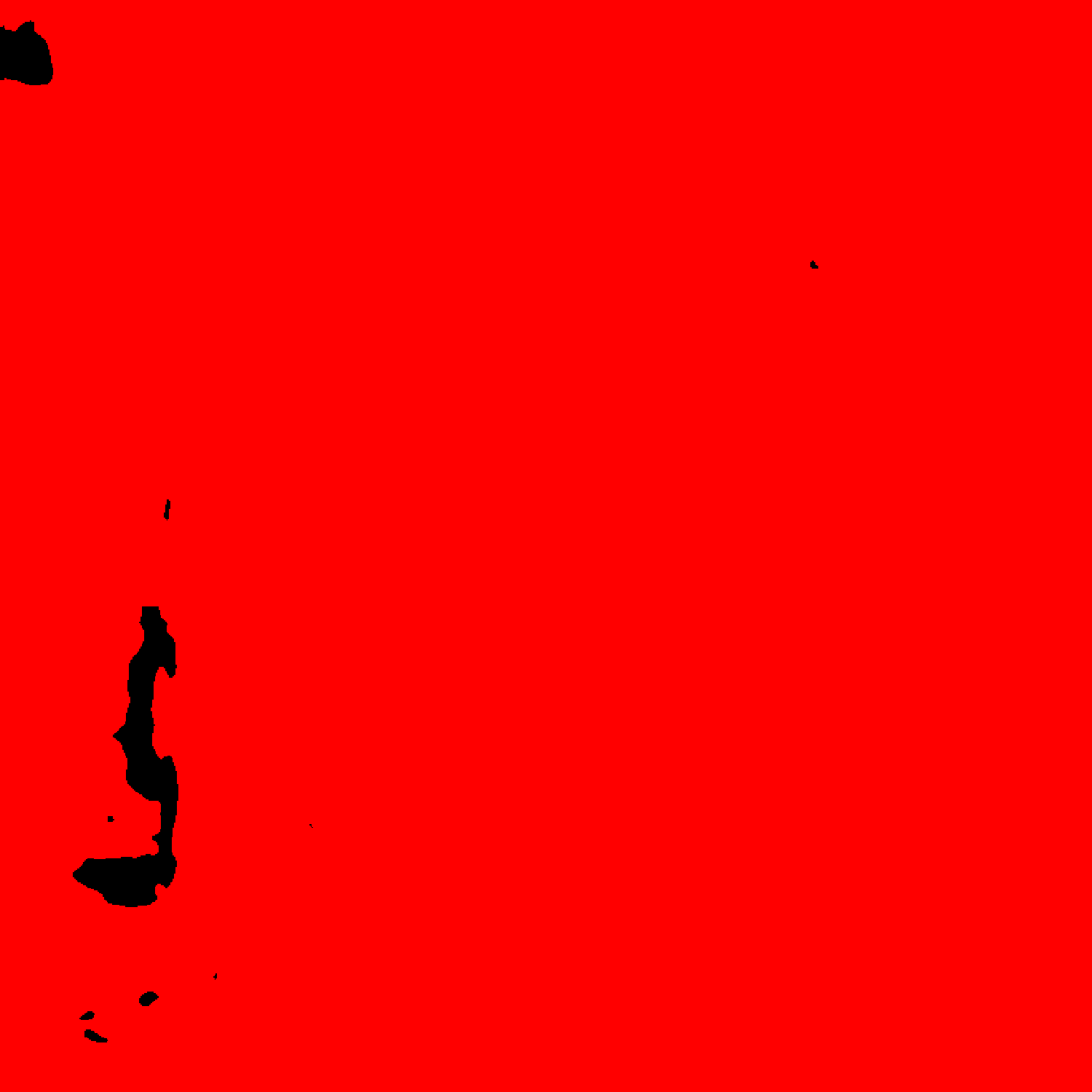}}
\hfill
\subfloat[AdaptSN S~\cite{tsai2018learning}]{\includegraphics[width = 0.165\linewidth]{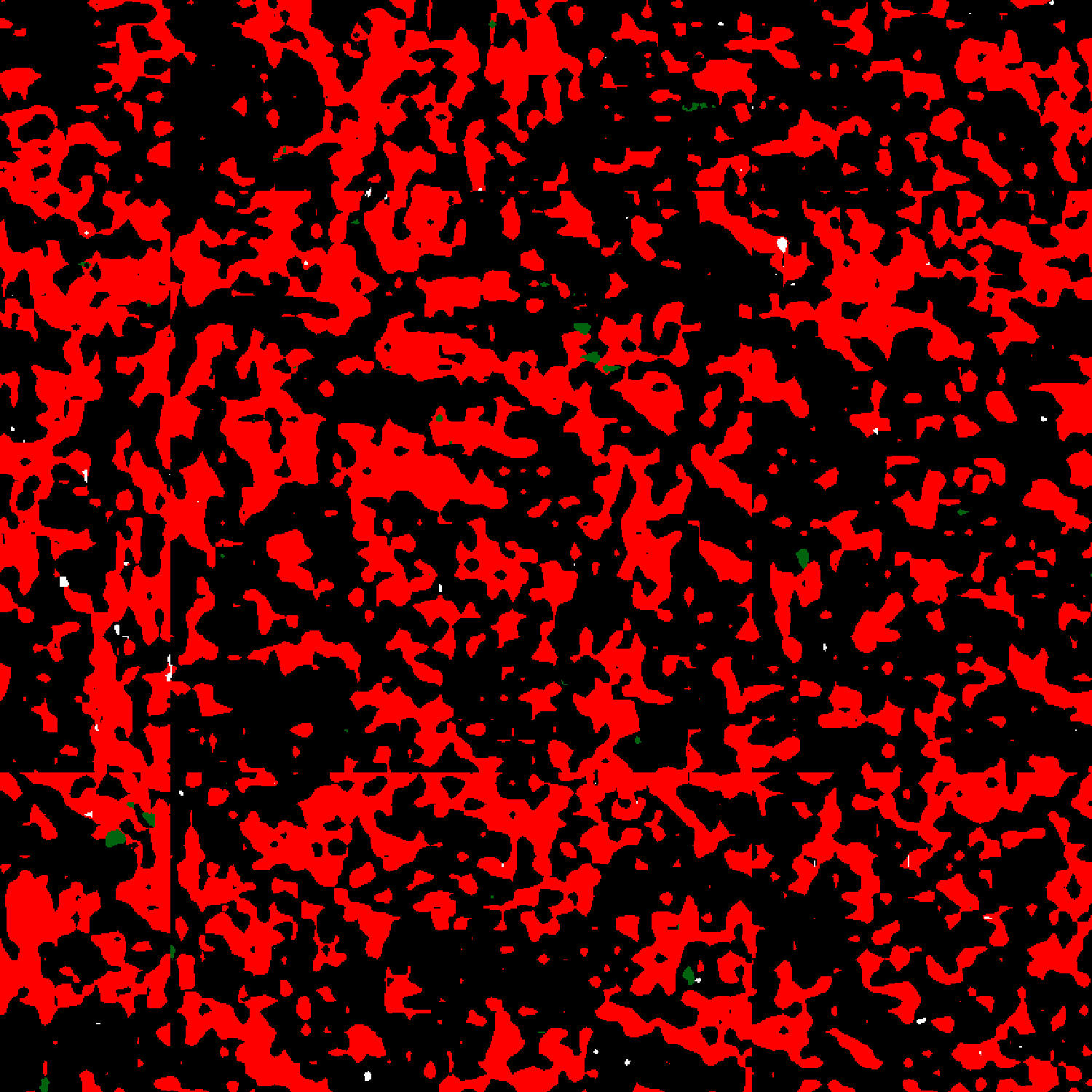}}
\hfill
\subfloat[AdaptSN M~\cite{tsai2018learning}]{\includegraphics[width = 0.165\linewidth]{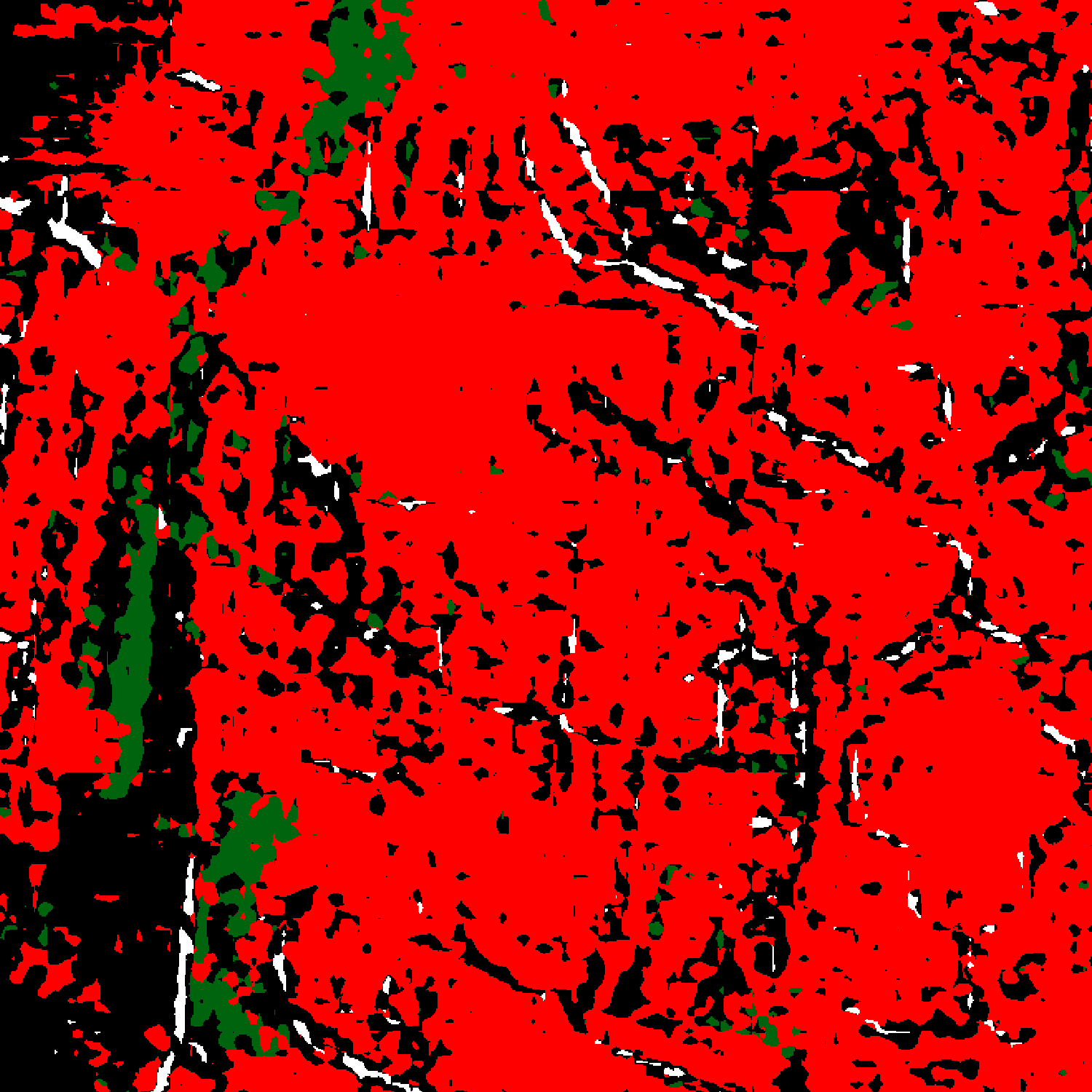}}
\hfill
\subfloat[CycleGAN~\cite{zhu2017unpaired}]{\includegraphics[width = 0.165\linewidth]{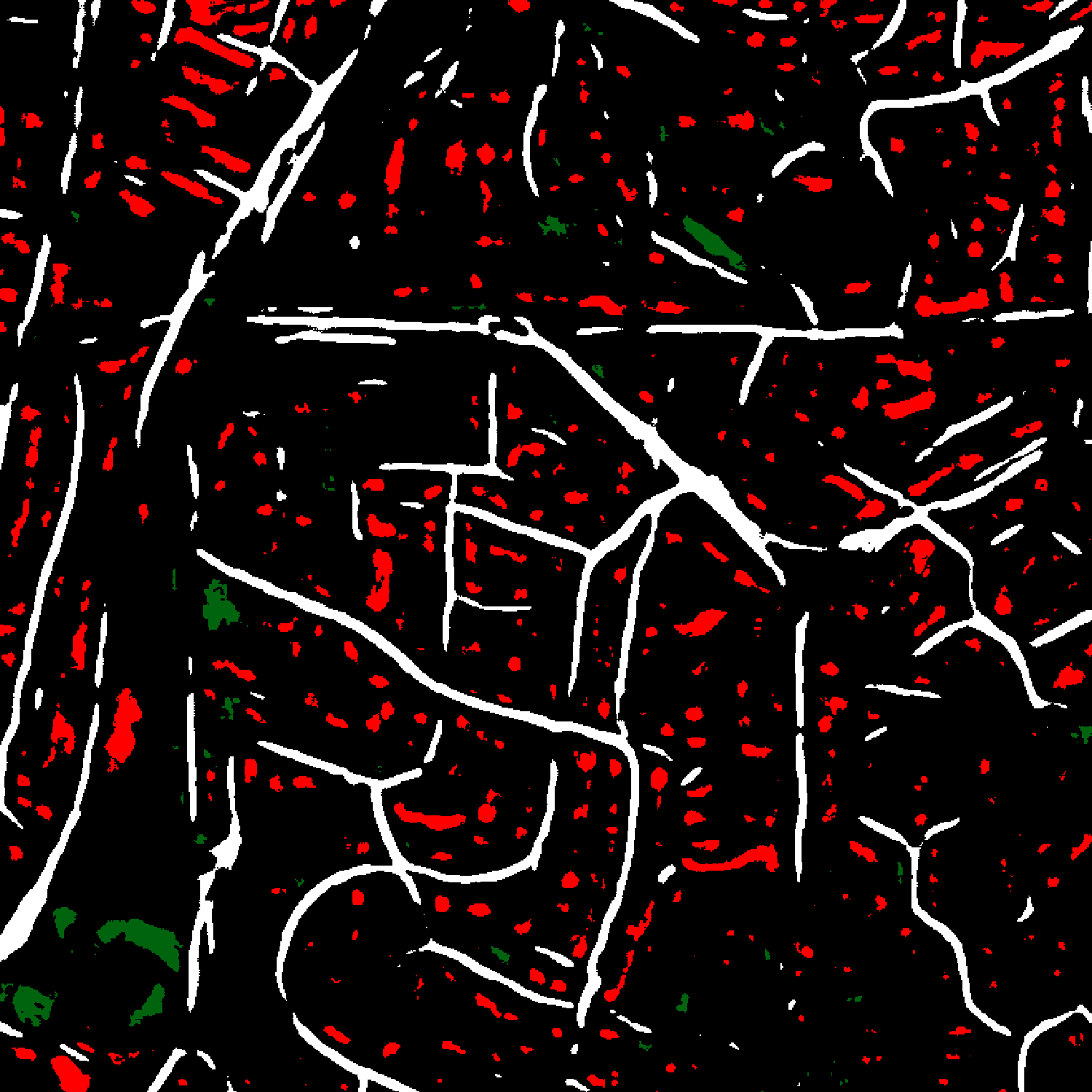}}

\subfloat[UNIT~\cite{liu2017unsupervised}]{\includegraphics[width = 0.165\linewidth]{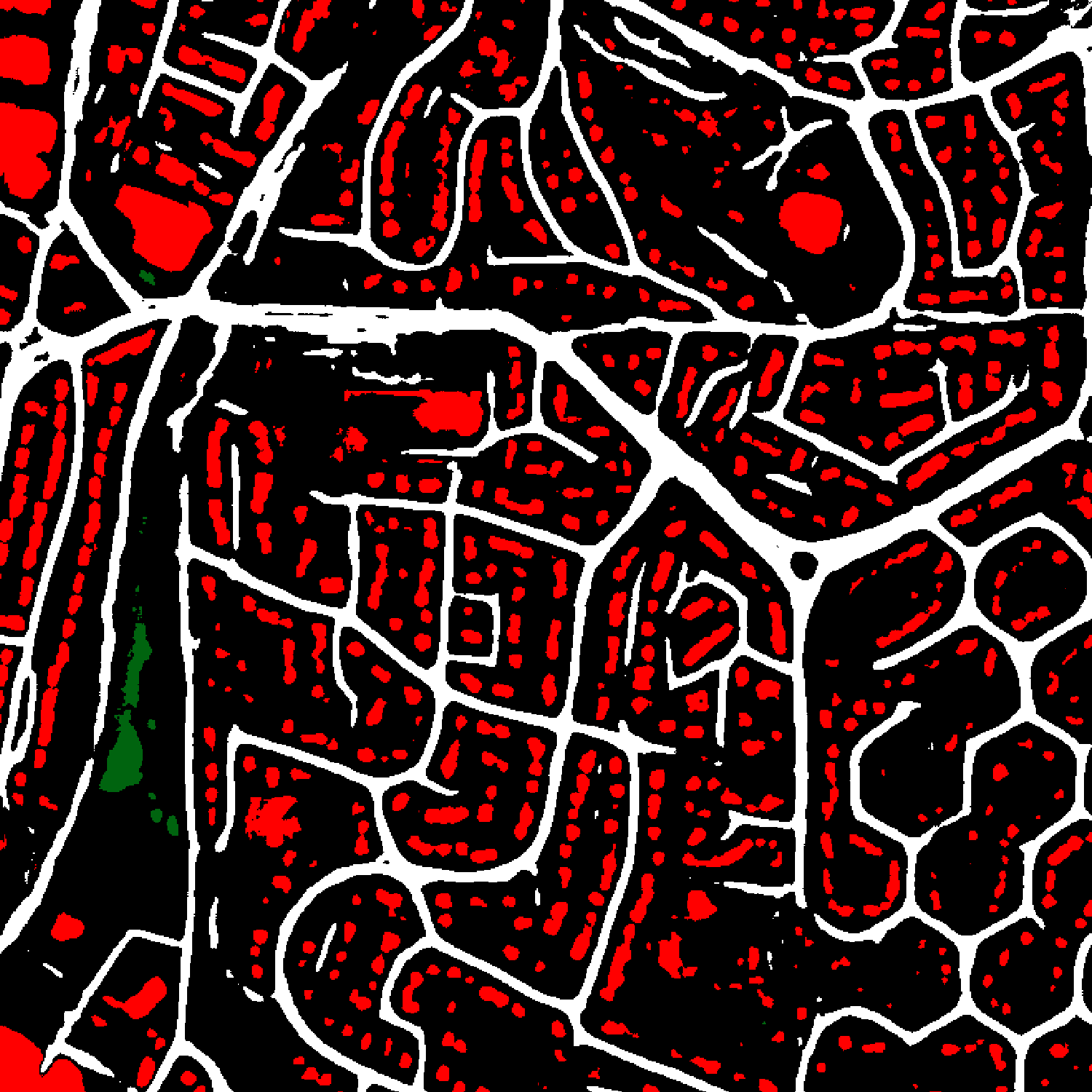}}
\hfill
\subfloat[MUNIT~\cite{huang2018multimodal}]{\includegraphics[width = 0.165\linewidth]{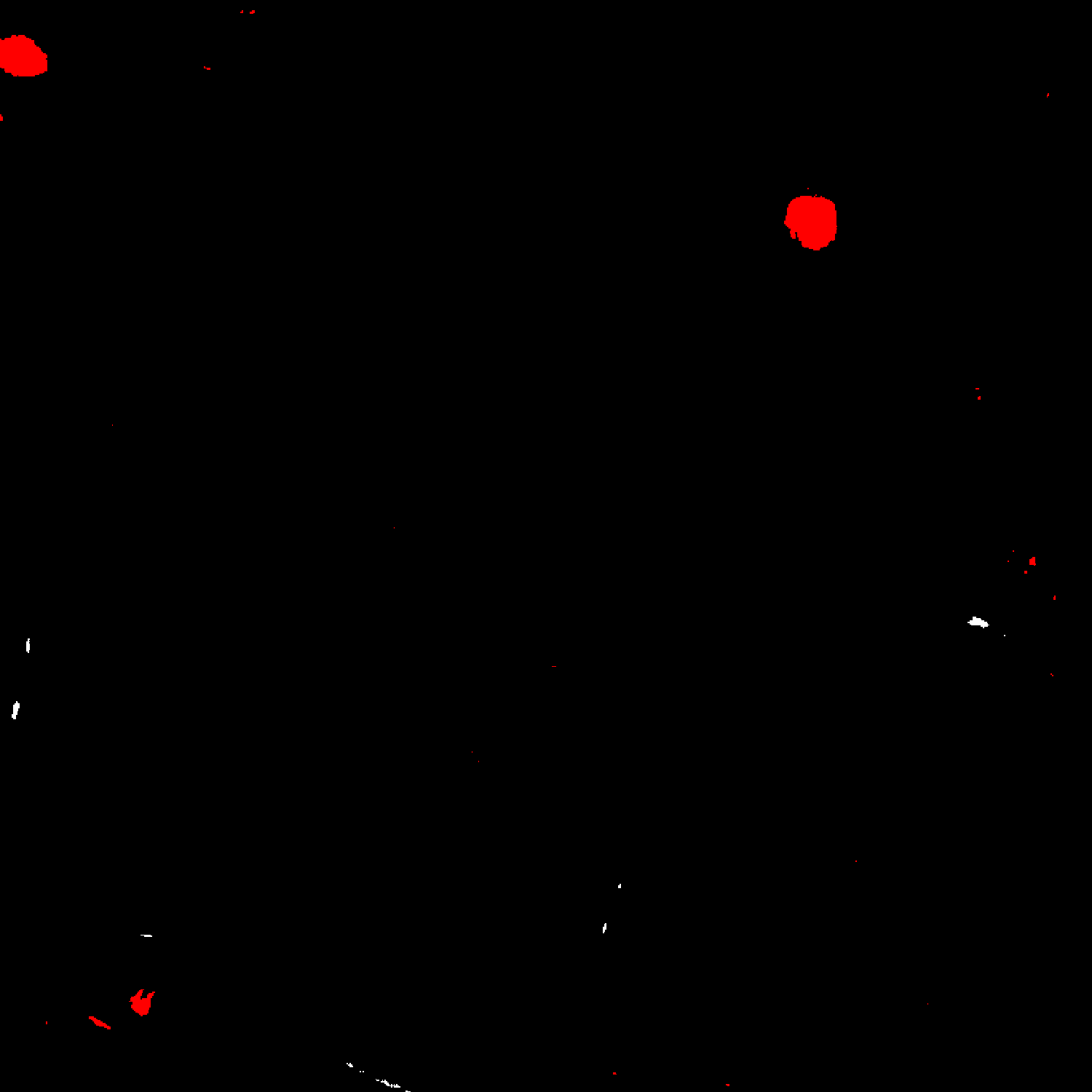}}
\hfill
\subfloat[DRIT~\cite{lee2018diverse} ]{\includegraphics[width = 0.165\linewidth]{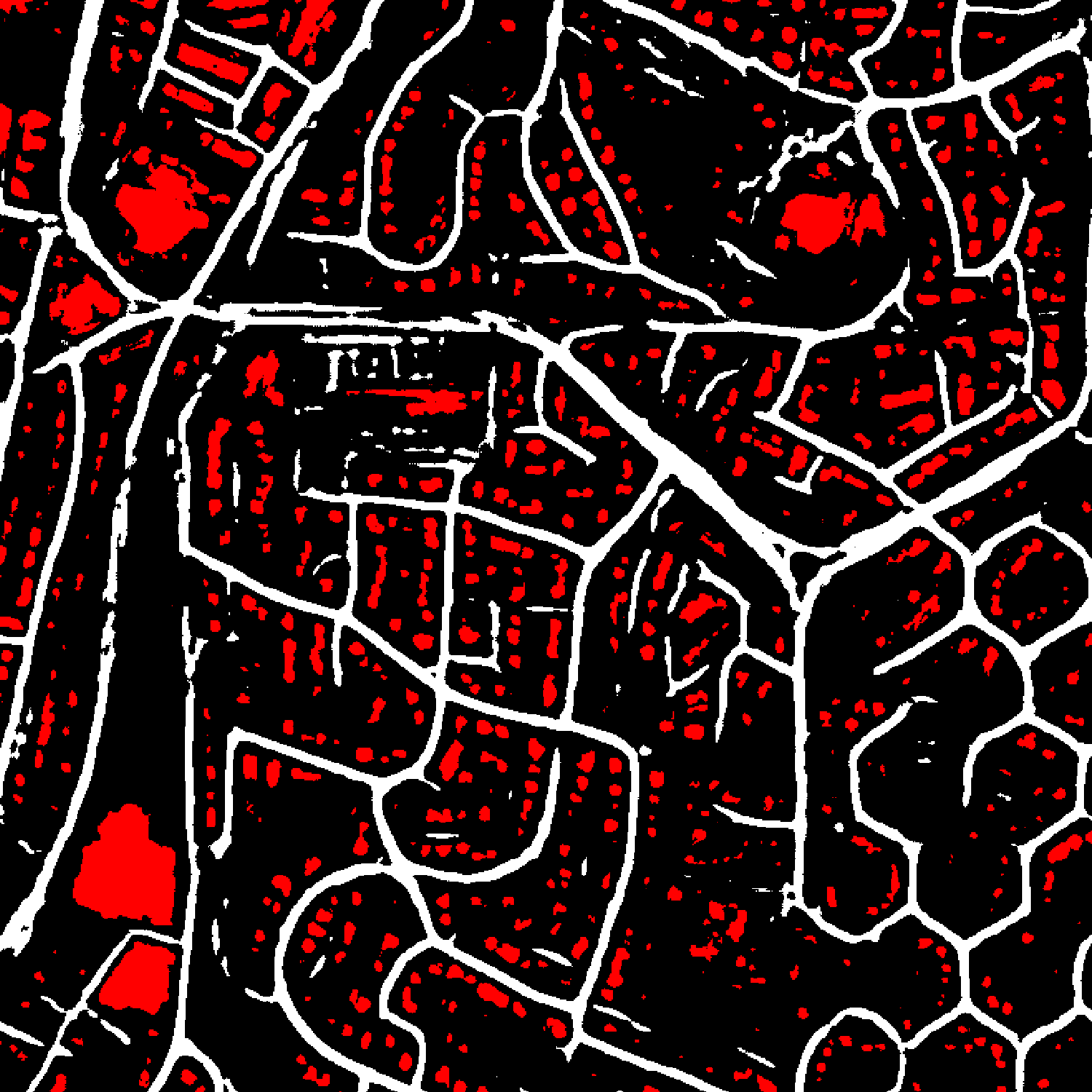}}
\hfill
\subfloat[Gray world~\cite{buchsbaum1980spatial}]{\includegraphics[width = 0.165\linewidth]{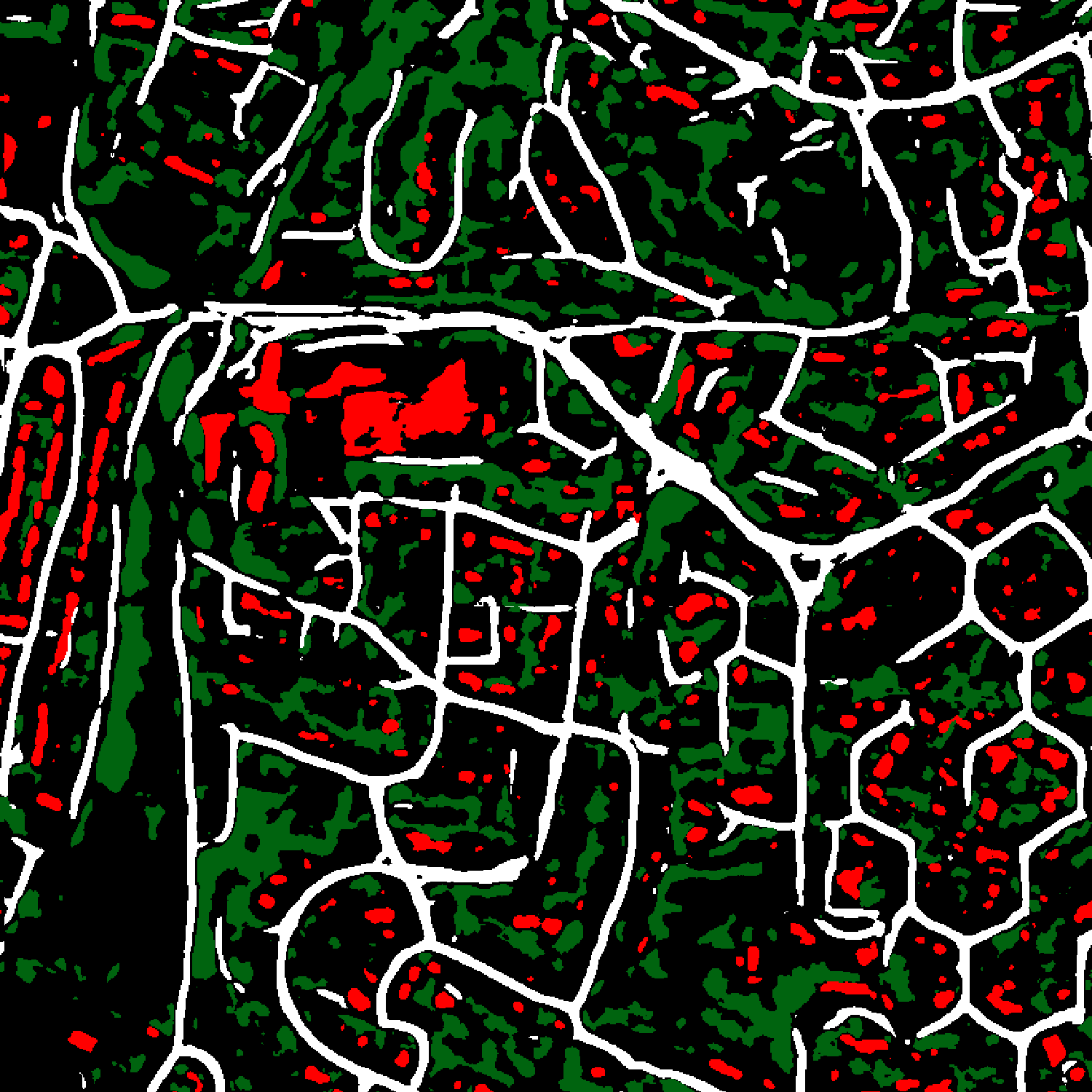}}
\hfill
\subfloat[Hist. match.~\cite{Gonzalez}]{\includegraphics[width = 0.165\linewidth]{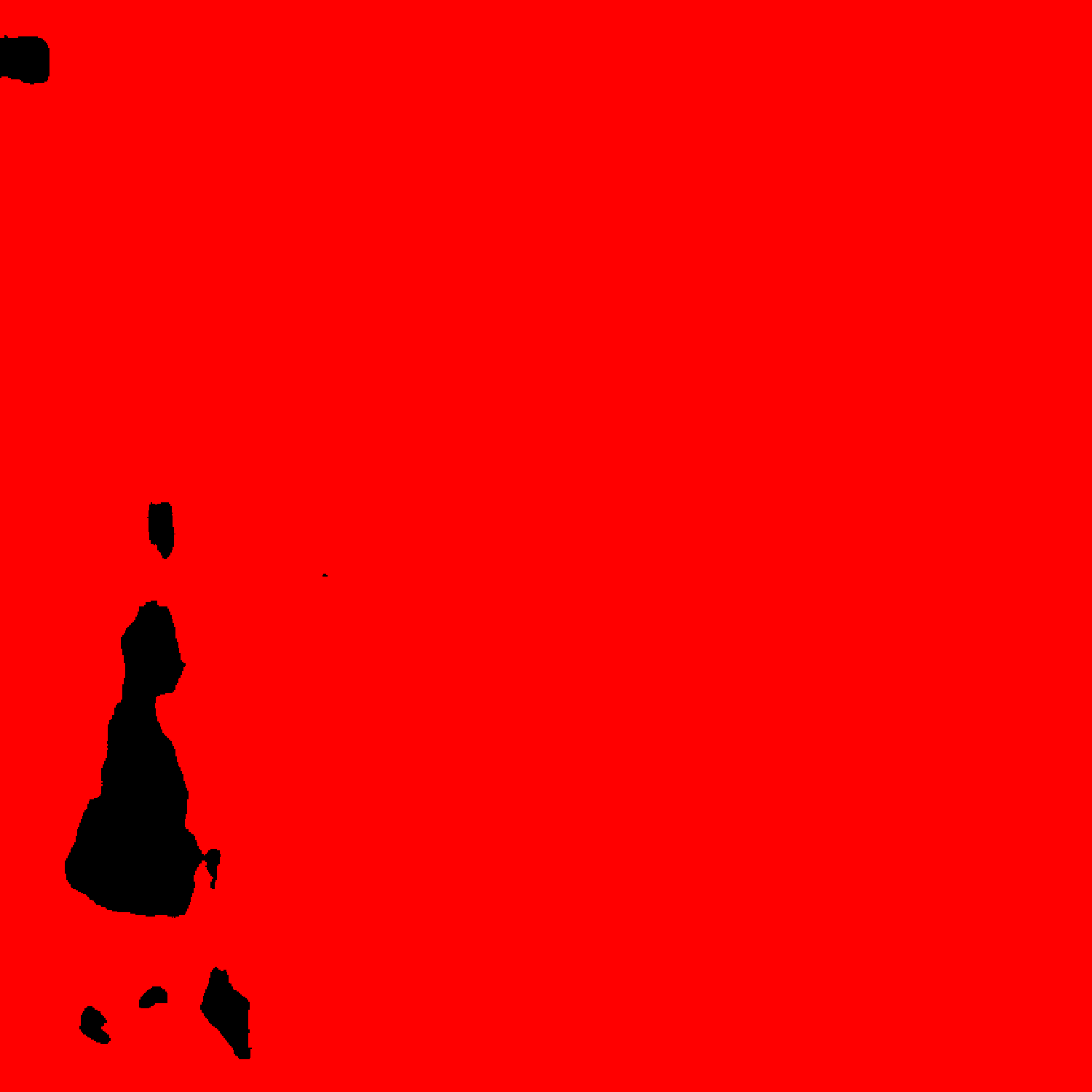}}
\hfill
\subfloat[ColorMapGAN]{\includegraphics[width = 0.165\linewidth]{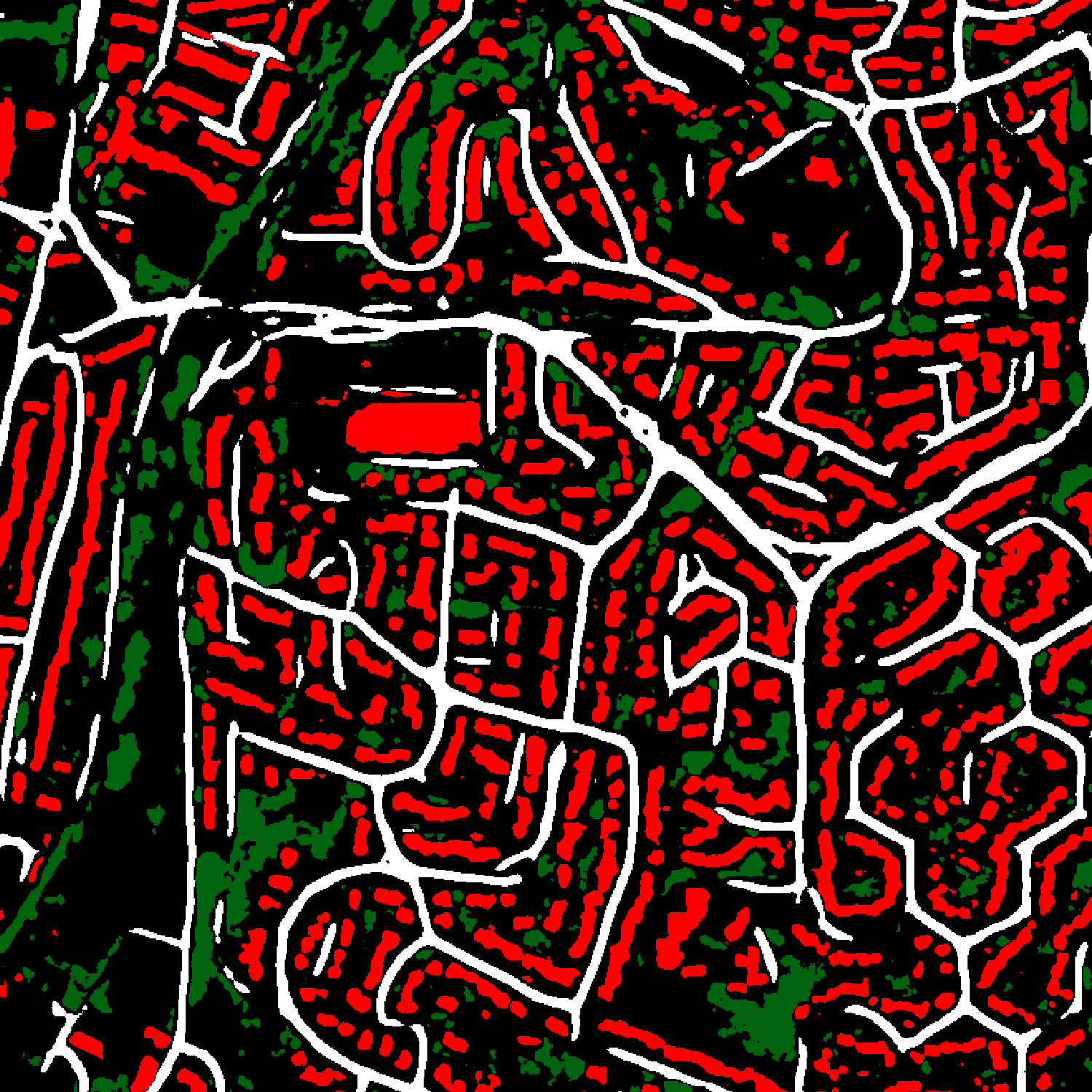}}
\caption{\textit{B{\'e}ziers}, ground-truth, and the predictions. \textit{Building, road, tree}, and background classes are represented by red, white, green, and black colors.}
\label{fig:beziers_preds}
\end{figure*}

\begin{figure*}
\subfloat[\textit{Roanne}]{\includegraphics[width = 0.165\linewidth]{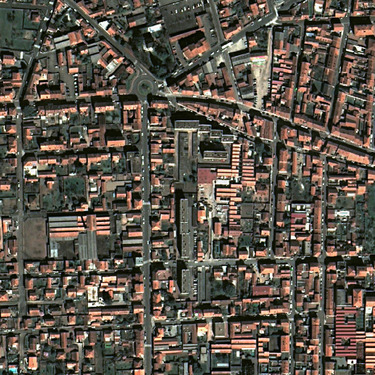}}
\hfill
\subfloat[Ground-truth]{\includegraphics[width = 0.165\linewidth]{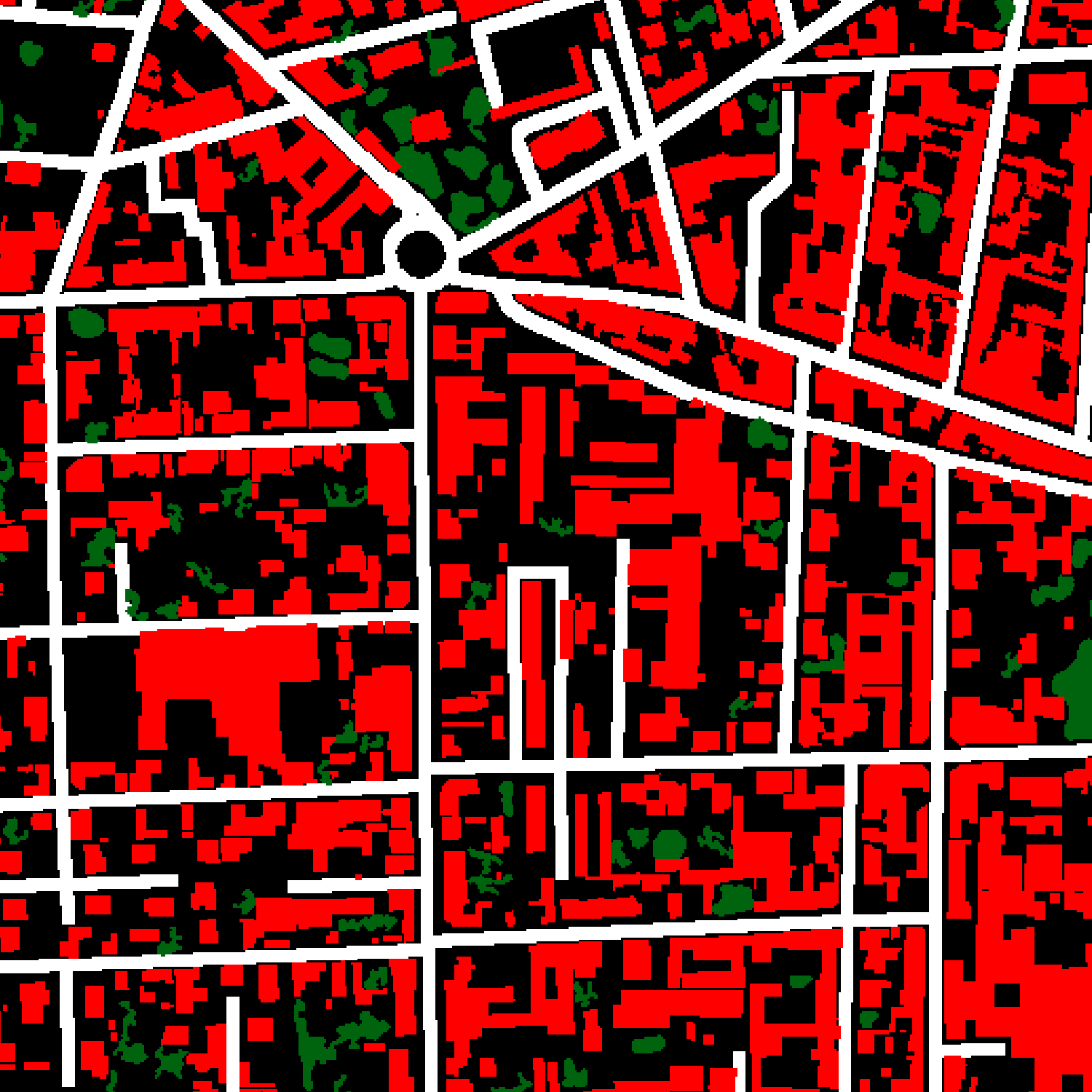}}
\hfill
\subfloat[U-net~\cite{ronneberger2015u}]{\includegraphics[width = 0.165\linewidth]{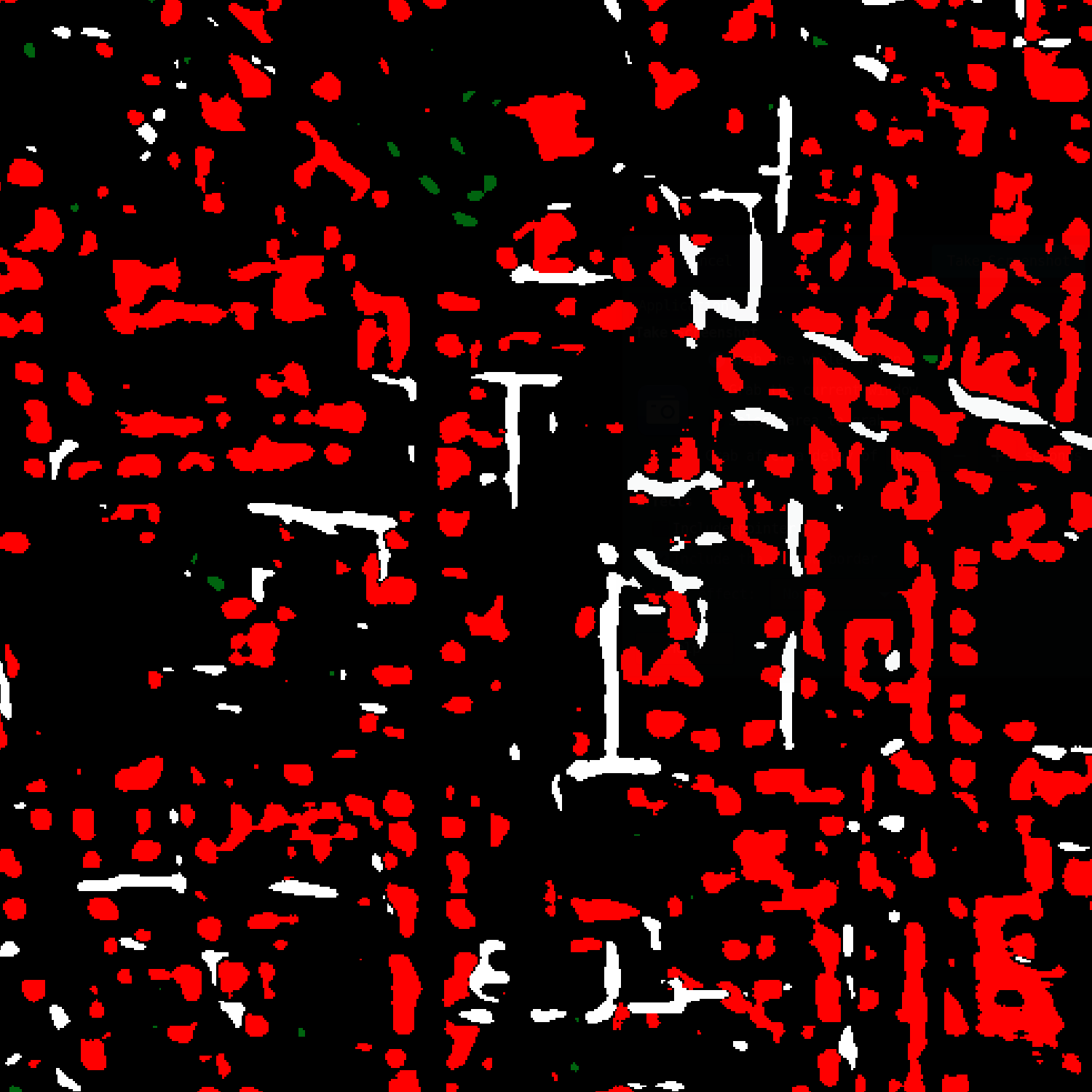}}
\hfill
\subfloat[AdaptSN S~\cite{tsai2018learning}]{\includegraphics[width = 0.165\linewidth]{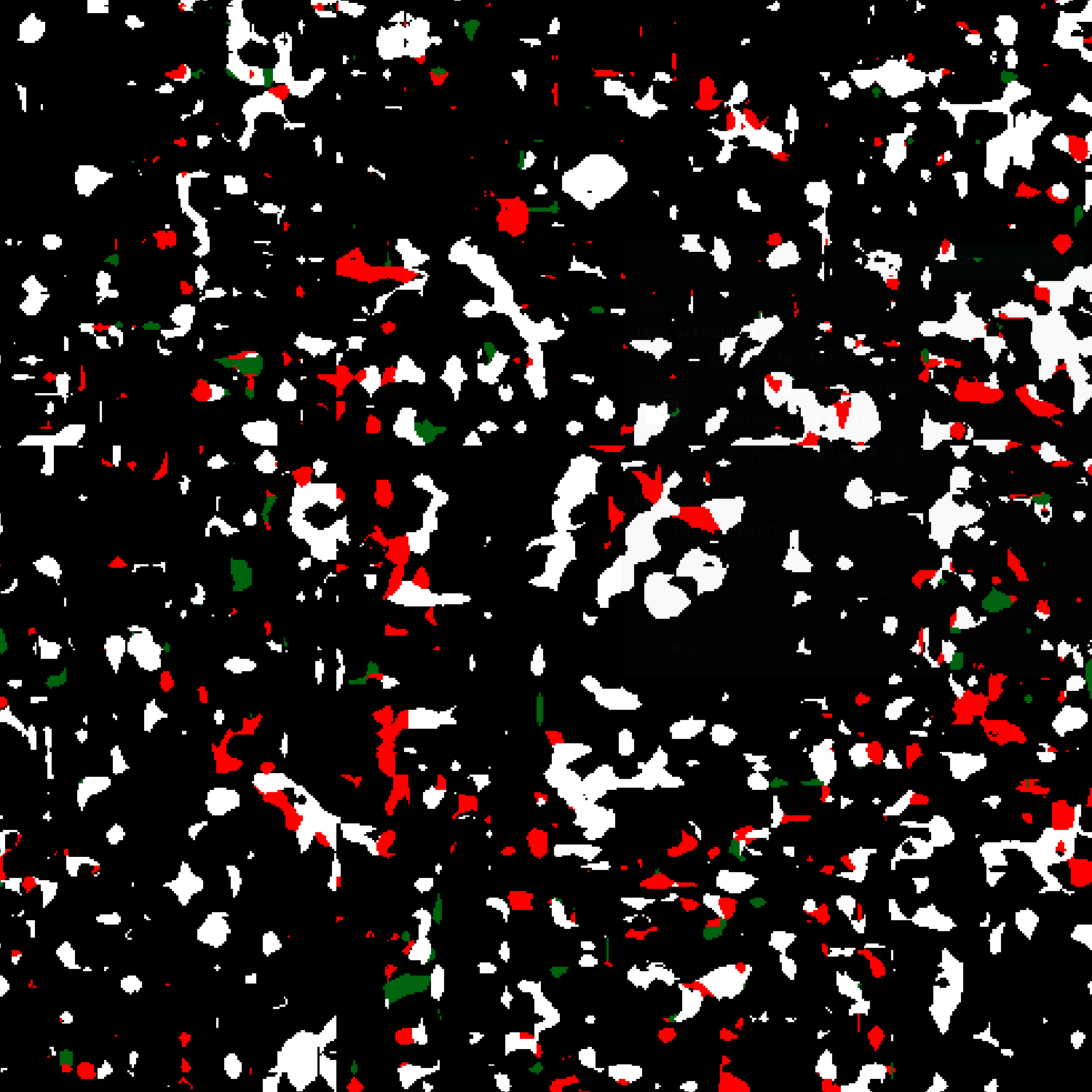}}
\hfill
\subfloat[AdaptSN M~\cite{tsai2018learning}]{\includegraphics[width = 0.165\linewidth]{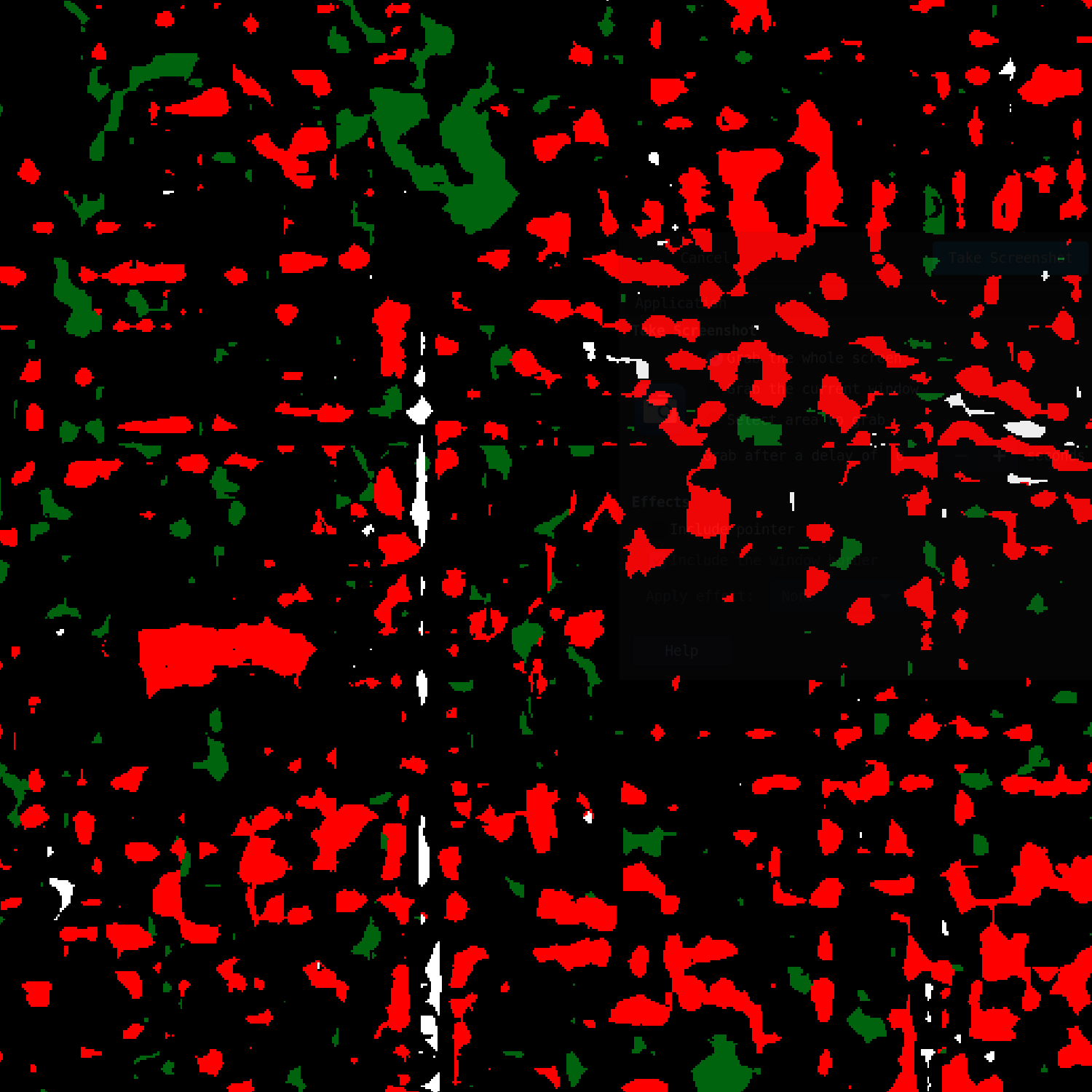}}
\hfill
\subfloat[CycleGAN~\cite{zhu2017unpaired}]{\includegraphics[width = 0.165\linewidth]{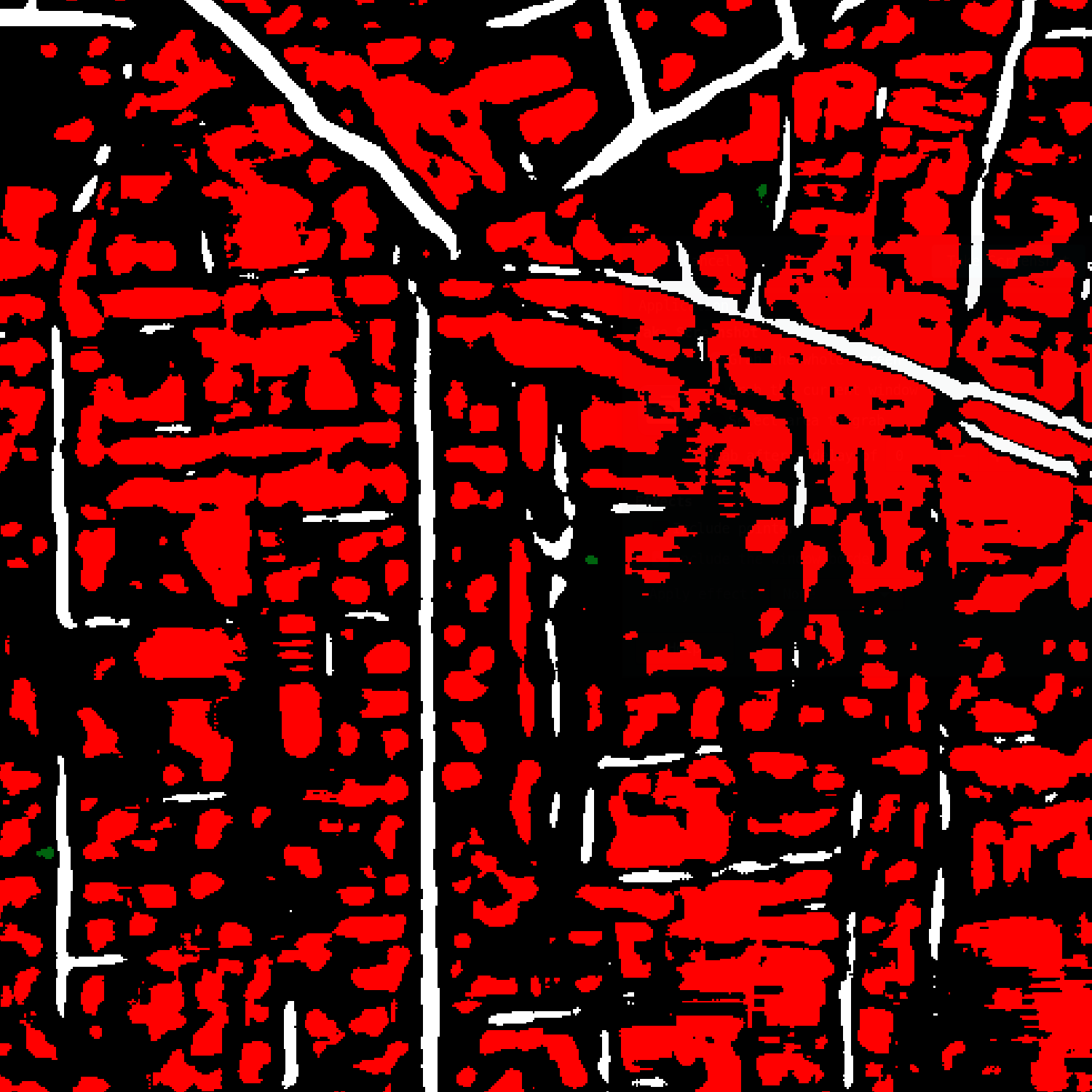}}

\subfloat[UNIT~\cite{liu2017unsupervised}]{\includegraphics[width = 0.165\linewidth]{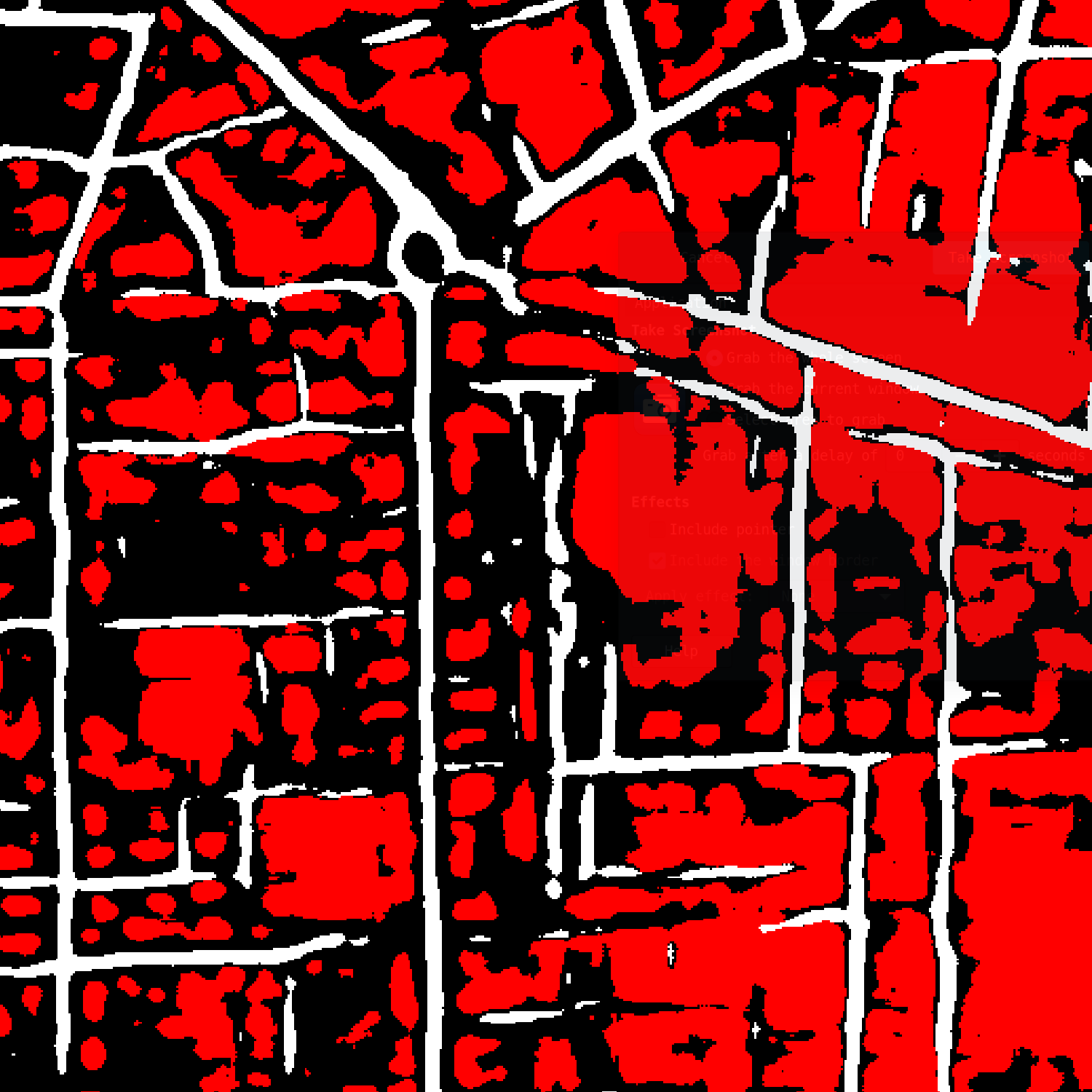}}
\hfill
\subfloat[MUNIT~\cite{huang2018multimodal}]{\includegraphics[width = 0.165\linewidth]{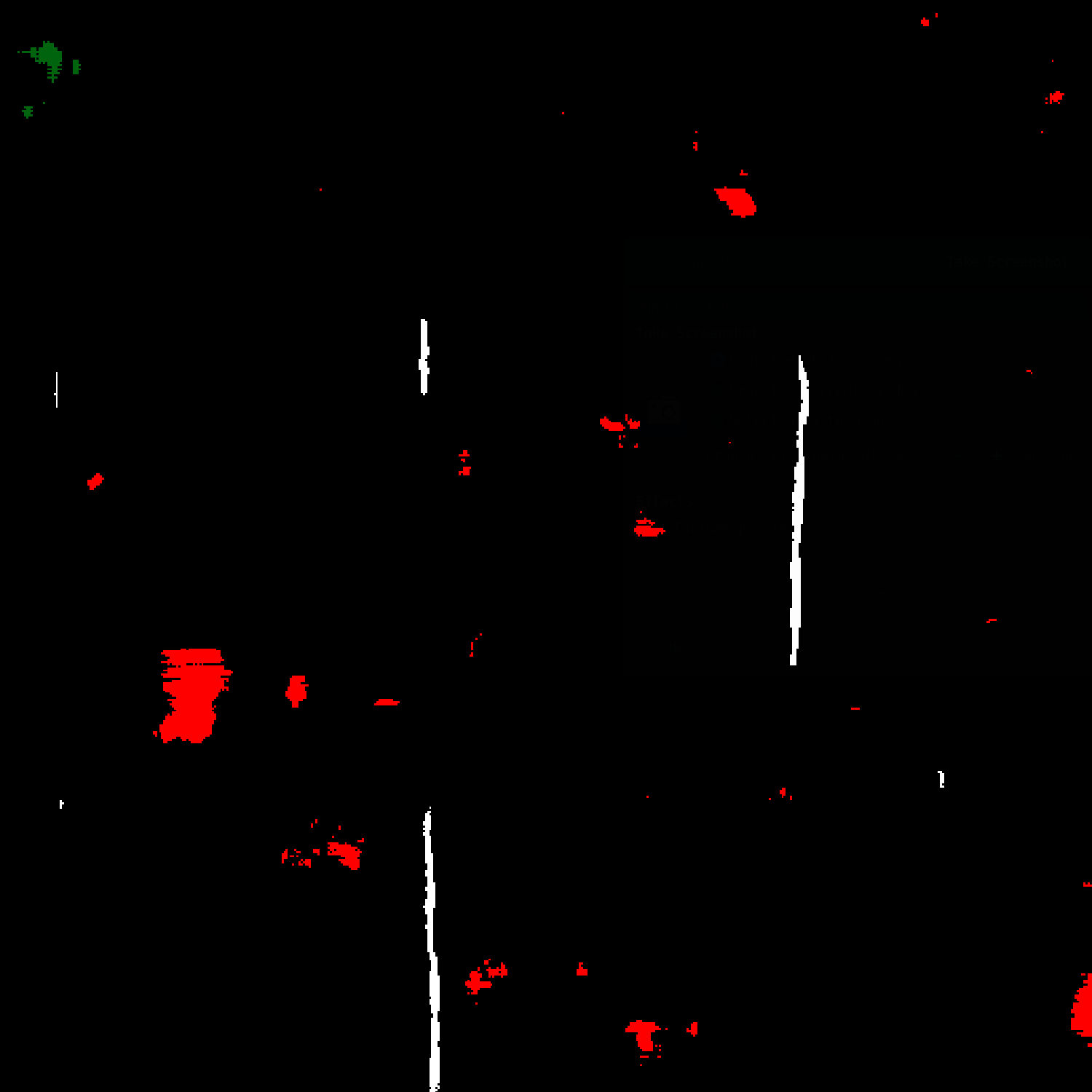}}
\hfill
\subfloat[DRIT~\cite{lee2018diverse} ]{\includegraphics[width = 0.165\linewidth]{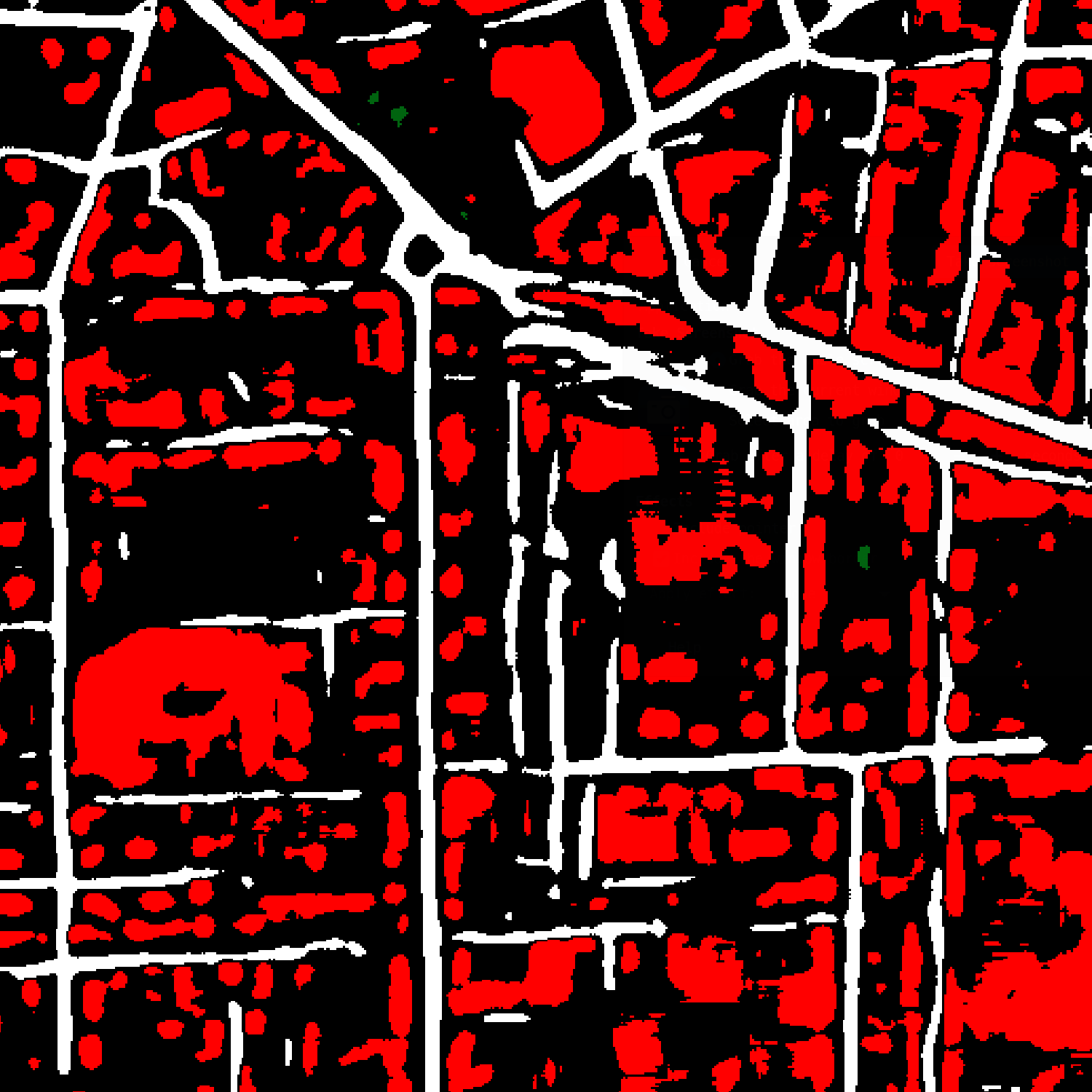}}
\hfill
\subfloat[Gray world~\cite{buchsbaum1980spatial}]{\includegraphics[width = 0.165\linewidth]{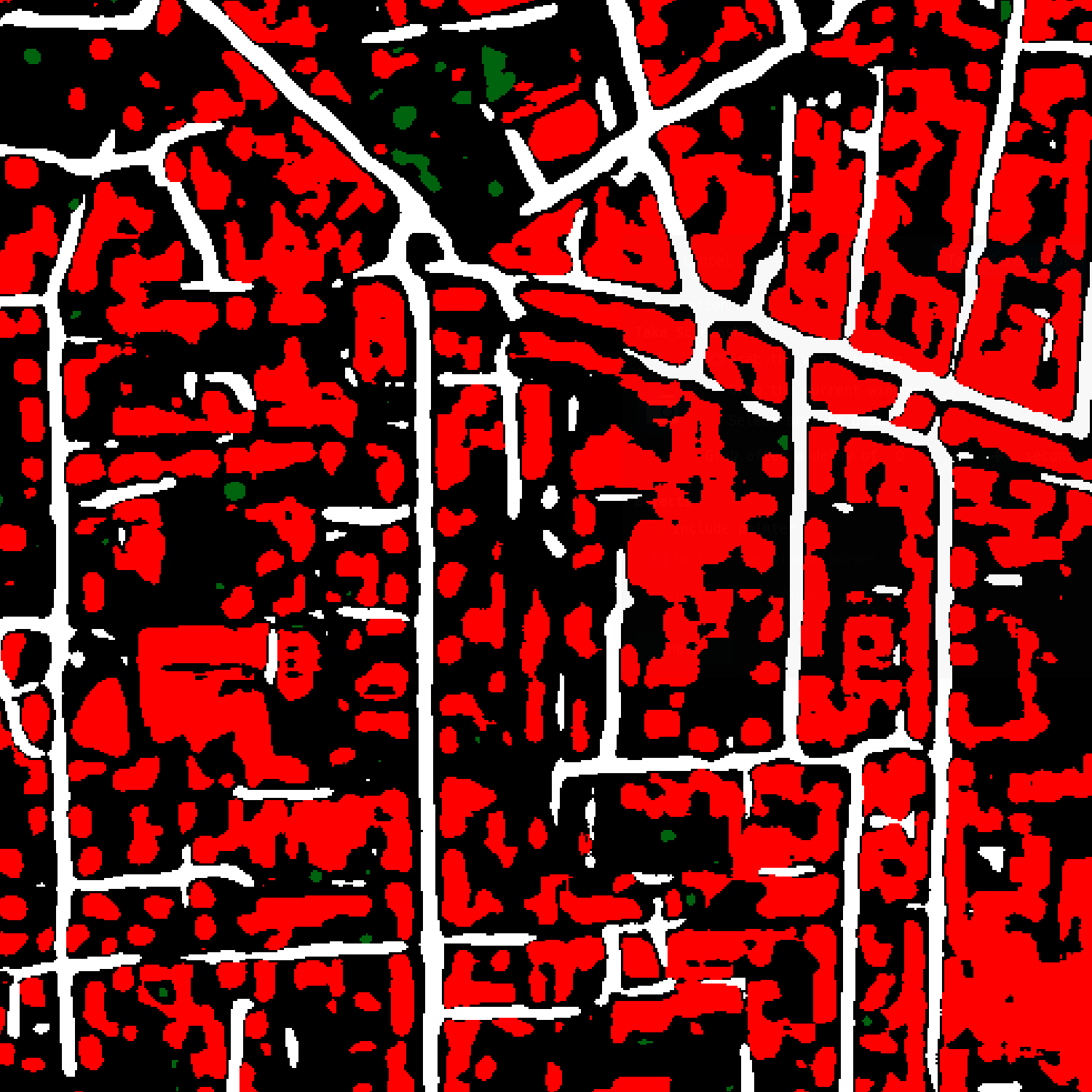}}
\hfill
\subfloat[Hist. match.~\cite{Gonzalez}]{\includegraphics[width = 0.165\linewidth]{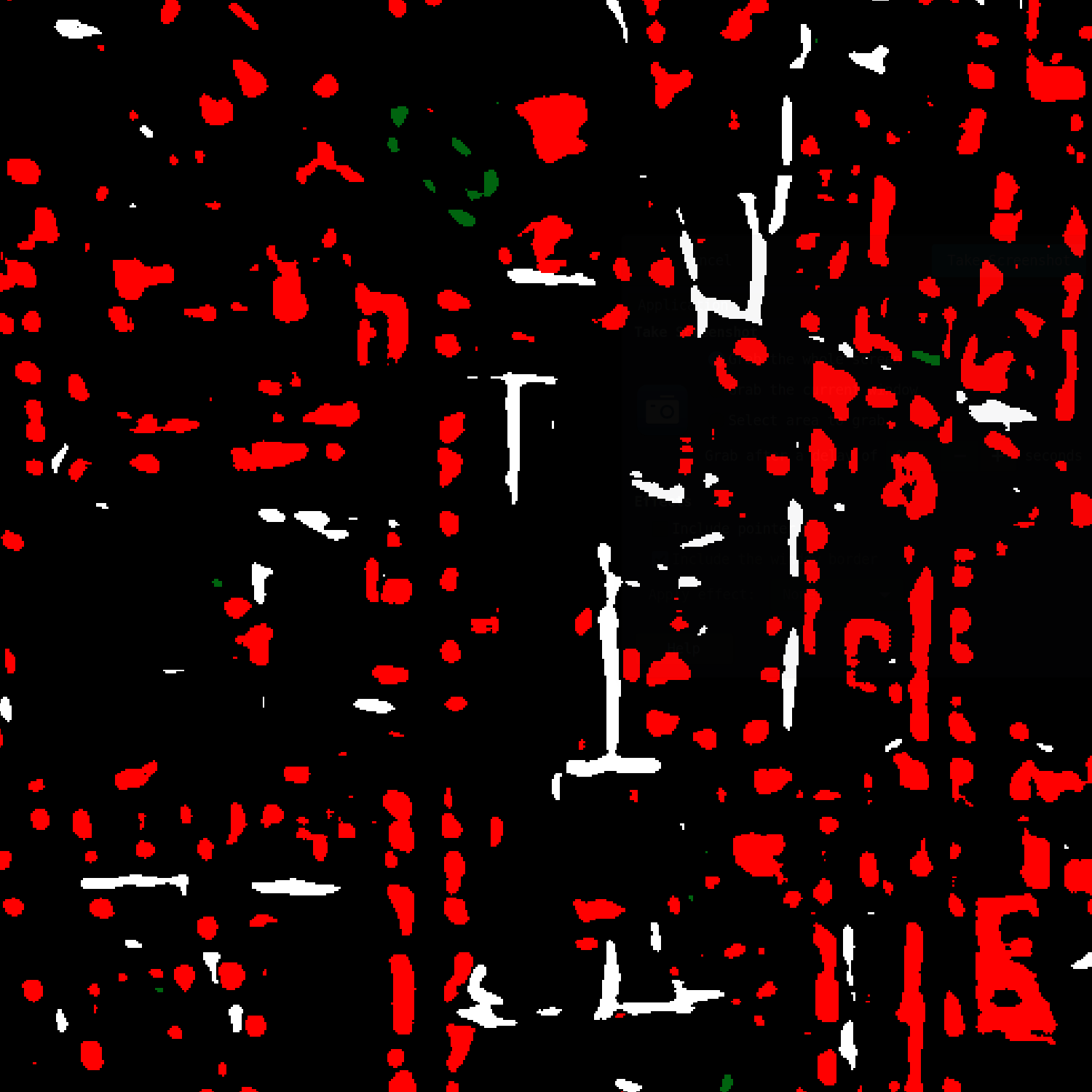}}
\hfill
\subfloat[ColorMapGAN]{\includegraphics[width = 0.165\linewidth]{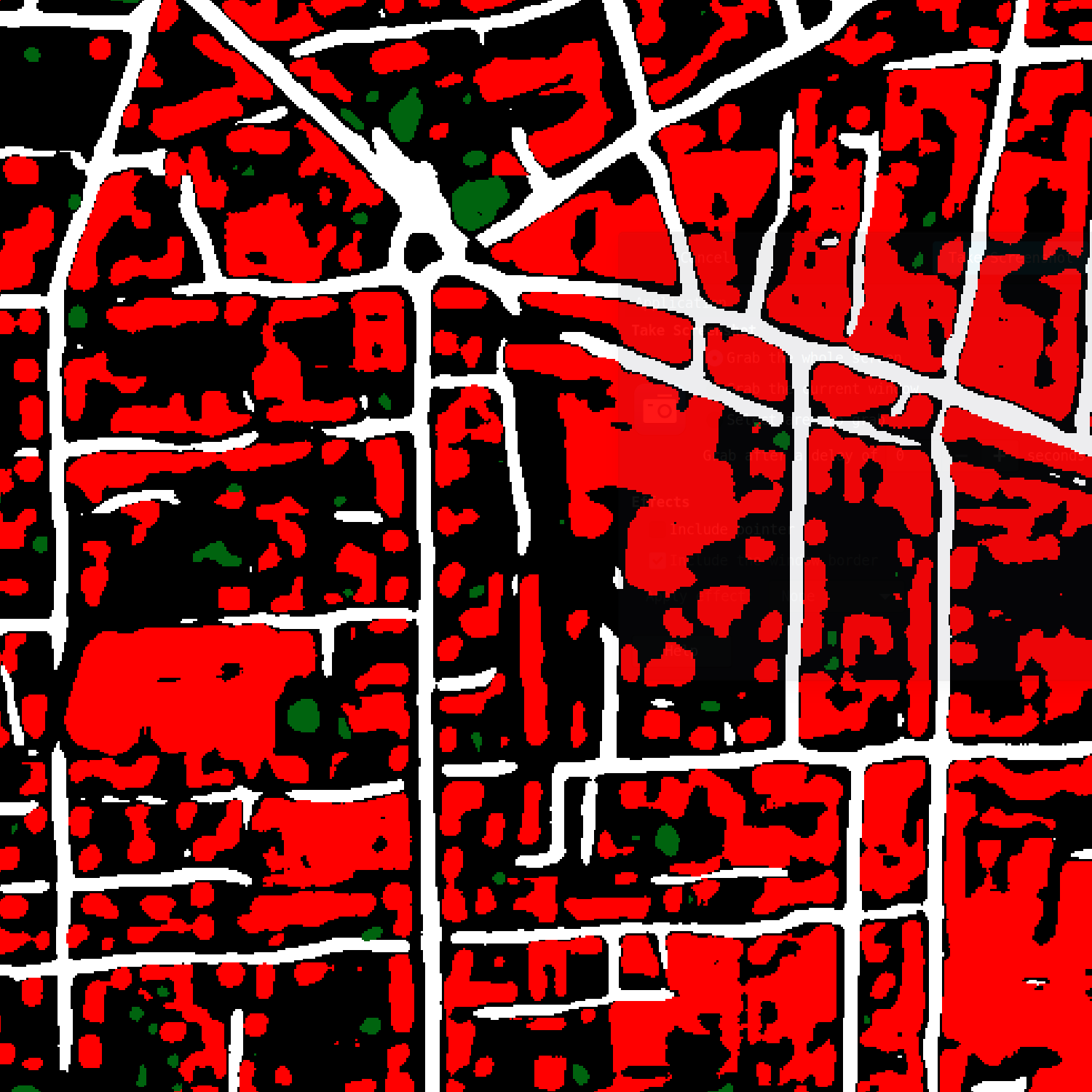}}
\caption{\textit{Roanne}, ground-truth, and the predictions. \textit{Building, road, tree}, and background classes are represented by red, white, green, and black colors.}
\label{fig:roanne_preds}
\end{figure*}

Tables~\ref{table:ious_pair1} and \ref{table:ious_pair2} depict IoU values for each method on the test set. To provide as reliable as possible results, we repeat the step 3 in our framework 20 times for each method that generates fake data, and report the average IoU values in the tables. Re-running the step 3 of the framework 20 times is feasible, since we fine-tune the classifier for only 750 iterations in this step. However, AdaptSegNet tries to adapt the classifier to the test data directly. Hence, we need to train a classifier from scratch every time when we repeat the experiment. For this reason, for AdaptSegNet Single and Multi, we show the average IoU values for only 3 runs. Figs.~\ref{fig:bad_ischl_preds} to \ref{fig:roanne_preds} depict the predictions. Because we run the experiments for each method multiple times, we illustrate the predictions that are obtained by majority voting. In these figures, CycleGAN, UNIT, MUNIT, DRIT, Gray world, Hist. match., and ColorMapGAN represent the results for our framework with these method. We do not add "our framework with" statement in the captions for the sake of simplicity.

Figs.~\ref{fig:bad_ischl_original_fake} to \ref{fig:roanne_original_fake} illustrate some parts of the original training images from the pairs and the fake training images generated by CycleGAN, UNIT, MUNIT, DRIT, gray world, histogram matching, and ColorMapGAN. From the images, we can clearly see that MUNIT and DRIT spoil the semantic identity of the images completely. For instance, the structures indicated by yellow and green rectangles in the original images are either replaced by other objects or distorted in the fake images. In some cases, UNIT seems to generate better results. For instance, the fake cities generated by UNIT in pair 2 are semantically relatively consistent with the original images, and style-wise similar to the test cities. On the other hand, the fake cities in pair 1 have plenty of artificial objects that do not exist in the original city. For UNIT, MUNIT, and DRIT, since the fake images and the ground-truth for the original images usually do not match, the network learns wrong information in the step 3 of our framework. Another problem with these approaches is that we observe tiling effect in the fake images. Especially in the fake \textit{Villach} images, the transition between the patches is not continuous. The reason is that since it is impossible to fit the large satellite images to GPU directly when generating fake images; we need to generate fake data from each patch and combine them to get the entire fake city. However, these approaches mostly generate irrelevant output for the neighboring patches. As a result, the proposed framework with these methods perform poorly on the test data, as confirmed by Tables~\ref{table:ious_pair1}, \ref{table:ious_pair2} and Figs.~\ref{fig:bad_ischl_preds} to \ref{fig:roanne_preds}.

As can be seen in Figs.~\ref{fig:bad_ischl_original_fake} to \ref{fig:roanne_original_fake}, the spectral difference between the training and the test images can be reduced by standardizing them using the gray world algorithm. However, between the fake images in each city pair, we still observe a spectral shift; it is not completely corrected. In consequence, the performance of gray world algorithm is mostly better than UNIT, MUNIT, and DRIT, but it is not as good as desired.

The network architecture used in AdaptSegNet single is very deep; therefore, aligning only the outputs of the final classification layer for the training and the test images does not yield a good performance. However, if the alignment is performed in multiple layers, a better performance could be obtained, especially when the objects of interests cover a large area such as forests. For instance, most of the trees in \textit{Villach} are located in forests areas. For \textit{tree} class on this city, IoU value for AdaptSegNet multi is 56.08$\%$, which is slightly lower than the performance of our framework. However, the performance of AdaptSegNet is not satisfactory in segmenting small objects such as building and thin objects like roads.

At the first sight, histogram matching seems to be working well; semantic structures of the training city are well preserved in the fake training city, and the style of the test city is perfectly transferred to the fake training city. However, it is noticeable in Tables~\ref{table:ious_pair1} and \ref{table:ious_pair2} that the quantitative results for this approach are very poor most of the time. Besides, the fake cities generated by CycleGAN and ColorMapGAN in pair 1 look similar. However, still there exists a large gap between the performance of the framework with ColorMapGAN and with CycleGAN. When segmenting \textit{Villach}, for \textit{road} class, the IoU of CycleGAN is around 9$\%$ lower than of ColorMapGAN. Similarly, ColorMapGAN outperforms CycleGAN by 6$\%$ for \textit{building} class, when segmenting \textit{Bad Ischl.} On the other hand, CycleGAN outperforms ColorMapGAN for \textit{tree} class. To better understand the reasons for this performance difference, the comparisons between ColorMapGAN and CycleGAN, and ColorMapGAN and histogram matching need further analysis.

\begin{figure}
\centering
\subfloat[\textit{Villach}]{\includegraphics[width = 0.33\linewidth]{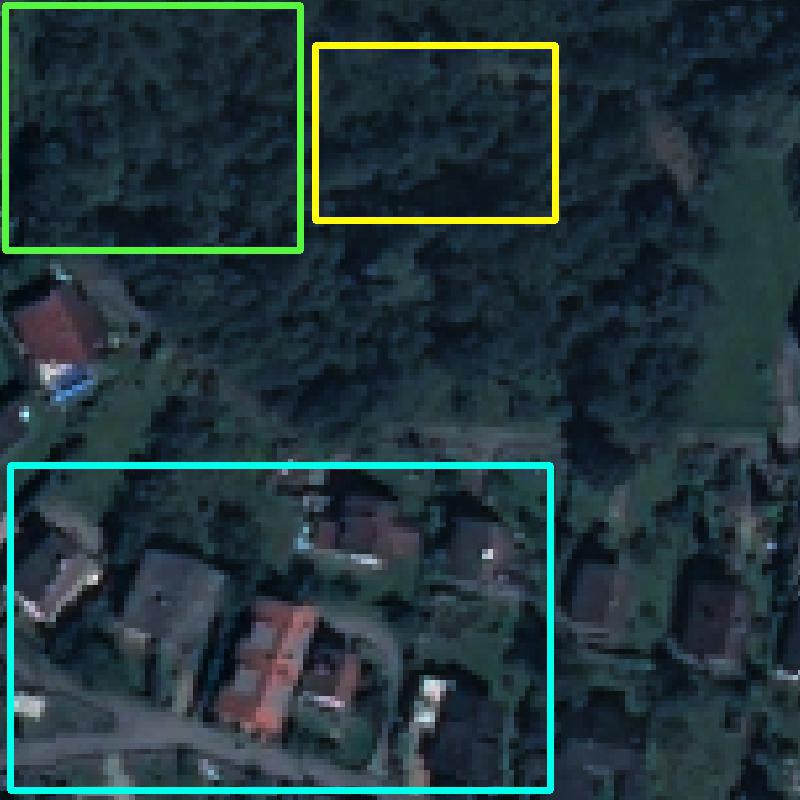}}
\hfill
\subfloat[CycleGAN\cite{zhu2017unpaired}]{\includegraphics[width = 0.33\linewidth]{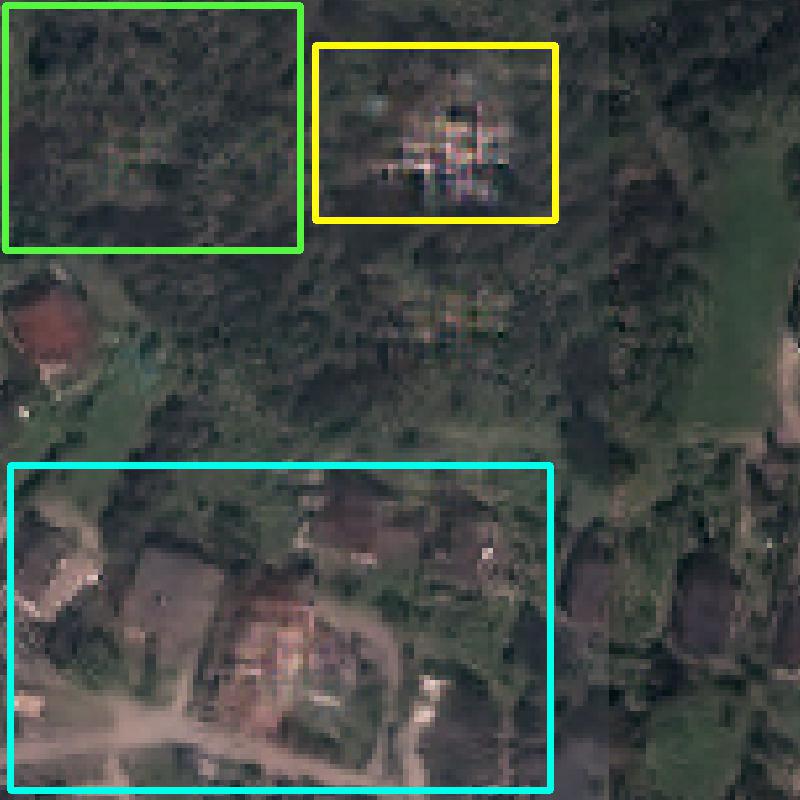}}
\hfill
\subfloat[ColorMapGAN\label{colormapgan_vs_cyclegan_colormapgan}]{\includegraphics[width = 0.33\linewidth]{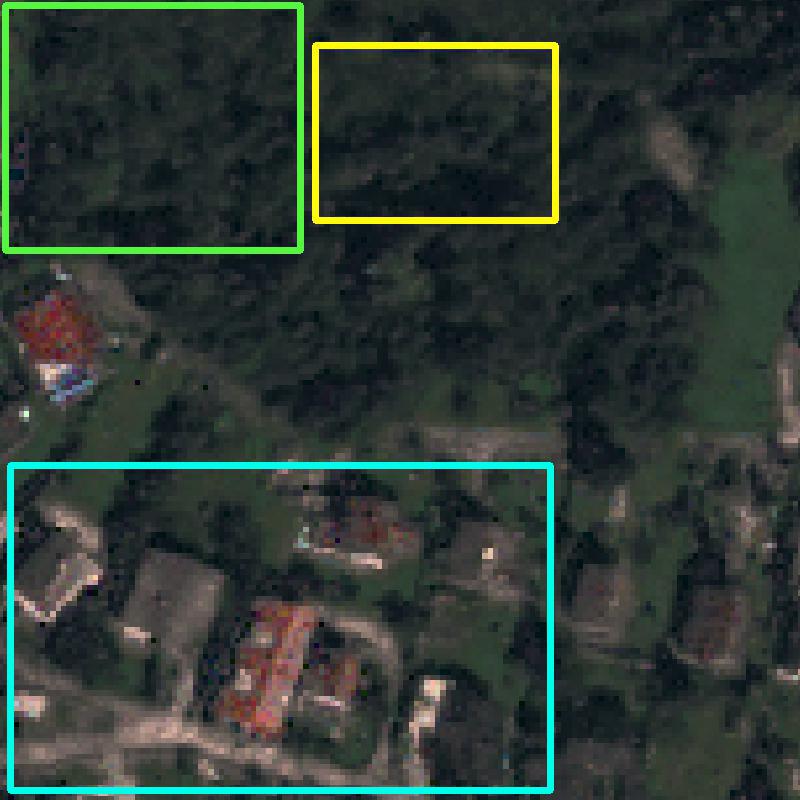}}
\hfill
\caption{A closeup from \textit{Villach} and the corresponding fake images.}
\label{fig:colormapgan_vs_cyclegan}
\end{figure}

\textit{ColorMapGAN vs CycleGAN}~\cite{zhu2017unpaired}: First of all, it is worth noting that the performance of CycleGAN is unstable. The proposed framework with CycleGAN performs unsatisfactorily on pair 2 because of its semantically inconsistent outputs with the original images. As highlighted by yellow rectangles in Figs.~\ref{fig:beziers_original_fake} and \ref{fig:roanne_original_fake}, CycleGAN removes some objects that exist in the original cities. There are several reasons why it performs worse than ColorMapGAN on pair 1 for \textit{building} and \textit{road} classes. Firstly, the resolution of its output is lower than the resolution of the original city and the output of ColorMapGAN. Fig.~\ref{fig:colormapgan_vs_cyclegan} illustrates a closeup from \textit{Villach} and the corresponding fake images generated by ColorMapGAN and CycleGAN. The resolution difference between the fake images can easily be noticed in the outlined areas by cyan rectangles. Learning from blurrier data obviously deteriorates the performance. Secondly, the output of CycleGAN has some artifacts as shown in the same figure by a yellow rectangle. The generator of ColorMapGAN does not have convolution, pooling, etc. operations; therefore, we do not observe such artifacts in the outputs of ColorMapGAN. Finally, since we generate fake cities patch by patch because of memory constraints, there exists a spectral difference between some of the neighboring patches in the fake images generated by CycleGAN (see the green rectangle in Fig.~\ref{fig:bad_ischl_cycle_gan}). This difference leads the network to exhibit a lower performance. ColorMapGAN does not have this problem, because it maps each color to another one. The pixels having the same color are mapped to another exactly the same color, irrespective of their locations. Therefore, the neighboring patches are spectrally consistent, and there is no tiling effect between them. One drawback of ColorMapGAN is that it seems to be slightly smoothing out trees, which results in the patterns on trees disappearing slightly (see the green rectangle in Fig.~\ref{colormapgan_vs_cyclegan_colormapgan}). This is probably why the framework with CycleGAN outperforms the framework with ColorMapGAN on pair 1 for \textit{tree} class.

\begin{figure}
\centering
\subfloat[\textit{Bad Ischl}]{\includegraphics[width = 0.33\linewidth]{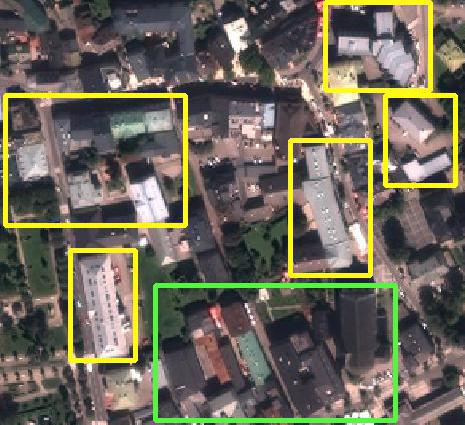}}
\hfill
\subfloat[Hist. matc.~\cite{Gonzalez}]{\includegraphics[width = 0.33\linewidth]{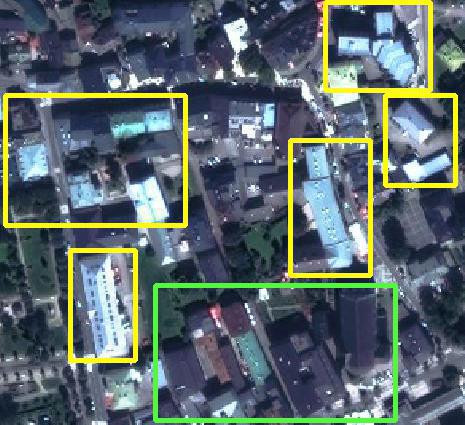}}
\hfill
\subfloat[ColorMapGAN]{\includegraphics[width = 0.33\linewidth]{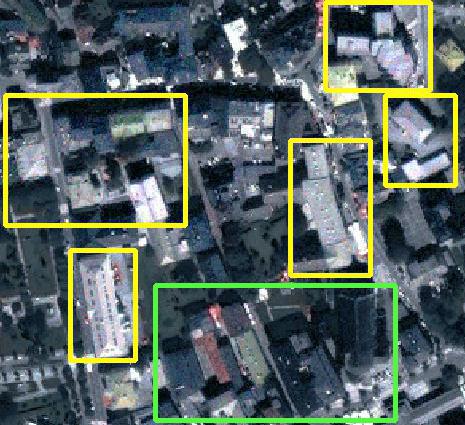}}
\hfill
\caption{A closeup from \textit{Bad Ischl} and the corresponding fake images.}
\label{fig:colormapgan_vs_histmatch}
\end{figure}

\begin{figure}
\centering
\subfloat[\textit{Bad Ischl}]{\includegraphics[width = 0.33\linewidth]{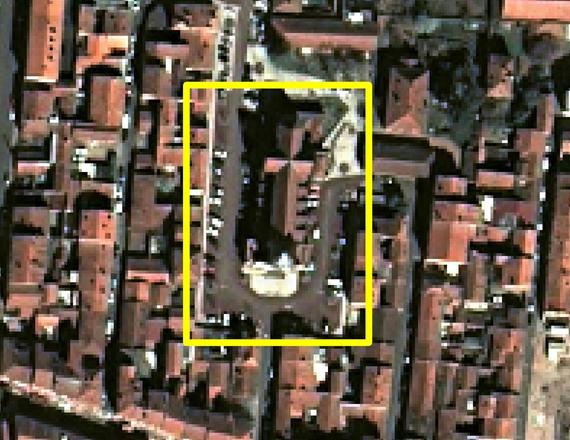}}
\hfill
\subfloat[Hist. matc.~\cite{Gonzalez}]{\includegraphics[width = 0.33\linewidth]{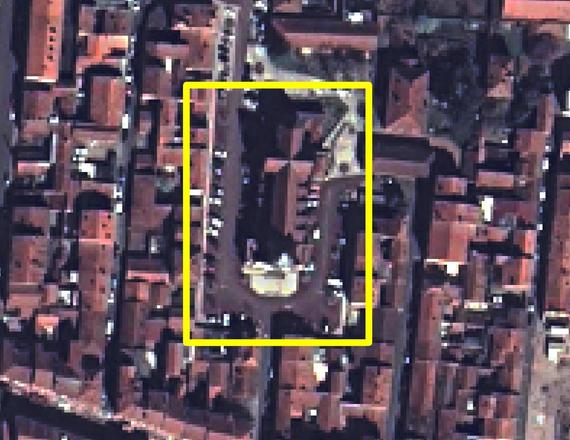}}
\hfill
\subfloat[ColorMapGAN]{\includegraphics[width = 0.33\linewidth]{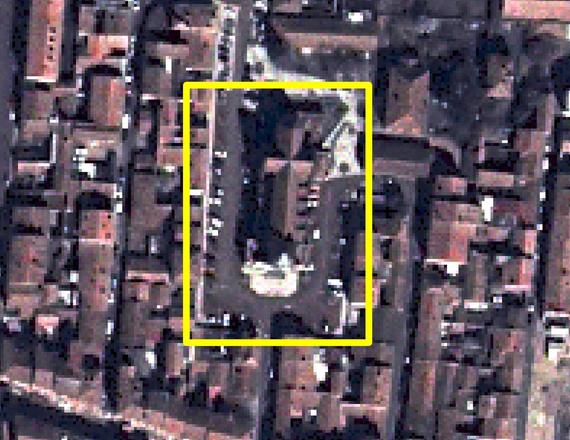}}
\hfill
\caption{A closeup from \textit{Roanne} and the corresponding fake images.}
\label{fig:colormapgan_vs_histmatch2}
\end{figure}

\begin{figure*}
\centering
\subfloat[\textit{Roanne}]{\includegraphics[width = \linewidth]{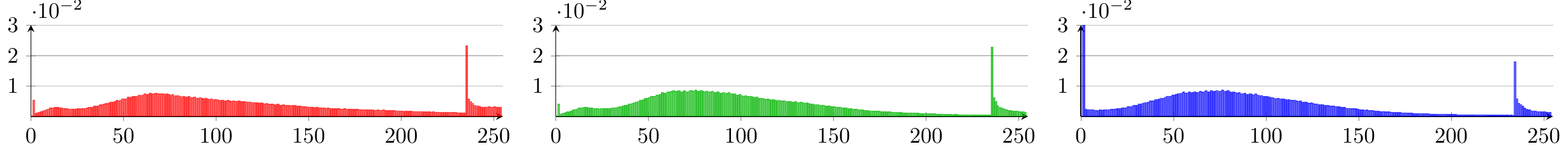}}
\hfill
\subfloat[Fake \textit{Roanne} generated by histogram matching\label{fig:histograms_hist_match}]{\includegraphics[width = \linewidth]{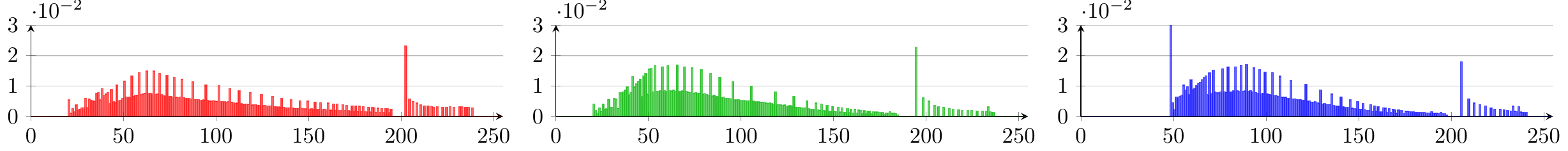}}
\hfill
\subfloat[Fake \textit{Roanne} generated by ColorMapGAN]{\includegraphics[width = \linewidth]{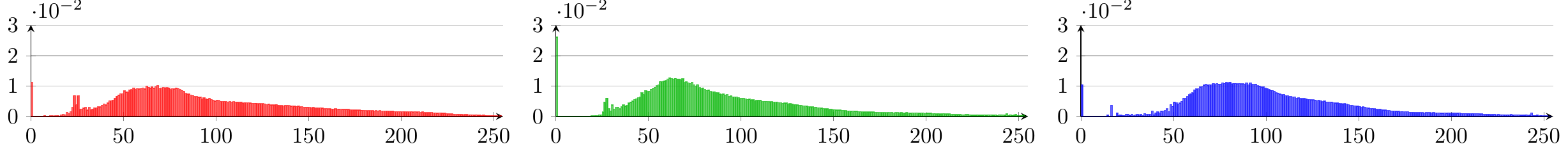}}
\hfill
\subfloat[\textit{B{\'e}ziers}]{\includegraphics[width = \linewidth]{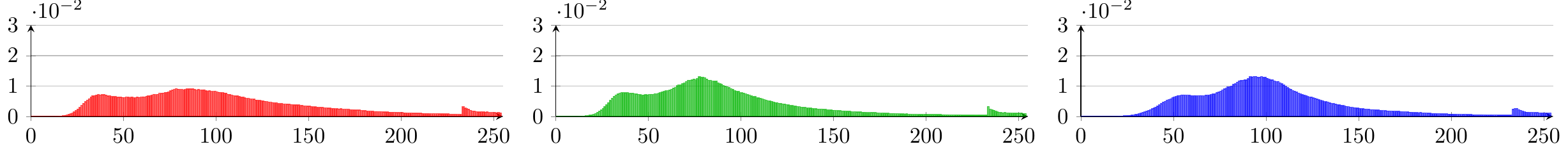}}
\hfill
\caption{Color histograms of building pixels. Red, green, and blue bins represent the histograms for red, green, and blue channels, respectively.}
\label{fig:histograms}
\end{figure*}

\textit{ColorMapGAN vs Histogram matching}~\cite{inamdar2008multidimensional}: The main problem of histogram matching is that it does not take into account the contextual information, it only tries to match the histogram of the whole training city with the histogram of the whole test city. Conversely, the discriminator of ColorMapGAN extracts high level features from the output of the generator and the test data to decide which one is real and which one is fake. In other words, the generator generates a fake training city in a way that its high level features align with the high level features extracted from the test city. For this reason, the proposed framework with ColorMapGAN yields substantially improved results. As shown in Fig.~\ref{fig:colormapgan_vs_histmatch} by yellow rectangles, when generating fake \textit{Bad Ischl}, histogram matching converts some gray rooftops to cyan ones, whereas ColorMapGAN keeps them gray. Similarly, in the same figure, we observe that the buildings highlighted by green rectangles have dark violet rooftops in the output of histogram matching, and black rooftops in the image generated by ColorMapGAN. In Fig.~\ref{fig:villach_original}, we can see that there is no building having a cyan or dark violet rooftop in \textit{Villach}, but there exists lots of buildings with gray or black rooftops. If the generator of ColorMapGAN generated cyan or dark violet colored rooftops, the discriminator would easily understand that these buildings were fake. For this reason, such buildings do not appear in the output of ColorMapGAN. Similarly, in the process of generating fake \textit{Roanne}, histogram matching algorithm generates some reddish roads, as shown in Fig.~\ref{fig:colormapgan_vs_histmatch2} by a yellow rectangle. In the same figure, we see that ColorMapGAN outputs gray roads. Moreover, ColorMapGAN generates buildings having brownish colored rooftops, which are probably more representative for the buildings in \textit{B{\'e}ziers} than the buildings with red rooftops generated by histogram matching. As discussed in Sec.~\ref{sec:introduction}, CNNs are extremely sensitive to the training data. A small domain shift between the training and the test data may affect the results significantly. Furthermore, in Fig.~\ref{fig:histograms}, we depict color histograms of the buildings in \textit{Roanne}, fake \textit{Roanne} generated by histogram matching and ColorMapGAN, and \textit{B{\'e}ziers.} Since \textit{Roanne} and \textit{B{\'e}ziers} are two different cities, we cannot expect the histograms of an ideal fake \textit{Roanne} and \textit{B{\'e}ziers} to be exactly the same. However, we expect them to resemble each other. Although histogram matching algorithm tries to match the histogram of the training city with the histogram of the test city, there exists a large difference between the class-wise histograms of the fake training and the test cities. As can be seen in Fig.~\ref{fig:histograms_hist_match}, there is a large deviation between some of the neighboring bins in the histogram. In contrast, color histograms of the buildings in the fake \textit{Roanne} generated by ColorMapGAN are more similar to the histograms of the buildings in \textit{B{\'e}ziers}. For histogram matching, we observe the same issue for the other classes as well, but we do not include the histograms of the other classes because of the lack of space.

\subsection{Running Times}
The proposed framework, ColorMapGAN, CycleGAN, UNIT, MUNIT, and DRIT were implemented in Tensorflow\footnote{https://www.tensorflow.org}. We conducted all the experiments on an Nvidia Geforce GTX1080 Ti GPU with 11 GB of RAM. Table \ref{table:running_times_learning_based} reports the running times for training ColorMapGAN and the other learning based compared methods for 1 iteration. The table demonstrates that the training time for ColorMapGAN is significantly shorter than the other learning based approaches. Let us also remark again that we optimize ColorMapGAN for only 8,000 iterations. In other words, we need only about 6.5 minutes to train it. On the other hand, the other learning based approaches with the default parameters requires long hours to train. It is also notable that non-learning based approaches generate an output in a substantially shorter time. The execution time for gray world algorithm is almost instant, and histogram matching needs less than half a minute. However, the quality of the results for non-learning based methods is deficient.
\begin{table}
\centering
\caption{Training times for generating fake cities.}
\label{table:running_times_learning_based}
\begin{tabular}{||c|c|}
\hline			
\textbf{Method}                  & \textbf{Training time for 1 Iter. (secs.)} \\
\hline
CycleGAN~\cite{zhu2017unpaired}  &  0.11         \\
UNIT~\cite{liu2017unsupervised}  &  0.47         \\    
MUNIT~\cite{huang2018multimodal} &  0.45         \\
DRIT~\cite{lee2018diverse}       &  0.29         \\
ColorMapGAN                      & \textbf{0.05} \\
\hline			
\end{tabular}
\end{table}

\begin{table}
\centering
\caption{Execution times for generating fake cities.}
\label{table:running_times_non_learning_based}
\begin{tabular}{||c|c|c||}
\hline			
\multirow{2}{*}{\textbf{City}} & \multicolumn{2}{c||}{\textbf{Execution time (seconds)}} \\
\cline{2-3}
                               & Gray world\cite{buchsbaum1980spatial} & Histogram matching\cite{Gonzalez} \\
\hline
\textit{Bad Ischl}             & \textbf{1.46} & 19.77 \\
\textit{Villach}               & \textbf{1.89} & 26.78 \\
\textit{B{\'e}ziers}           & \textbf{1.37} & 18.05 \\
\textit{Roanne}                & \textbf{1.24} & 20.32 \\
\hline			
\end{tabular}
\end{table}

\section{Concluding Remarks}
In this work, we presented a novel framework to generate high quality maps from satellite images even when there exists a large domain shift between spectral bands of the training and the test images. We validated our approach on two city pairs, where we performed a city to city domain adaptation in each pair. Our experimental results demonstrated that the proposed framework exhibits a much better performance than the existing approaches. We also showed that the proposed ColorMapGAN generates fake images in a significantly shorter time than some of the competitive unpaired image to image translation methods in the computer vision community.

A possible future direction could be to investigate a more difficult domain adaptation problem, where in addition to the large spectral difference, shapes of the objects such as buildings in the training and test images are considerably different.

\section*{Acknowledgment}
The authors would like to thank \href{https://www.acri-st.fr/}{ACRI-ST} and \href{https://cnes.fr/en}{CNES} for initializing and funding this study. The authors also thank Alain Giros and S{\'e}bastien Clerc for fruitful discussions.

\bibliographystyle{IEEEtran}
\bibliography{refs}

\begin{IEEEbiography}[{\includegraphics[width=1in,
height=1.25in,clip,keepaspectratio]{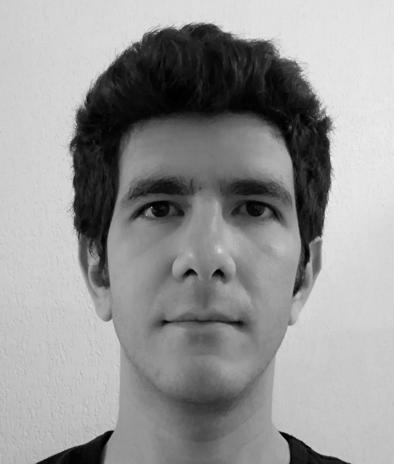}}]{Onur Tasar}
received the B.S. degree in computer engineering department from Hacettepe University, Ankara, Turkey in 2014, and the M.S. degree in computer engineering department from Bilkent University, Ankara, Turkey in 2017. He is currently working towards his Ph.D. at Inria Sophia Antipolis-M\'editerran\'ee within TITANE team, Valbonne, France.

His research interests include computer vision, machine learning, and computational geometry with applications to remote sensing.
\end{IEEEbiography}

%\vspace{-8mm}

\begin{IEEEbiography}[{\includegraphics[width=1in,
height=1.25in,clip,keepaspectratio]{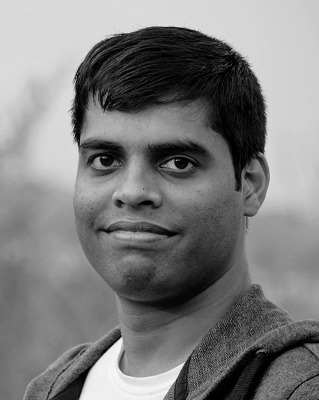}}]{S L Happy } has completed the joint MS -- PhD degree from Indian Institute of Technology Kharagpur, India in 2018. Currently, he is working as a postdoctoral researcher at Inria Sophia Antipolis, France. His research interests include machine learning, computer vision, hyperspectral image classification, medical image analysis, and facial expression analysis.
\end{IEEEbiography}

%\vspace{-8mm}

\begin{IEEEbiography}[{\includegraphics[width=1in,
height=1.25in,clip,keepaspectratio]{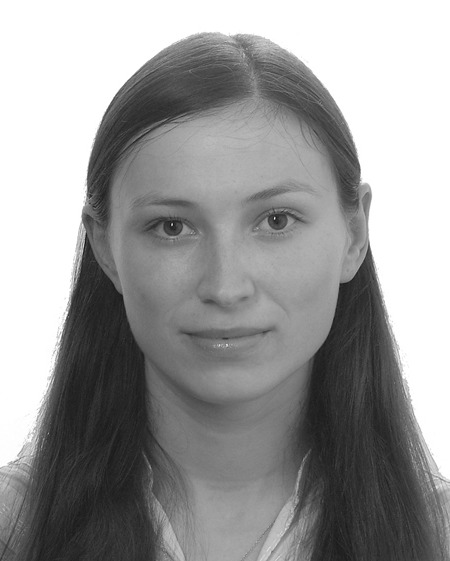}}]{Yuliya Tarabalka}

(S'08--M'10) received the B.S. degree in computer science from Ternopil Ivan Pul'uj State Technical University, Ukraine, in 2005 and the M.Sc. degree in signal and image processing from the Grenoble Institute of Technology (INPG), France, in 2007. She received a joint Ph.D. degree in signal and image processing from INPG and in electrical engineering from the University of Iceland, in 2010.

From July 2007 to January 2008, she was a researcher with the Norwegian Defence Research Establishment, Norway. From September 2010 to December 2011, she was a postdoctoral research fellow with the Computational and Information Sciences and Technology Office, NASA Goddard Space Flight Center, Greenbelt, MD. From January to August 2012 she was a postdoctoral research fellow with the French Space Agency (CNES) and Inria Sophia Antipolis-M\'editerran\'ee, France. From 2012 to 2019 she was a researcher with the TITANE team of Inria Sophia Antipolis-M\'editerran\'ee. She is currently the research director of \href{https://luxcarta.com}{LuxCarta Technology}. Her research interests are in the areas of image processing, pattern recognition and development of efficient algorithms. She is Member of the IEEE Society.
\end{IEEEbiography} 

%\vspace{-8mm}

\begin{IEEEbiography}[{\includegraphics[width=1in,height=1.25in,clip,keepaspectratio]{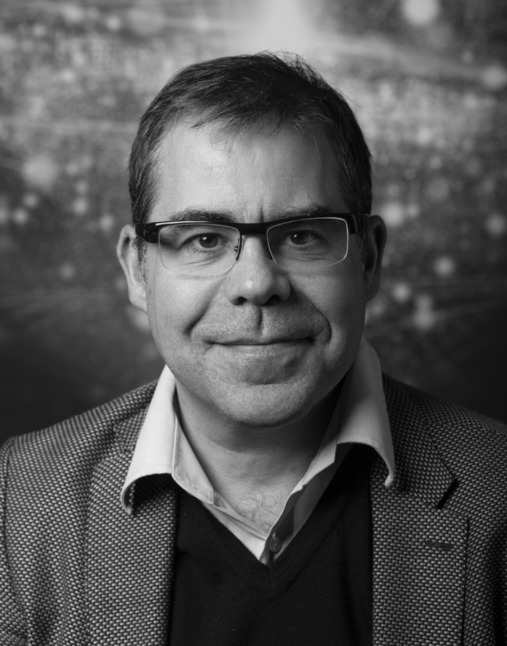}}]{Pierre Alliez}
is Senior Researcher and team leader at Inria Sophia-Antipolis - Mediterranee. He has authored scientific publications and several book chapters on mesh compression, surface reconstruction, mesh generation, surface remeshing and mesh parameterization. He was awarded in 2005 the EUROGRAPHICS young researcher award for his contributions to computer graphics and geometry processing. He was co-chair of the Symposium on Geometry Processing in 2008, of Pacific Graphics in 2010 and Geometric Modeling and Processing 2014. He was awarded in 2011 a Starting Grant from the European Research Council on Robust Geometry Processing.
\end{IEEEbiography}

\end{document}